\documentclass[draftcls,12pt,onecolumn]{IEEEtran} 
\usepackage{booktabs}
\usepackage{multirow}
\usepackage{amsfonts}
\usepackage{amssymb}
\usepackage{amsbsy} 
\usepackage{amsmath}
\usepackage[ansinew]{inputenc}%
\usepackage{cite}
\usepackage{multirow}
\usepackage{color}
\usepackage[usenames,dvipsnames]{xcolor}
\usepackage{lscape}
\usepackage{graphicx}
\usepackage{colortbl}
\usepackage{hyperref}

\usepackage{algorithm}
\usepackage{algorithmic}
\usepackage{slashbox}
\def\X{\ensuremath{\mathbf{X}}}
\def\A{\ensuremath{\mathbf{A}}}
\def\B{\ensuremath{\mathbf{B}}}

\def\G{\ensuremath{\mathbf{G}}}
\def\L{\ensuremath{\mathbf{L}}}
\def\S{\ensuremath{\mathbf{S}}}
\def\D{\ensuremath{\mathbf{D}}}
\def\W{\ensuremath{\mathbf{W}}}

\def\F{\ensuremath{\mathbf{F}}}

\def\x{\ensuremath{\mathbf{x}}}

\usepackage{eso-pic}

\newcommand{\f}{{\mathbf f}}

\definecolor{resi}{RGB}{255,140,0}
\definecolor{comm}{RGB}{255,0,0}
\definecolor{mead}{RGB}{34,139,34}
\definecolor{harv}{RGB}{222,184,135}
\definecolor{bare}{RGB}{139,69,19}
\definecolor{road}{RGB}{190,190,190}
\definecolor{pool}{RGB}{0,0,205}
\definecolor{park}{RGB}{111,111,111}
\definecolor{tree}{RGB}{124,252,0}
\newcommand{\blue}[1]{\textcolor[rgb]{0,0,0}{#1}}
\newcommand{\blueR}[1]{\textcolor[rgb]{0,0,0}{#1}}
\newcommand{\blueRR}[2]{\textcolor[rgb]{0,0,0}{#2}}

\newcommand{\red}[1]{\textcolor[rgb]{1,0,0}{#1}}
\newcommand{\white}[1]{\textcolor[rgb]{1,1,1}{#1}}

\newcommand{\orange}[1]{\textcolor[rgb]{1,0.5,0}{#1}}

\newcommand{\mytitle}{Semisupervised Manifold Alignment of Multimodal Remote Sensing Images} 

\begin{document}

\title{\mytitle}
\author{Devis Tuia,~\IEEEmembership{Member,~IEEE},
Michele Volpi,~\IEEEmembership{Student Member,~IEEE},\\ Maxime Trolliet, and Gustau Camps-Valls,~\IEEEmembership{Senior Member,~IEEE}
\thanks{Manuscript received 2013;}
\thanks{This work has been partly supported by the Swiss National Science Foundation (grants PZ00P2-136827 (http://p3.snf.ch/project-136827) and P2LAP2-148432 (http://p3.snf.ch/Project-148432)) and by the spanish Ministry of Economy and Competitiveness (MINECO) under project LIFE-VISION TIN2012-38102-C03-01.}
\thanks{DT and MT are with the Laboratoire des Syst\`emes d'Information G\'eographique (LaSIG), Ecole Polytechnique F{\'e}d{\'e}rale de Lausanne (EPFL), Switzerland. devis.tuia@epfl.ch, http://devis.tuia.googlepages.com, Phone: +41-216935785, Fax : +41-216935790.
\newline MV was with the Centre for Research on Terrestrial Environment, Universit{\'e} de Lausanne, Switzerland. He is now with the CALVIN at the Institute of Perception, Action and Behaviour, the University of Edinburgh, United Kingdom. Telephone: +44 (0) 131 650 8741, Fax: +44 (0) 131 651 5651, Email: michele.volpi@ed.ac.uk, Web: https://sites.google.com/site/michelevolpiresearch. 
\newline GCV is with the Image Processing Laboratory (IPL), Universitat de Val\`encia, C/ Catedr\'atico A. Escardino, 9 - 46980 Paterna, Val\`encia (Spain). gustavo.camps@uv.es, http://isp.uv.es, Phone: +34-963544064, Fax: +34-963543261.
}}

\markboth{Preprint. Published in IEEE Transactions on Geoscience and Remote Sensing (10.1109/TGRS.2014.2317499)}{Tuia et al.: Semisupervised Manifold Alignment}

\maketitle

\begin{abstract}
\textbf{This is the pre-acceptance version, to read the final version published in 2014 in the IEEE Transactions on Geoscience and Remote Sensing (IEEE TGRS), please go to: \href{https://doi.org/10.1109/TGRS.2014.2317499}{10.1109/TGRS.2014.2317499}}\\
We introduce a method for manifold alignment of different modalities (or domains) of remote sensing images. The problem is recurrent when a set of multitemporal, multisource, multisensor and multiangular images is available. In these situations, images should ideally be spatially coregistred, corrected and compensated for differences in the image domains. Such procedures require the interaction of the user, involve tuning of many parameters and heuristics, and are usually applied separately. Changes of sensors and acquisition conditions translate into shifts, twists, warps and foldings of the image distributions (or manifolds). The \blue{proposed semisupervised manifold alignment (SS-MA)} method aligns the images working directly on their manifolds, and is thus not restricted to images of the same resolutions, either spectral or spatial. SS--MA pulls close together samples of the same class while pushing those of different classes apart. At the same time, it preserves the geometry of each manifold along the transformation. The method builds a linear invertible transformation to a latent space where all images are alike, and reduces to solving a generalized eigenproblem of moderate size. 
We study the performance of SS--MA in toy examples and in real multiangular, multitemporal, and multisource image classification problems. The method performs well for strong deformations and \blue{leads to accurate classification for all domains}. A Matlab implementation of the proposed method is provided at \url{http://isp.uv.es/code/ssma.htm}.
\end{abstract}


\section{Introduction}
\label{sec:intro}

\PARstart{R}{emote} sensing analysts can nowadays exploit images of unprecedented spatial resolution (e.g. higher than the meter for commercial sensors such as WorldView 2) with increasing revisit time (which will be increased with the upcoming Sentinels)~\cite{Cam11,Pra11}. With these very high resolution (VHR) images, it becomes possible to perform a \blue{large variety} of monitoring studies, since the geographical area of interest can be covered with high frequency. \blue{This led to a variety of} techniques for multitemporal~\cite{Cam08,Pet12} and multiangular~\cite{Lon12,Dor12} data processing \blue{for VHR data}.

The increased access to information also induces a series of processing problems for the \blue{analyst} performing the monitoring task: the images undergo a series of spectral distortions related to variations in the acquisition geometry, the atmospheric conditions, and the acquisition angle. To these acquisition-related differences, we must add a series of temporal specificities, such as the variations in the phenological cycle, or the appearance of small objects contaminating the class signature (flower pots on roofs, for example). All these changes in acquisition conditions and geometry, as well as differences in the properties of the sensor\blue{s}, \blue{produce} local changes in the probability distribution function (PDF) of the images from one acquisition to another one, which in turn affect the performances of the classifier\blue{, when predicting data from another domain}. As a consequence, the tempting direct application of a classifier optimal for one scene to another scene can lead to catastrophic results. The application of a classifier to a series of newly acquired scenes remains a priority for image analysts, who would like to reuse the available labeled samples, minimize the time/effort devoted to photointerpretation (or terrestrial campaigns) and eventually use images acquired by other sensors. Therefore, a classifier should be applicable to new scenes regardless of the sensor and acquisition specificities. This property of {\em portability} is of the greatest importance, especially in the VHR context or when considering archives of images, for which labels are often unavailable.

To meet this objective, research has considered strategies of \emph{adaptation}. By adaptation, we mean the ability of a method to modify its characteristics to new scenes acquired under different acquisition conditions, but representing a similar problem (typically sharing the same classes). Adaptation has been carried out in three main ways in remote sensing data processing: 1) at the level of the classifier, 2) to encode invariances of interest, and 3) at the level of the data representation. Let us briefly review the three families of approximations.

Regarding the adaptation of the classifier, most strategies are issued from  semi-supervised learning: the adaptation is usually performed \blue{either} by modifying the weights of the classifier using unlabeled data coming from the PDF of the image to be classified~\cite{Gom08f,Bru10,Kim10}, \blue{by} spatial regularization~\cite{Jun11}, or \blue{by} adding few informative labeled examples carefully chosen from the new image~\cite{Tui11d,Mat12}. Th\blue{is family of approaches often} comes with several free parameters, \blue{requires} expertise in machine learning and statistics, and \blue{involves} high computational costs.

When considering the incorporation of invariances in the classifier, one aims at making the classifier robust to variations in the data representation. In the specific example of remote sensing images, such variations can be rotations (objects can be arbitrarily oriented), presence of shadows (attenuations of the signal), and scale (objects of the same class can have different sizes, or the new image may have different spatial resolution), just to name a few. Invariance to these changes can be achieved in several ways. \blue{For some of these effects,} one may develop physical models~\cite{Gua10}\blue{, for example to remove atmospheric of illumination effects}. This, however, requires the accurate modeling of the physical processes involved \blue{and a detailed knowledge of the atmospheric conditions at the time of acquisition}. Alternatively, the classifier can become invariant if one includes in the training set synthetic examples representing the phenomena that the classifier should be invariant to. For example, if using patches of the images as inputs, adding rotated versions of them will force invariance to rotation of the objects represented in the patches~\cite{Ver13}. While good results are obtained in general, it becomes complex to encode several invariances simultaneously that cover a reasonable subspace of the image manifold.

The third \blue{family} is the one considered in this paper. The rationale is to modify the data representation to perform the adaptation. Such a modification can be driven by physical properties of the atmosphere~\cite{Gua09,Gua10}, by a data-driven feature extraction~\cite{Nie02,Tui12a}, or both~\cite{Mat13b}. In all cases, however, the aim is to find a common data representation, in which the data are more similar (or \emph{aligned}) to each other~\cite{Wan11}, independently from the processing steps that will follow. Since the data representation itself is modified, it is then possible to use any classifier on the aligned data. 

This type of adaptation can be seen in a global or local perspective: in the first case, the PDFs of all images are compared and aligned, either by matching the histograms~\cite{Hon05,Yan07} or by projecting the data distributions onto a common space using, for instance, Principal Component Analysis (PCA,~\cite{Mat13}), Canonical Correlation Analysis (CCA~\cite{Nie02}) or other more advanced nonlinear kernel multivariate methods~\cite{Jero13spm}. 
In the case of local matching, the point clouds are matched regionally, in order to account for deformations that affect only certain parts of the image manifold. Local methods generally either minimize a cost function bringing the nodes of two graphs closer while maintaining the local neighborhoods in the projected space unchanged~\cite{Sha85,Bun99,Mca11} or consider mixtures of local models~\cite{Kam97}, which are then merged in a global representation~\cite{Row02,Ver02,Teh03,Bra03}. In the remote sensing literature, an application of the first of these two approaches is found in~\cite{Tui12a}, where an iterative method is used to match graphs of the same sensor under angular and temporal variations. 

A desirable alignment method must be capable of aligning images 1) under strong distortions, 2) of different sensors and 3) not co-registered. The methodologies so far proposed in remote sensing literature do not address all these issues simultaneously: PCA does not require co-registration, but fails in multisensor scenarios and under strong distortions. On the contrary, (K)CCA is naturally multisensor and works under strong distortions, but requires strict co-registration. Table~\ref{tab:comp} summarizes these properties for a series of alignment methodologies found in remote sensing.

In this paper, we tackle these problems simultaneously by proposing a linear, efficient method for feature extraction and dataset alignment issued from the manifold alignment literature~\cite{Ham05,Wan11}. The main idea of the method is to use labeled samples from both domains to bring the manifolds closer, while keeping their respective inherent structure unchanged using two proximity graphs built with unlabeled samples~\cite{Wan11b}. Therefore, the method includes constraints on the local manifold geometry in the aligned space. The alignment transformation is defined by a linear projection function that depends both on labeled and unlabeled samples, making the method semi-supervised. In the \blue{aligned} space, pixels from both domains can be used simultaneously and a classifier effective in all domains can be learned. We will refer to this method as \emph{semi-supervised manifold alignment} (SS--MA).

SS--MA is effective for large deformations, since it is not based on inter-graphs distances, as \blue{the methods} in~\cite{Tui12a,Yang11}. It does not require co-registration of the image sources as~\cite{Zur08,Nie02}, and is naturally multisource, as it can align images of different dimensionality, unlike standard domain adaptation algorithms~\cite{Bru10,Kim10}. \blue{Since the eigenvectors  are defined with a discriminative term and are sorted by their eigenvalues, the classifier can then be learned using only the first dimensions: this makes SS--MA an interesting solution also for dimensionality reduction.} The price to pay is the availability of some (typically few) labeled pixels in each domain.

The remainder of the paper is organized as follows. Section~\ref{sec:methodo} presents the proposed methodology and the manifold alignment algorithm. Section~\ref{sec:data} presents the VHR data used in the experiments, that are detailed and discussed in Section~\ref{sec:res}. We will illustrate the method in several problems: 2D toy datasets under different deformations, as well as multiangular adaptation for the same sensor, multitemporal and multisource image adaptation. Section~\ref{sec:concl} concludes the paper with some discussion and further work.

\section{Semisupervised manifold alignment}\label{sec:methodo}

\begin{figure}
\centering
\begin{tabular}{cc}
(a) Original data & (b) Manifolds \\
\includegraphics[width=0.21\textwidth]{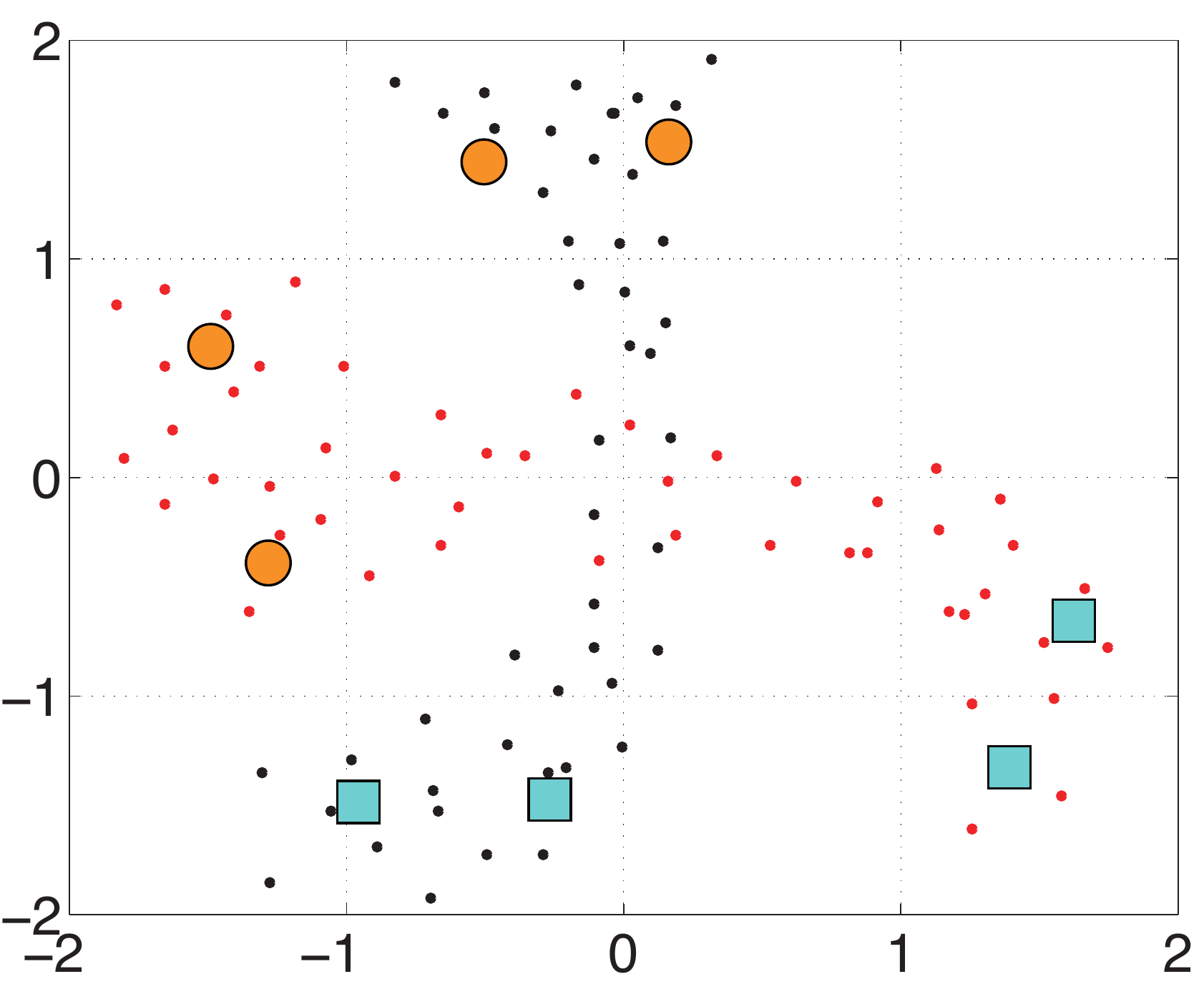}&
\includegraphics[width=0.21\textwidth]{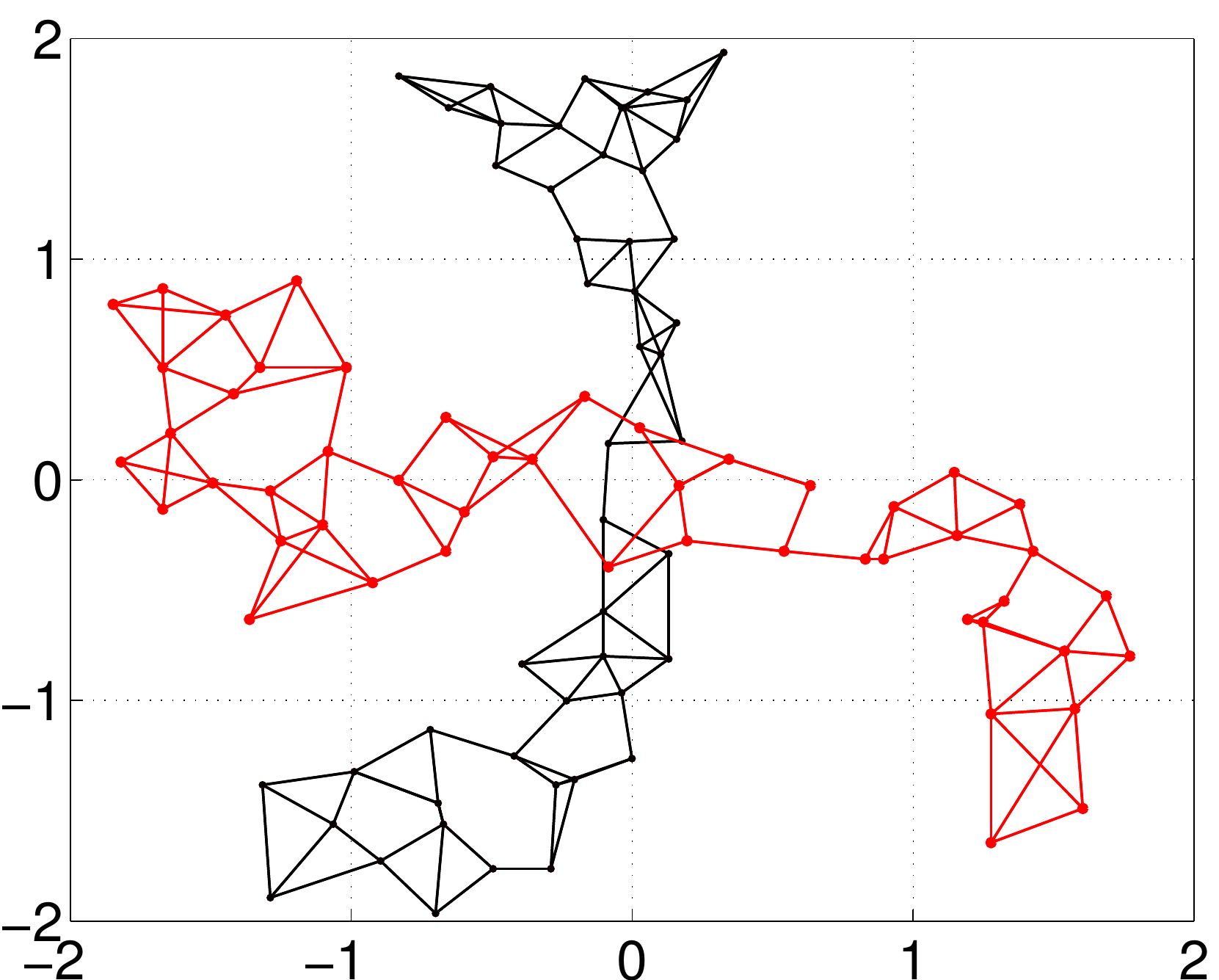}\\
(c) Class correspondence & (d) Aligned data \\
\includegraphics[width=0.21\textwidth]{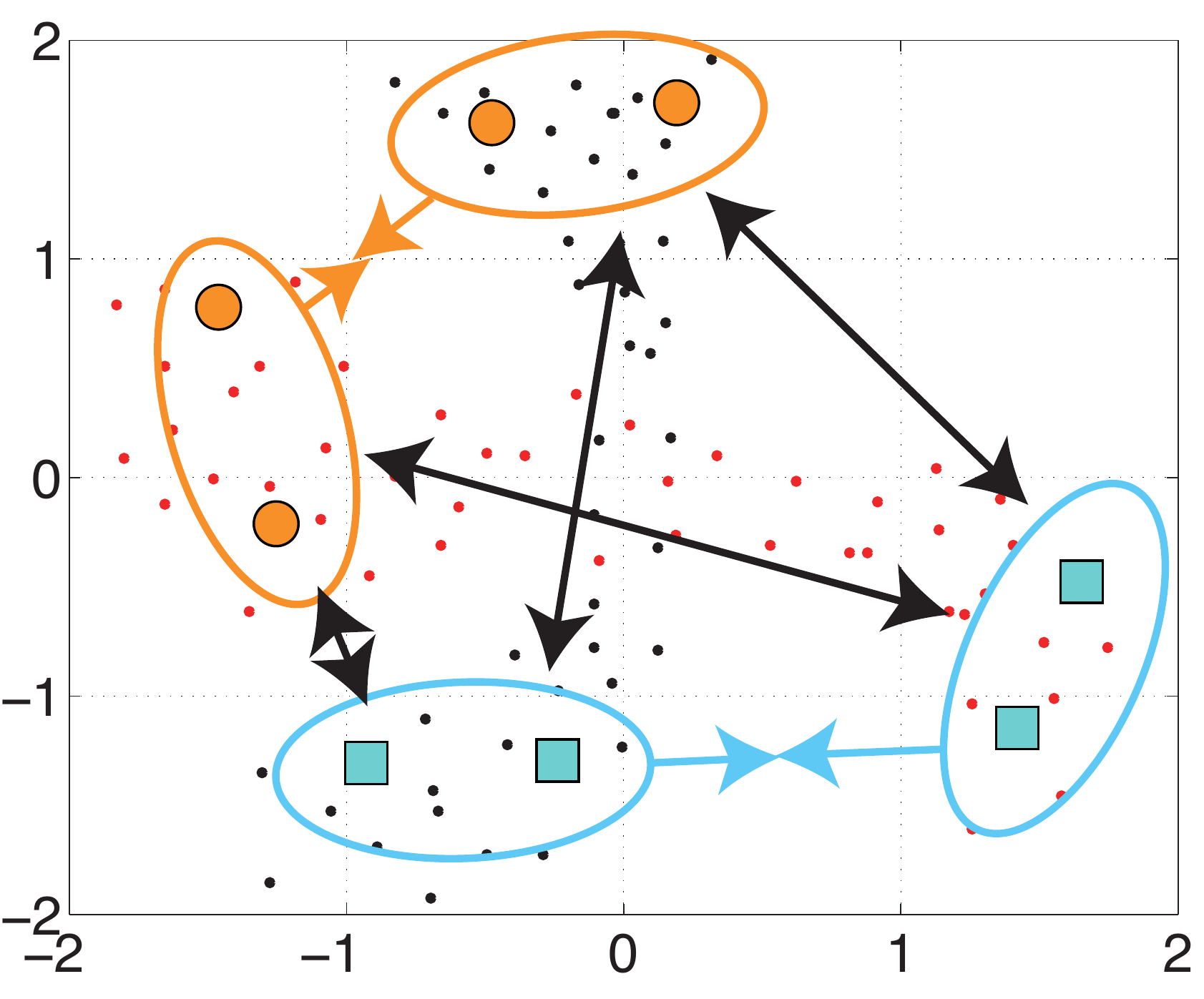}&
\includegraphics[width=0.21\textwidth]{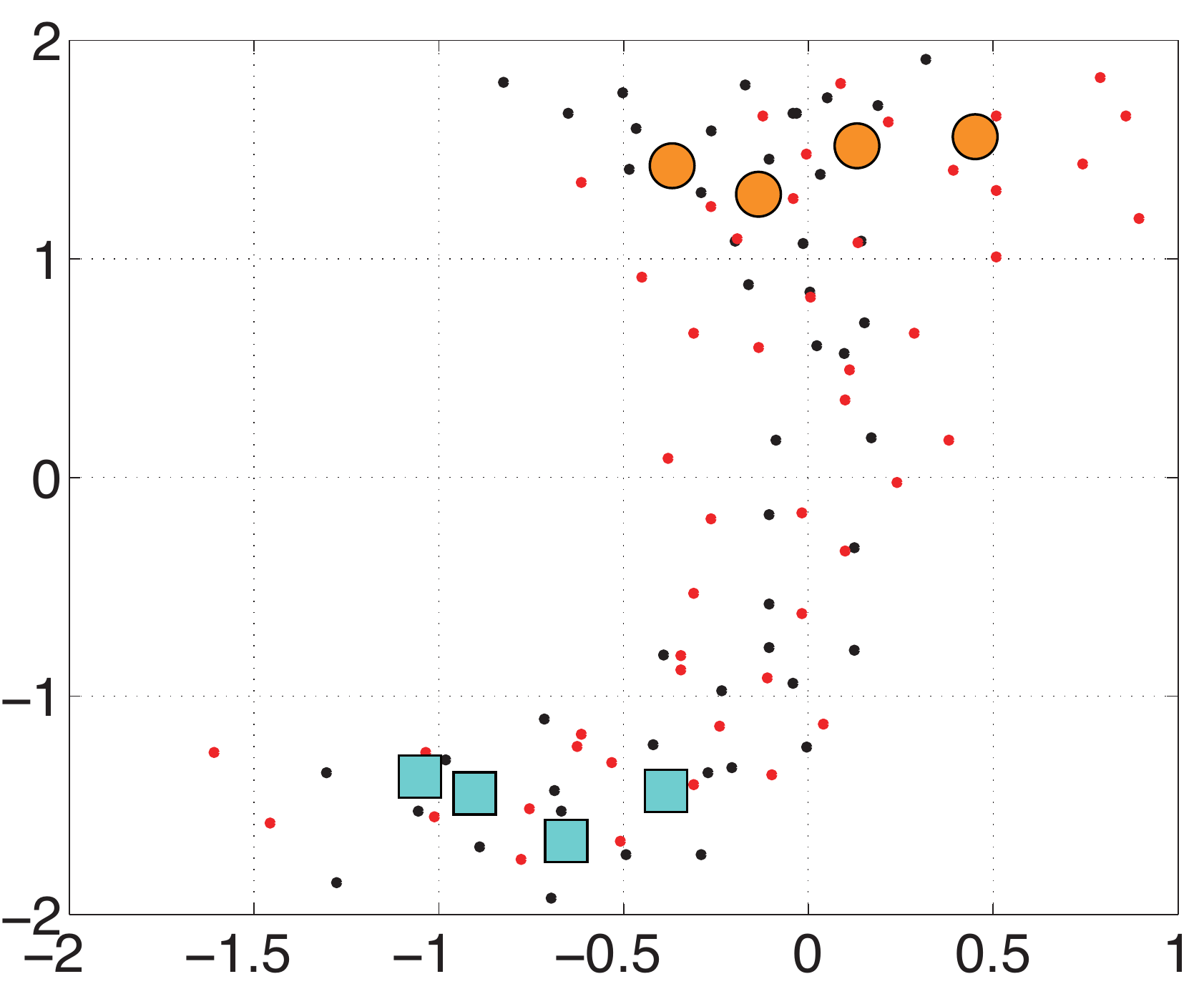}\\
\end{tabular}
\caption{Ingredients of the proposed algorithm: (a) two datasets ($\red{\bullet}$ and ${\bullet}$) distorted by a $90^\circ$ rotation and labeled pixels of two classes ($\orange{\bullet}$ and $\textcolor{SeaGreen}{\blacksquare}$); (b) Geometrical structures of each manifold to be preserved ($G$ term, Eq.~\eqref{g}); (c) Label similarity induced attraction (orange and cyan arrows, $S$ term, Eq.~\eqref{s}) and repulsion (black arrows, $D$ term, Eq.~\eqref{d}); (d) aligned dataset.}
\label{fig:ma}

\end{figure}

The main idea of the proposed method is to align the data manifolds by projecting them into a common representation, or \emph{joint latent space}, ${\mathcal F}$. Such space has two desirable properties: 1) it preserves the local geometry of each dataset and 2) it brings regions belonging to the same class closer together, while pushing those belonging to different classes apart. The method searches for a set of projection functions, one {\em per} image, that achieve this double objective~\cite{Wan11b}. Figure~\ref{fig:ma} illustrates the \blue{two properties and the expected alignment}.

\subsection{Notation}

Consider a series of $M$ images (or \emph{domains}) and their corresponding data matrices $\X^m$, $m=1,\ldots,M$. Each matrix $\X^m$ contains a large set of unlabeled samples  $\{\x_i^m\}_{i = 1}^{u_m}$, and some input-output labeled sample pairs $\{\x_j^m,y_j^m\}_{j=1}^{l_m}$, typically with $l_m \ll u_m$. To fix notation, examples are column vectors and the data matrices are $\X^m\in\mathbb{R}^{d_m\times n_m}$ where $n_m=l_m+u_m$. These matrices   contain both labeled and unlabeled points. Let us additionally denote the block diagonal data matrix containing all samples $\X^m$, $m=1,\ldots,M$, as $\X = diag(\X_1,\ldots,\X_M)\in\mathbb{R}^{d\times N}$, $N=\sum_m n_m$, $d = \sum_m d_m$. It is important to stress here that the images do not necessarily represent the same location and can be acquired by different sensors with different spatial and spectral resolutions, so in principle we may have different numbers of samples per data matrix, $n_m\neq n_{m'}$, and different dimensions \blue{in each domain}, i.e. $\x^m \in \mathbb{R}^{d_m}$ possibly with $d_m \neq d_{m'}$, $\forall m, m'=1,\ldots,M$. 

\subsection{Semi-supervised loss function}

The problem of aligning the $M$-images dataset to a common representation boils down to constructing $M$ mapping functions to ${\mathcal F}$ by means of $M$ projection matrices, $\f^m\in\mathbb{R}^{d_m\times d}$, $m=1,\ldots,M$. The common latent space ${\mathcal F}$ is of dimension $d = \sum_{m=1}^M d_{m}$. Mapping to ${\mathcal F}$  requires that samples belonging to the same class become closer, while those of different classes are pushed far apart. Moreover, the mapping should preserve the geometry of each data manifold. Since we have terms to be minimized and others to be maximized, \blue{the problem reduces to solving a standard Rayleigh quotient}\footnote{\blue{Also known as the Rayleigh--Ritz ratio, and widely used in the field of Fisher's discriminant analysis.}}:
\begin{equation}
\F^{opt} = \blueRR{}{\arg} \min_{\F} \{\text{tr}( (\F^\top \B \F)^{-1} \F^\top \A\F ) \}
\label{costABC}
\end{equation}  
\blueRR{where $\F=[\f^1,\ldots,\f^M]^\top$ is a \blue{projection} matrix of size $d\times d$}{where $\F$ is a $d \times d$ projection matrix, whose row-blocks correspond to the $d_m \times d$ domain-specific projection functions, $\f^m$}.  $\A$ \blue{is a matrix that} contains the quantity of data relations to be \blueRR{increased}{decreased} by the projection, and  $\B$ \blue{is a matrix that} contains the quantity of data relations to be \blueRR{decreased}{increased}. 
\blueR{In our case, we want find the set of projectors $\F^{opt}$ that maximizes the distances between samples of different classes\blue{, represented by an affinity matrix} $\D$ (thus $\B = \D$) and at the same time minimizes the distances between samples that are either close in each manifold (\blue{affinity matrix} $\G$) or of the same class (\blue{affinity matrix} $\S$).} These two last \blue{matrices} are combined as $\A = \mu \G + \S$, where $\mu$ is a tradeoff parameter. $\A$ favors discriminant projections (induced by $\S$) and at the same time \blue{preserves} the original geometry of each manifold (\blue{summarized in} $\G$). The effect of the three terms $\G, \S$ and $\D$ is illustrated in Fig.~\ref{fig:ma}. \blue{The three matrices $\D$, $\S$ and $\G$ will be approached by graphs Laplacians, as detailed below.} 

Let us now describe the computation of the corresponding scalar terms $G$, $S$, and $D$\blue{, which can be interpreted as the global quantities to be minimized (for $\S$ and $\G$), respectively maximized (for $\D$), by the optimization}.

The term $G$ ensures that the geometry of each manifold is preserved in the transform. For this reason it does not act among the sources, but only within each source. To this end, local similarity matrices $\mathbf{W}_g^{m}\in\mathbb{R}^{n_m\times n_m}$ need to be defined. In this paper, similarity matrices are built using standard graphs, such as the $k$NN or $\epsilon$-ball~\cite{Cha06b}\blueRR{}{, computed for each domain $m$ separately. Entries of the matrices are $W_g^{m}(i,j) = 1$ if samples $\x_i$ and $\x_j$ from domain $m$ are neighbors in the $k$NN graph of that domain and $0$ otherwise. }
The term ensures that neighboring samples in the original domains are mapped close in the projected space\footnote{The term acts as a form of graph regularization \blue{enforcing} the smoothness (or manifold) assumption. Intuitively, this is equivalent to penalize ``rapid changes'' of the projection function $\f^{m}$ evaluated between samples \blue{that are close} in the graphs \blue{$\mathbf{W}_g^m$.}}. \blueRR{}{No geometric relation between domains are considered, the geometry is preserved solely within each graph.} 
The geometry term reduces to: 

\begin{align}\label{g}
G &= \sum_{m=1}^M \sum_{i,j=1}^{{n_{m}}} W_g^{m}(i,j) {\|{\f^{m}}^\top{\x_i^m} - {\f^{m}}^\top{\x_j^{m}} \|^2}\nonumber\\ &  = \text{tr}(\F^\top\X\L\blueRR{}{_g}\X^\top\F)
\end{align}
where $\f^m$ are the functions projecting the domain $m$ in the latent space ${\mathcal F}$ \blueRR{}{and $\L\blueRR{}{_g}=diag(\L\blueRR{}{_g^1},\ldots,\L\blueRR{}{_g^M})$ is a block-diagonal matrix containing the}  \blueRR{}{ domain-specific graph Laplacians,  reflecting the geometric similarity in each domain. For domain $m$, the corresponding Laplacian is  $\L\blueRR{}{_g^m}=\mathbf{U}\blueRR{}{_g^m}-\W\blueRR{}{_g^m}$, with  degree matrix  $\mathbf{U}\blueRR{}{_g^m}(i,i) = \sum_{j}\mathbf{W}\blueRR{}{_g^m}(i,j)$.}

The term $S$ enhances class similarity between the labeled instances of all domains. Its role is to pull labeled samples of the same class close together. To this end, we use a matrix of class-similarities, 
$\W_s = [\W_s^{(m,m')}]$, $m,m'=1,\ldots, M$, 

where the components of each $\W_s^{m,m'}\in\mathbb{R}^{n_m\times n_m}$ are $W_s^{m,m'}(i,j)=1$ for samples with the same label and $0$ otherwise (including unlabeled data). Minimizing $S$ for all the image sources simultaneously corresponds to minimize the distance among all samples of the same class in the latent~space:

\begin{align}\label{s}
S &= \sum_{m,m'=1}^M\sum_{i,j=1}^{l_{m},l_{m'}} W_s^{m,m'}(i,j) {\| {\f^{m\top}}\x_i^m - {\f^{m'\top}}\x_j^{m'} \|^2} \nonumber\\ & = \text{tr}(\F^\top\X\L_s\X^\top\F),
\end{align}
where $\L_s$ is the corresponding joint graph Laplacian to $\W_s$. 
What we want to achieve is to map all samples of the same class close to each other, independently from the source they come from. Note that this term  considers relations of labeled samples in different domains.

The dissimilarity term $D$ encodes the opposite behavior. By maximizing $D$, one tries to pull samples belonging to different classes apart from each other. 
To this end, we employ a matrix of class-dissimilarities, 
$\W_d = [\W_d^{(m,m')}]$ 

where the components of each $\W_s^{m,m'}\in\mathbb{R}^{n_m\times n_m}$ are $W_s^{m,m'}(i,j)=1$ for samples of different classes and 0 otherwise (including unlabeled data). The dissimilarity term is:
\begin{align}\label{d}
D &= \sum_{m,m'=1}^M\sum_{i,j=1}^{l_{m},l_{m'}} W_d^{m,m'}(i,j) {\| {\f^{m\top}}\x_i^m - {\f^{m'\top}}\x_j^{m'} \|^2} \nonumber\\ & = \text{tr}(\F^\top\X\L_d\X^\top\F),
\end{align}
where $\L_d$ is the corresponding joint graph Laplacian for $\W_d$. The term is maximized when labeled samples of different classes are mapped far from each other in the latent space.

\subsection{Projection functions}
All the $\W$ matrices \blueRR{}{($\W_g$, $\W_s$ and $\W_d$)} and corresponding graph Laplacians are of size $N \times N$. Now, by plugging Eqs.~\eqref{g}, \eqref{s} and \eqref{d} into~\eqref{costABC}, we obtain:
\begin{equation}
\min_{\F} \{\text{tr}( (\F^\top \X \L_d\X^\top \F)^{-1} \F^\top \X (\mu\L\blueRR{}{_g}+\L_s)\X^\top \F ) \}.
\label{costABC2}
\end{equation}
The solution of the minimization problem in Eq.~\eqref{costABC} is given by the eigenvectors $\boldsymbol{\varphi}_i$ corresponding to the smallest eigenvalues of the following generalized eigenvalue decomposition~\cite{Wan11b}: 
\begin{equation}
\X(\mu \L\blueRR{}{_g} + \L_s)\X^\top \boldsymbol{\varphi} = \lambda \X \L_d \X^\top \boldsymbol{\varphi}.
\end{equation}
The optimal matrix $\F^{opt}$ contains the projectors from the original spaces to the joint latent space in row blocks: 
\begin{equation}
\F^{opt} = \Big[\sqrt{\lambda_1}\boldsymbol{\varphi}_1  | \ldots |\sqrt{\lambda_d}\boldsymbol{\varphi}_d \Big] = 
	\begin{bmatrix} 
		\f^1_{1} & \ldots & \f^1_{d} \\
		\f^2_{1} & \ldots & \f^2_{d} \\
      \vdots & \vdots & \vdots   \\
		\f^M_{1} & \ldots & \f^M_{d}
	\end{bmatrix} 
\label{eq:gamma}
\end{equation}

By looking at the structure of the $\F^{opt}$ matrix, we observe that each row block contains the projection function from domain $m$ to the latent space. Each domain can thus be projected to the joint space by simple matrix multiplication (as in any multivariate analysis method like PCA). Projecting a data matrix in domain $m$, $\X_\ast^m$ to the $d$-dimensional latent space ${\mathcal F}$  just involves:
$${\mathcal P}_f(\X_\ast^{m}) = {{\f^m}^\top{\X_\ast^m}}.$$ 

\subsection{Properties}

The method has the following properties:
\begin{itemize} 
\item[a)] \emph{Linearity}: the method defines explicitly the projection functions, which can in turn be used to project large datasets into the latent space explicitly optimized for joint classification of all the sources.
\item[b)] \emph{Multisensor}: since it only exploits the geometry of each manifold separately, there is no restriction on the number of bands to be aligned nor \blue{on} their properties.
\item[c)] \emph{Multidomain}: as for CCA, the method can align an arbitrary number of domains in a common latent space. It does not necessarily require (as in the experiments below) a leading source domain to which all the others are aligned to (contrarily to PCA, TCA 
 and graph matching~\cite{Tui12a}).
\item[d)] \emph{PDF-based}: the method aligns directly the PDFs of the sources, without requiring co-registered samples. Unlike CCA, it can thus align images, which are unregistered, of different areas or of different spatial resolutions.
\item[e)] \emph{Invertibility}: using the projection functions defined, it is possible to synthesize the bands of one sensor from images acquired with another one. This opens many possibilities for the design and evaluation of new instruments.
\end{itemize}
These properties are also summarized and compared to other feature extractors used to align domains in Table~\ref{tab:comp}. 

\begin{table}[!t]
\caption{Properties of the data representation methods.}
\label{tab:comp}
\vspace{-0.25cm}
\setlength{\tabcolsep}{2pt}
\centering
\begin{tabular}{p{3.6cm}|p{2.3cm}|p{2.4cm}|p{2.3cm}|p{2.2cm}|p{2cm}}
\hline
Method & No labeled pixels in targets & Multisensor & No Co-registration & More than 2 images simult. & Invertible \\
\hline\hline
PCA & $\checkmark$ & $\times$ & $\checkmark$ & $\times$ & $\checkmark$  \ \\\hline
k-PCA & $\checkmark$ & $\times$ & $\checkmark$ & $\times$ & $\times$  \ \\\hline
TCA & $\checkmark$ & $\times$ & $\checkmark$ & $\times$ & $\times$   \\\hline
CCA & $\checkmark$ & $\checkmark$ & $\times$ & $\checkmark$ & $\checkmark$ \\\hline
k-CCA & $\checkmark$ & $\checkmark$ & $\times$ & $\checkmark$ & $\times$ \\\hline
Graph matching~\cite{Tui12a} & $\checkmark$ & $\times$ & $\checkmark$ & $\times$ & $\checkmark$ \\\hline\hline
SS--MA & $\times$ & $\checkmark$ & $\checkmark$ & $\checkmark$ &$\checkmark$  \\\hline
\end{tabular}
\end{table}

\subsection{Computational complexity}

\blue{
The proposed method reduces to solving a generalized eigenvalue problem of relatively small size, involving $d\times d$ matrices. The problem is  linear with the number of data sources (in our case, images). Most of the computational effort is associated to the construction of the graph Laplacian $\L\blueRR{}{_g}$ involved in the geometry-preserving term $G$, because the two other graph Laplacians $\L_s$ and $\L_d$ are constructed with a simple convolution between the vectors of labels, and hence the cost of their construction is negligible. Dealing with graph Laplacians with many vertices and edges can be computationally demanding for two main reasons: the calculation/storage of the matrix itself and the cost of operations such as eigendecomposition and inversion when dealing with these large matrices. }

\blue{
The memory requirements cannot be solved easily, but  strategies to decrease the size of the matrices exist, such as intelligent node selection via landmark points~\cite{ChenCrawfordGhosh06}, active learning~\cite{Tuia11}, inclusion of additional constraints to regularize the graph spatially~\cite{Chi13} or graph partitioning and sparsification~\cite{SpielmanTeng08}. Efficient solvers to construct the graph Laplacian have  been recently proposed: the solver proposed in~\cite{LivneBrandt12} scales linearly with the number of edges in both run time and storage. Also note that in SS--MA the graph Laplacian is computed only once and its calculation is performed off-line.}

\blue{
The second problem (the cost of the eigendecomposition) is not critical for SS--MA: since Eq.~\eqref{costABC2}  decomposes an eigenproblem of relatively small size ($d \times d$), there is no need to design specific strategies to reduce the size of the problem (by sample/nodes selection) or to improve the efficiency of the algebraic process.  }
\blue{
Note however that iterative methods can be used to reduce the computational load of the eigendecomposition to a logarithmic complexity (which is otherwise $\mathcal{O}(d^3)$ for our $d \times d$ matrix): examples can be found in~\cite{Dhillon98,PanZhaoChen99}.}

\begin{figure*}[!t]
\centering
\begin{tabular}{ccccc}
$- 38.79^\circ$ &  $-29.16^\circ$ & $6.09^\circ$ & $26.76^\circ$ & $39.5^\circ$ \\
\includegraphics[width=3cm]{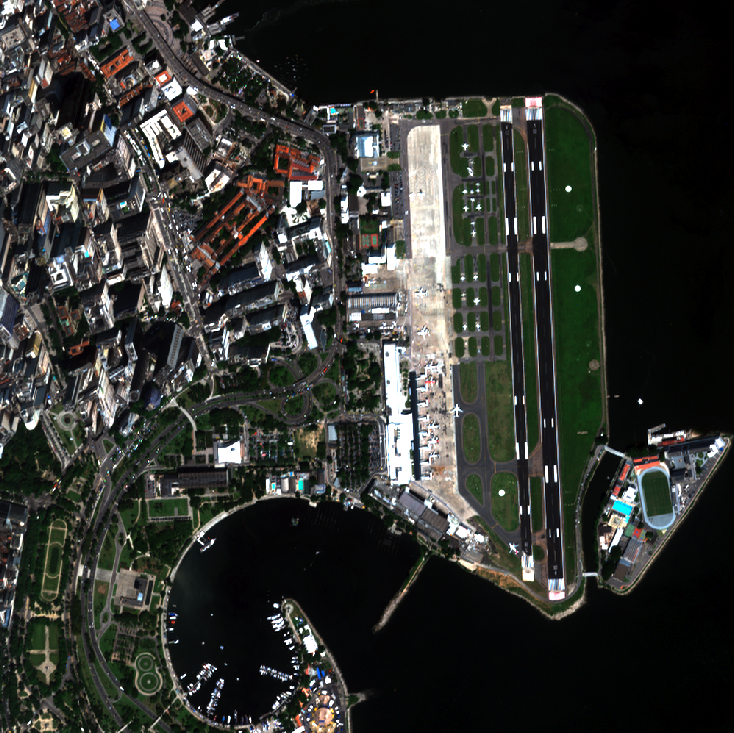}&
\includegraphics[width=3cm]{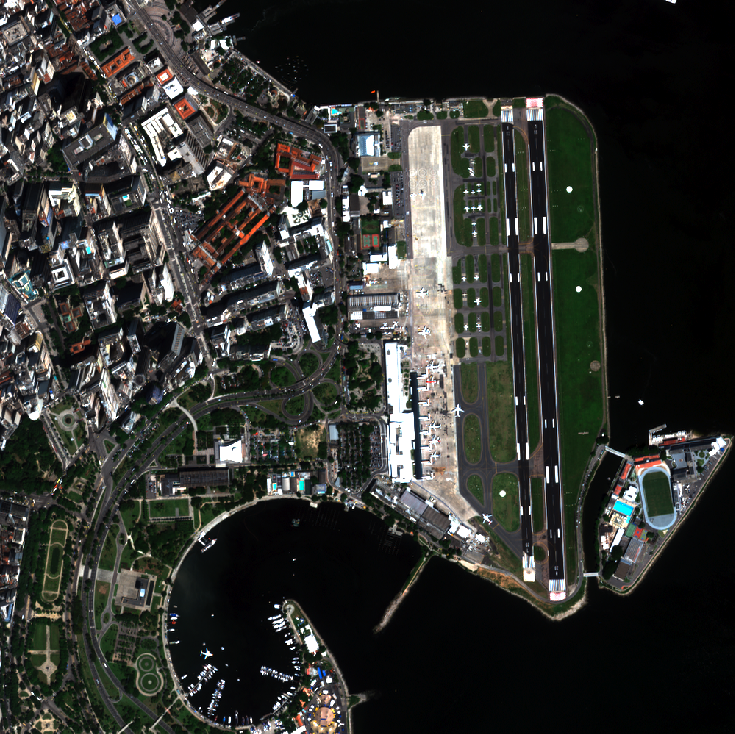}&
\includegraphics[width=3cm]{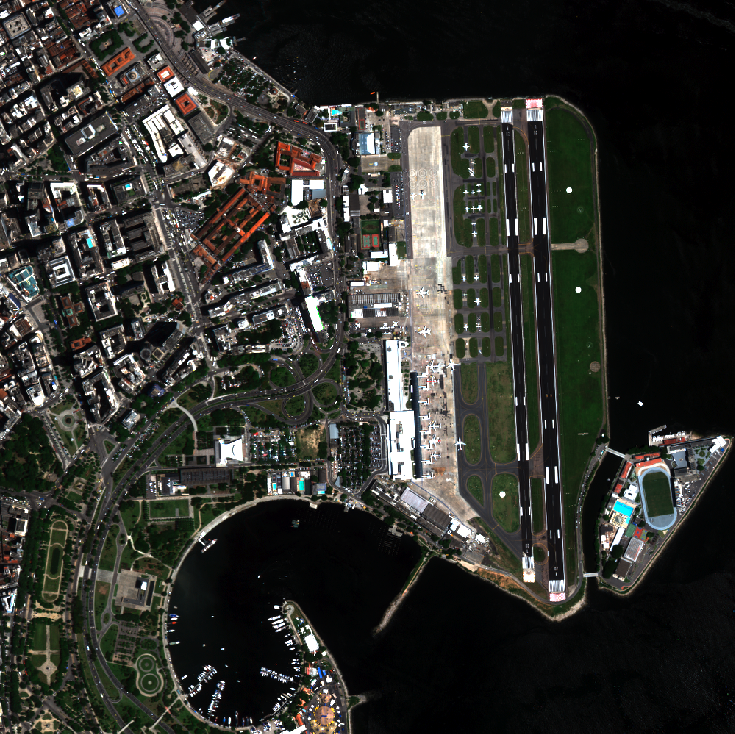}&
\includegraphics[width=3cm]{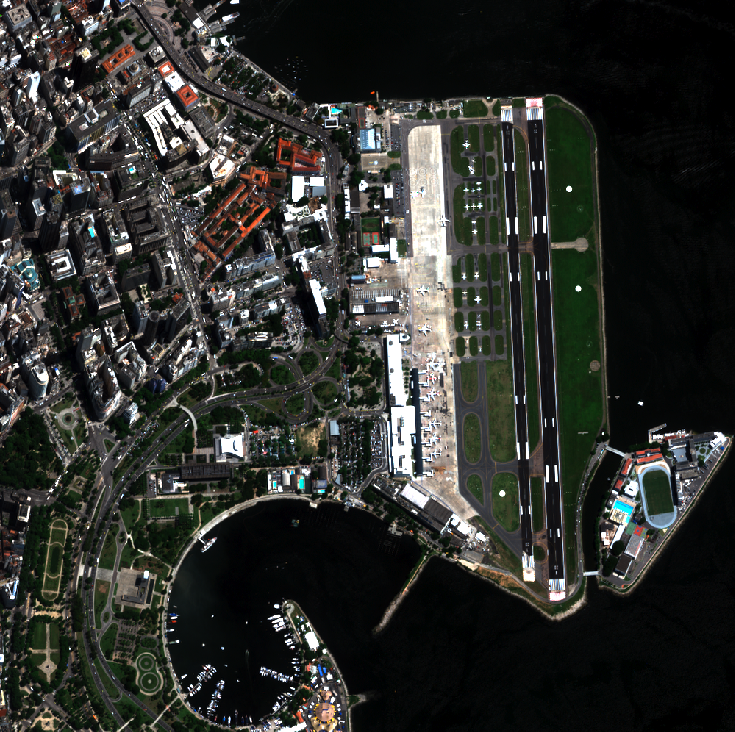}&
\includegraphics[width=3cm]{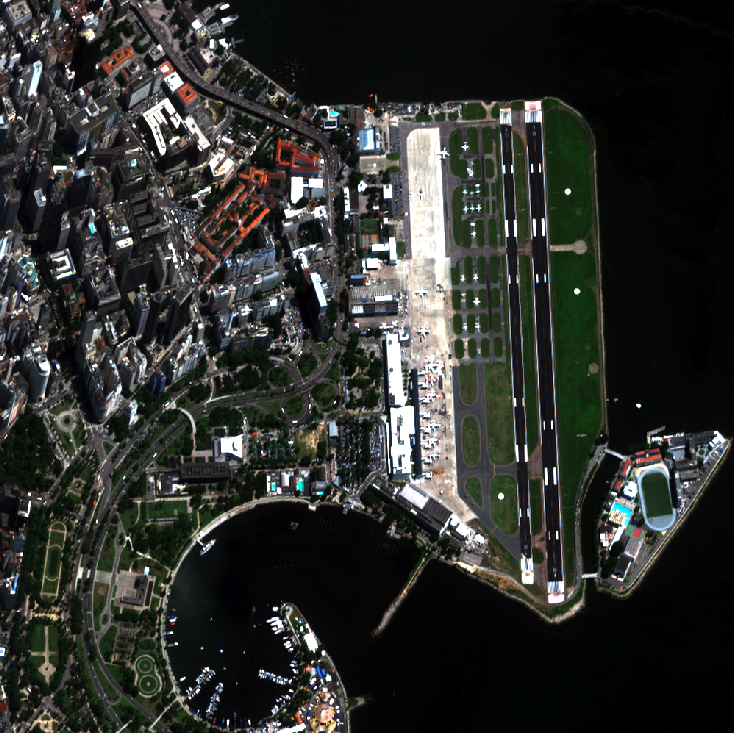}\\
\end{tabular}
\caption{The five images used in the multiangular experiments.}
\label{fig:ang}
\end{figure*}

\section{Data and setup}\label{sec:data}

This section describes the data and the setups used in the experiments presented in Section~\ref{sec:res}. 

\subsection{Datasets and evaluated scenarios}

\begin{figure}[!t]
\centering
\begin{tabular}{ccc}
&Images & Classes \\
\rotatebox{90}{ Prilly, WV2}&
\includegraphics[width=3.6cm]{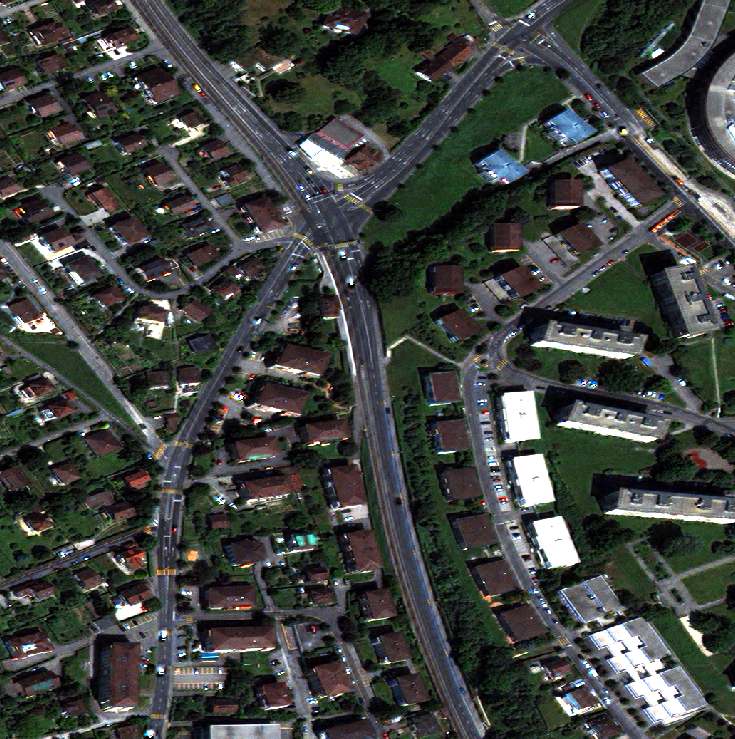}&
\includegraphics[width=3.6cm]{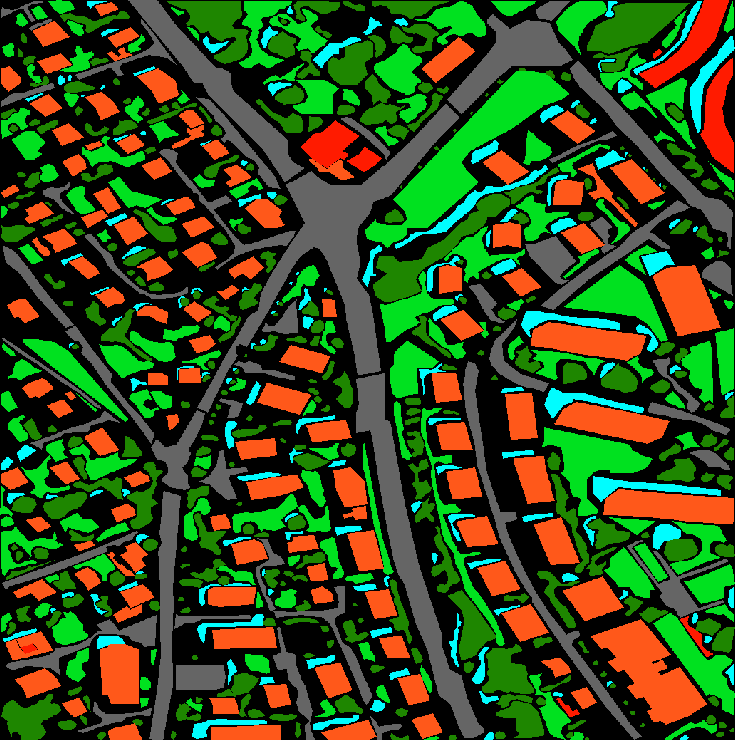}\\
\rotatebox{90}{ Montelly, WV2}&
\includegraphics[width=3.6cm]{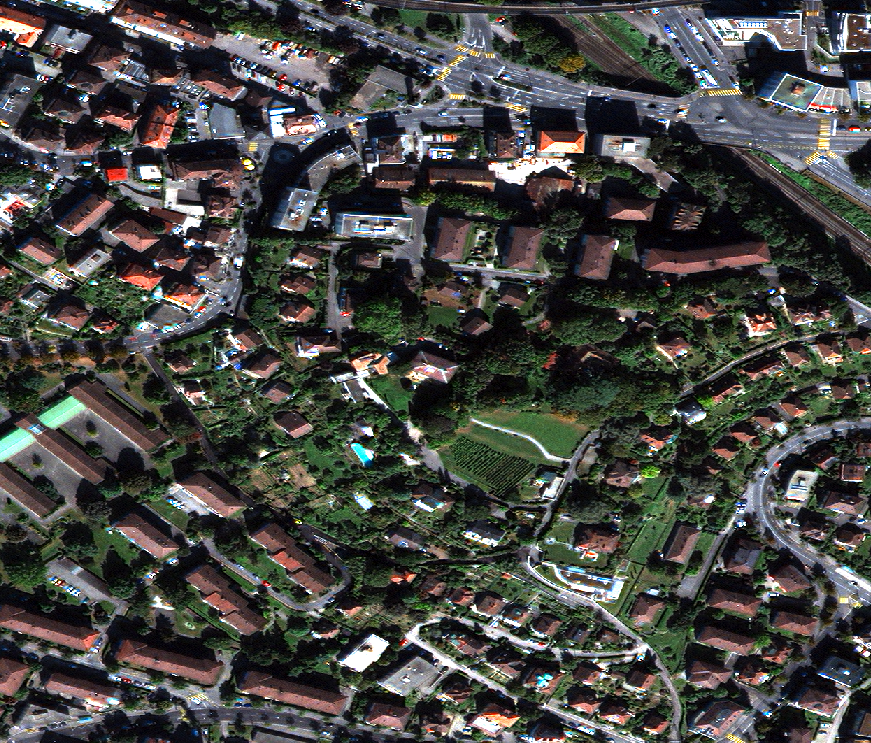}&
\includegraphics[width=3.6cm]{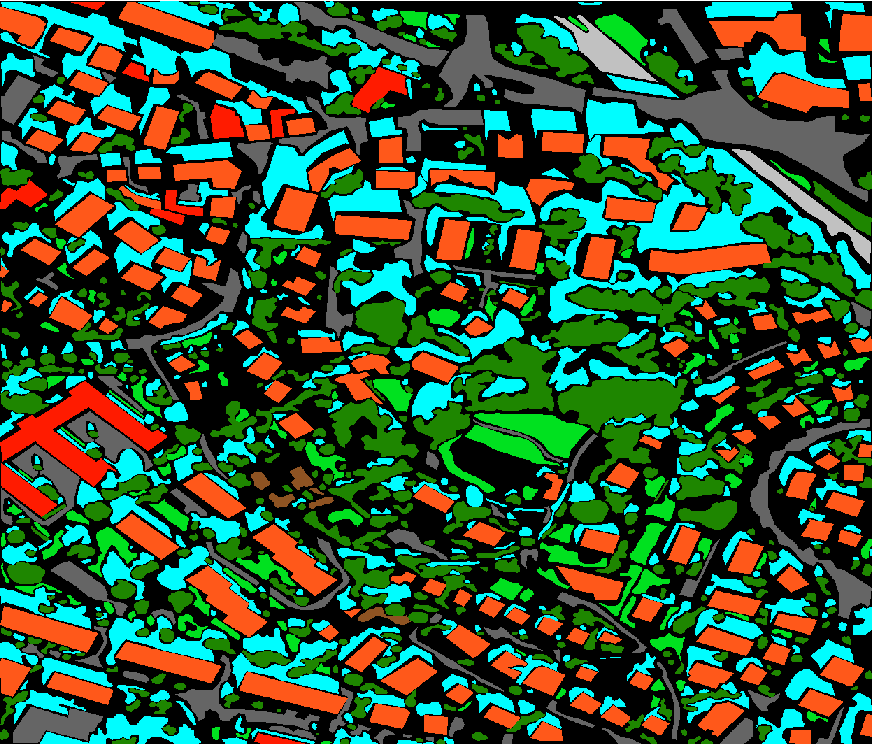}\\
\rotatebox{90}{Malley, WV2}&
\includegraphics[width=3.6cm]{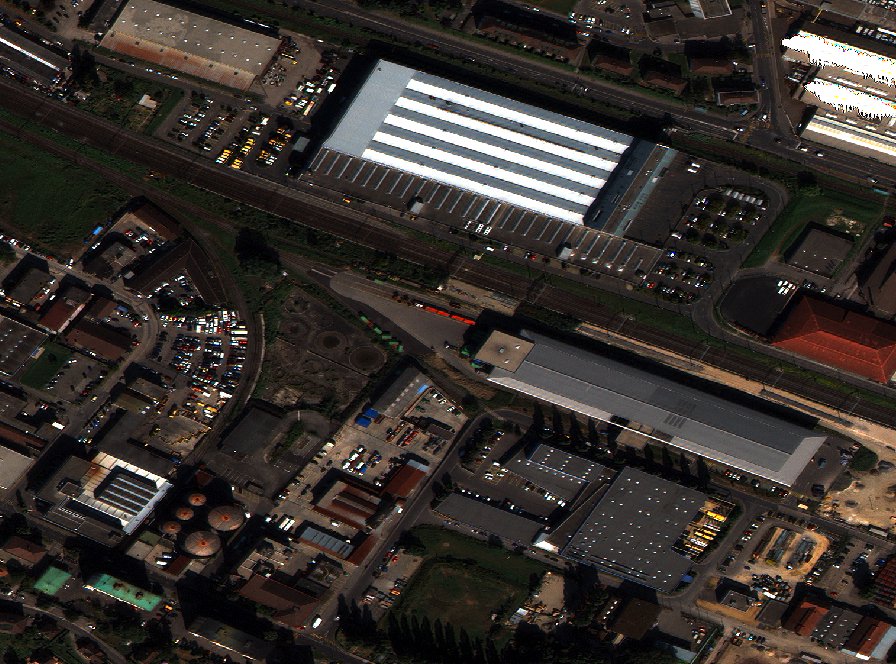}&
\includegraphics[width=3.6cm]{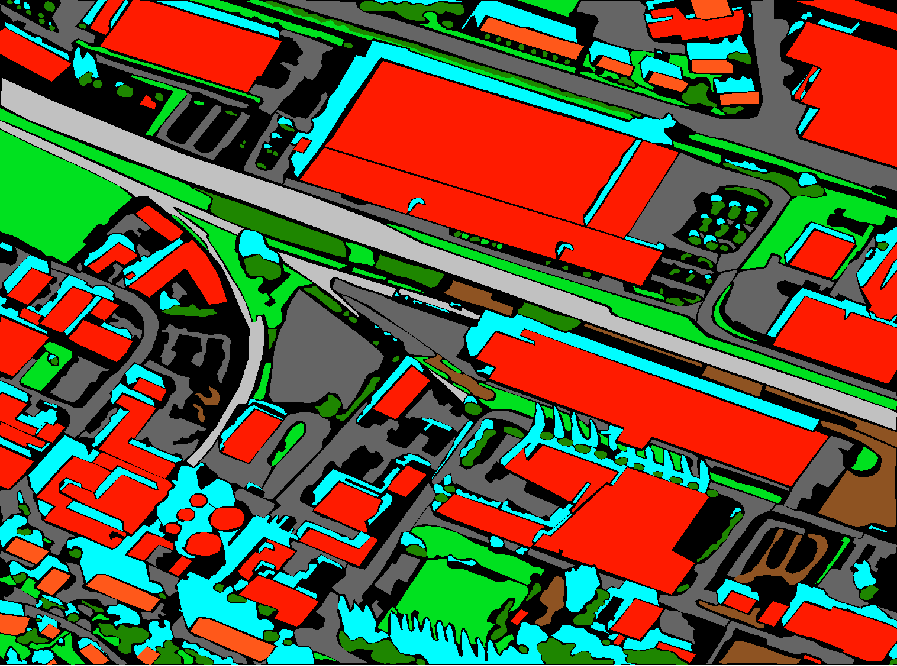}\\
\rotatebox{90}{Zurich, QB}&
\includegraphics[width=3.6cm]{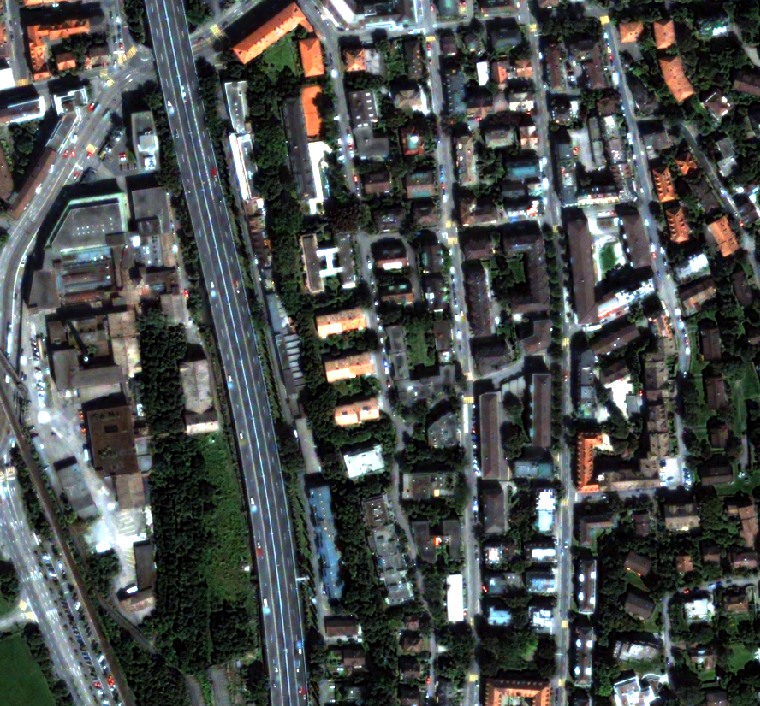}&
\includegraphics[width=3.6cm]{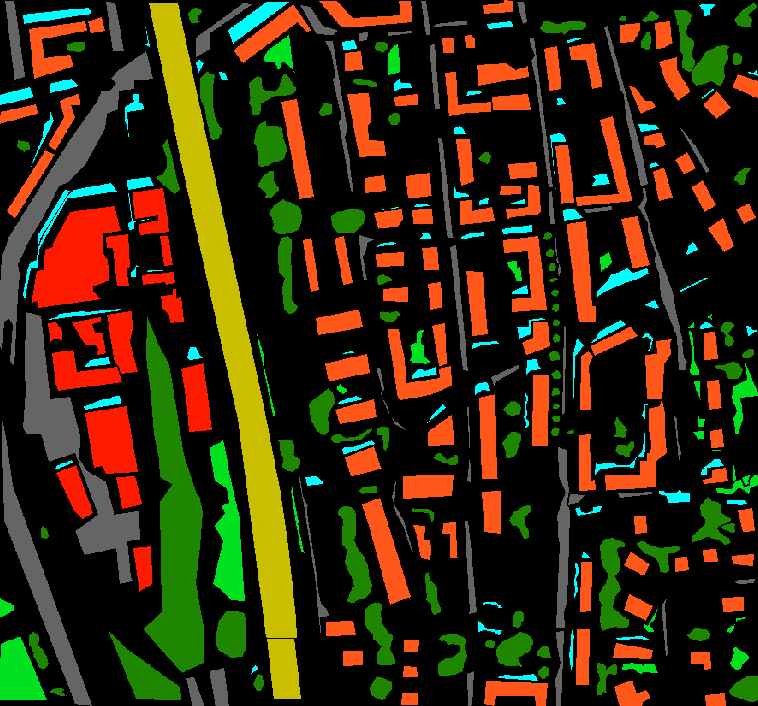}\\
\end{tabular}
\caption{The four images used in the multitemporal and multisource experiments.}
\label{fig:im}
\end{figure}

\begin{figure*}[!t]

\begin{tabular}{cccccc}
& (a) & (b) & (c) & (d) & (e)\\
& Domains  & 3 Classes & Alignment  & Predicting test samples   & Predicting test samples\\
&($\red{\bullet}$ vs. $\blue{\bullet}$ )&&SS--MA&from the first spiral ($\red{\bullet}$) &from the second spiral ($\blue{\bullet}$) \\
\rotatebox{90}{\hspace{0.5cm} S}&
\includegraphics[width=2.3cm]{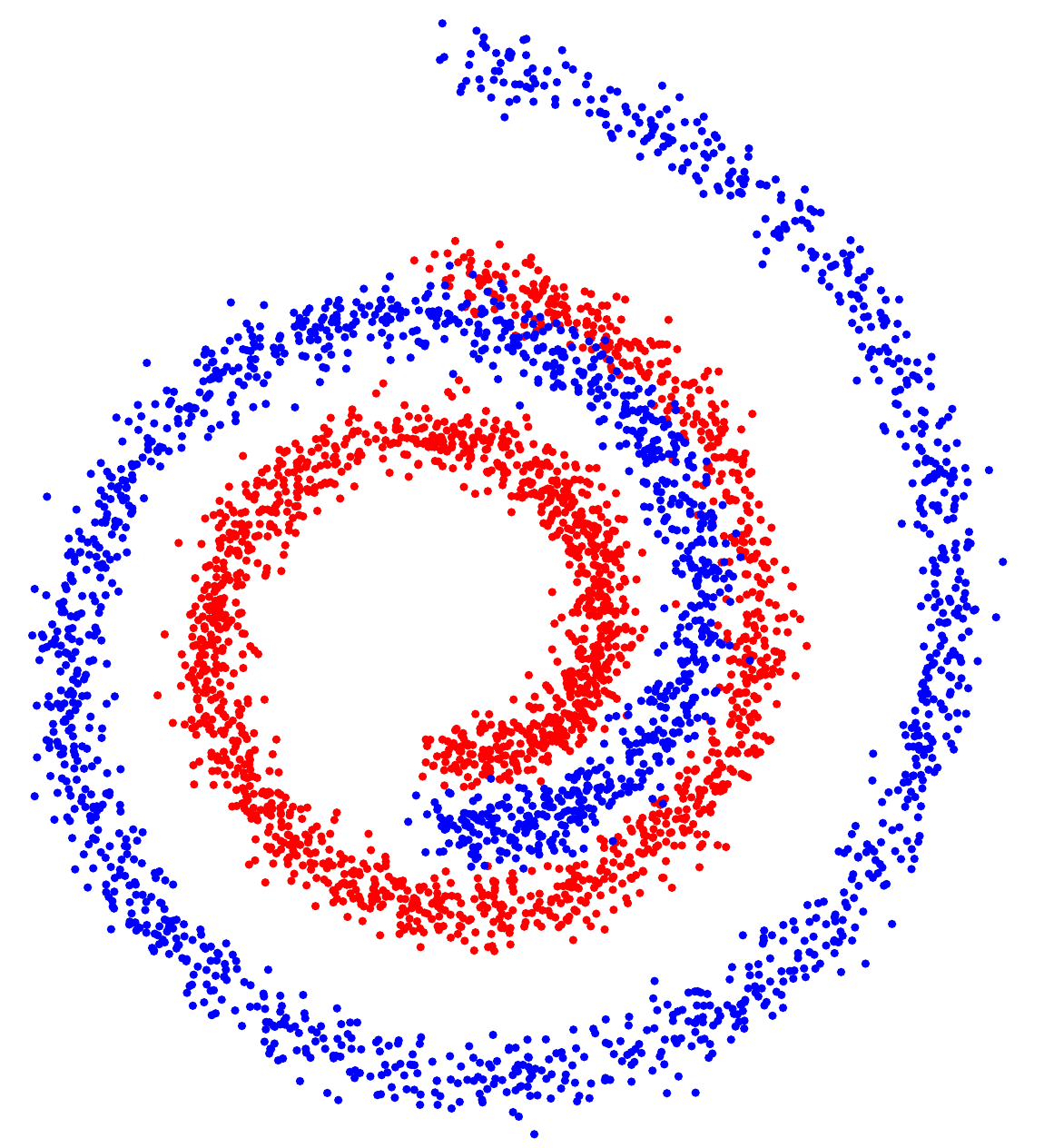}&
\includegraphics[width=2.3cm]{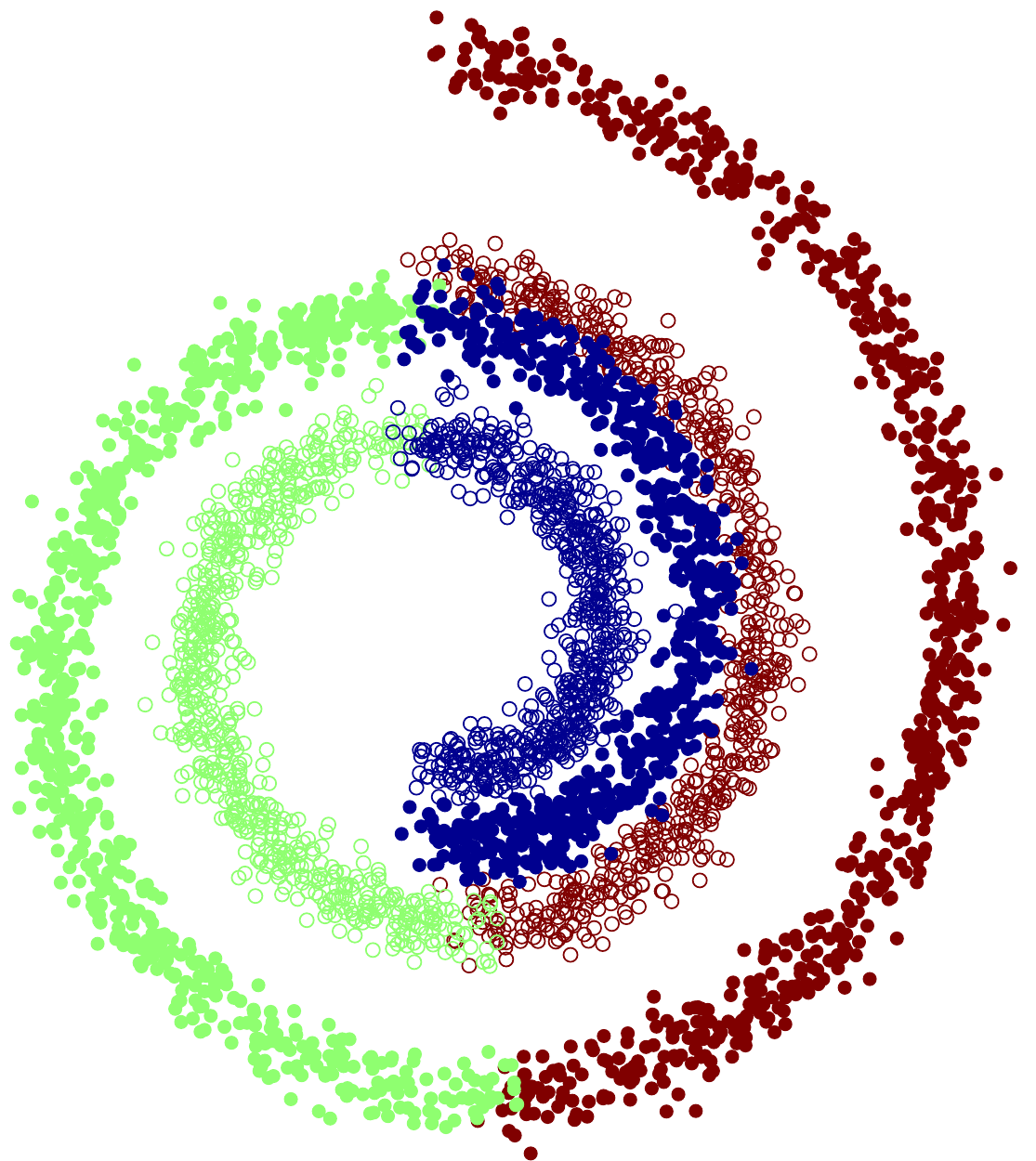}&
\includegraphics[width=2.3cm]{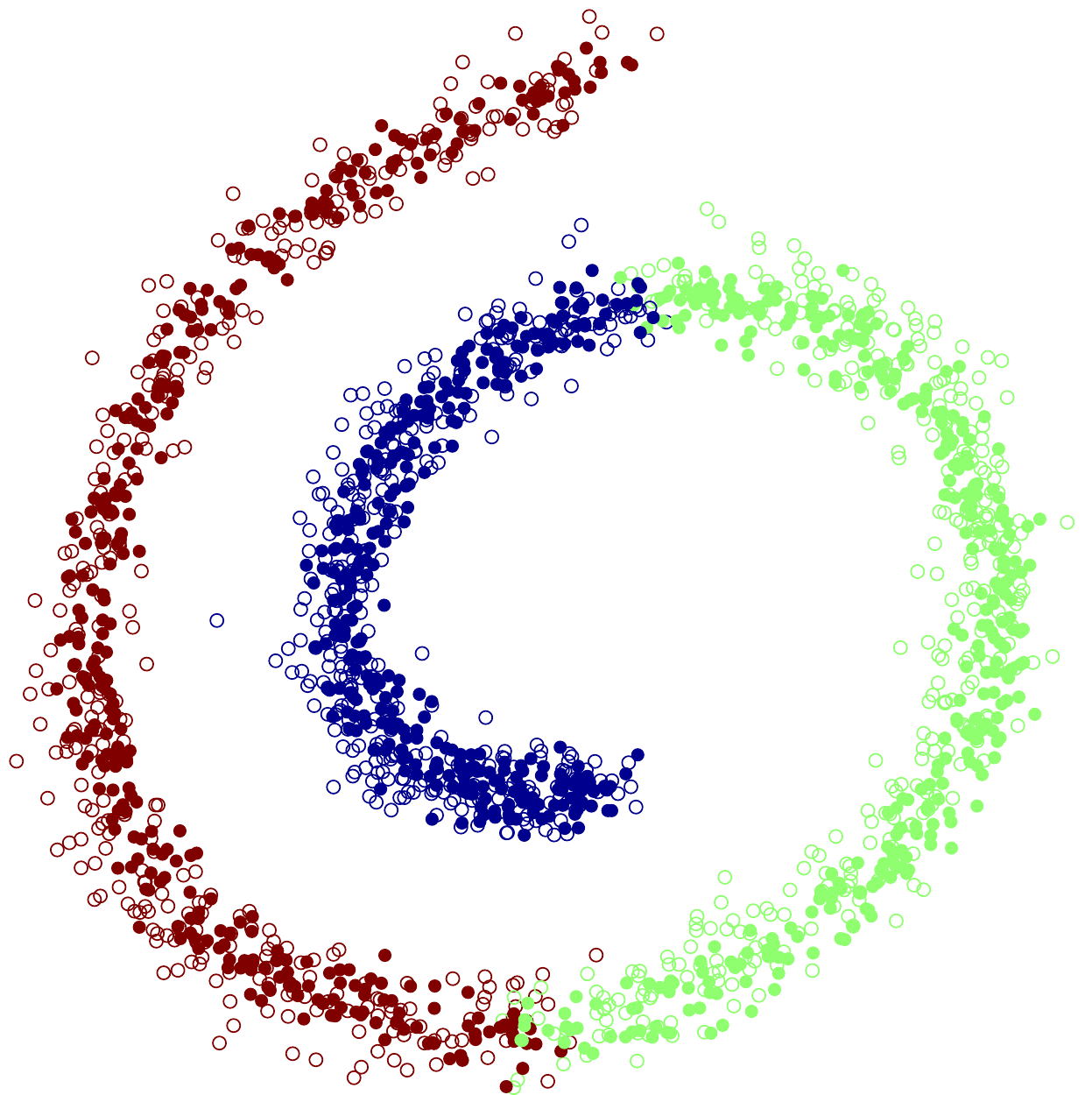}&
\includegraphics[width=3.5cm]{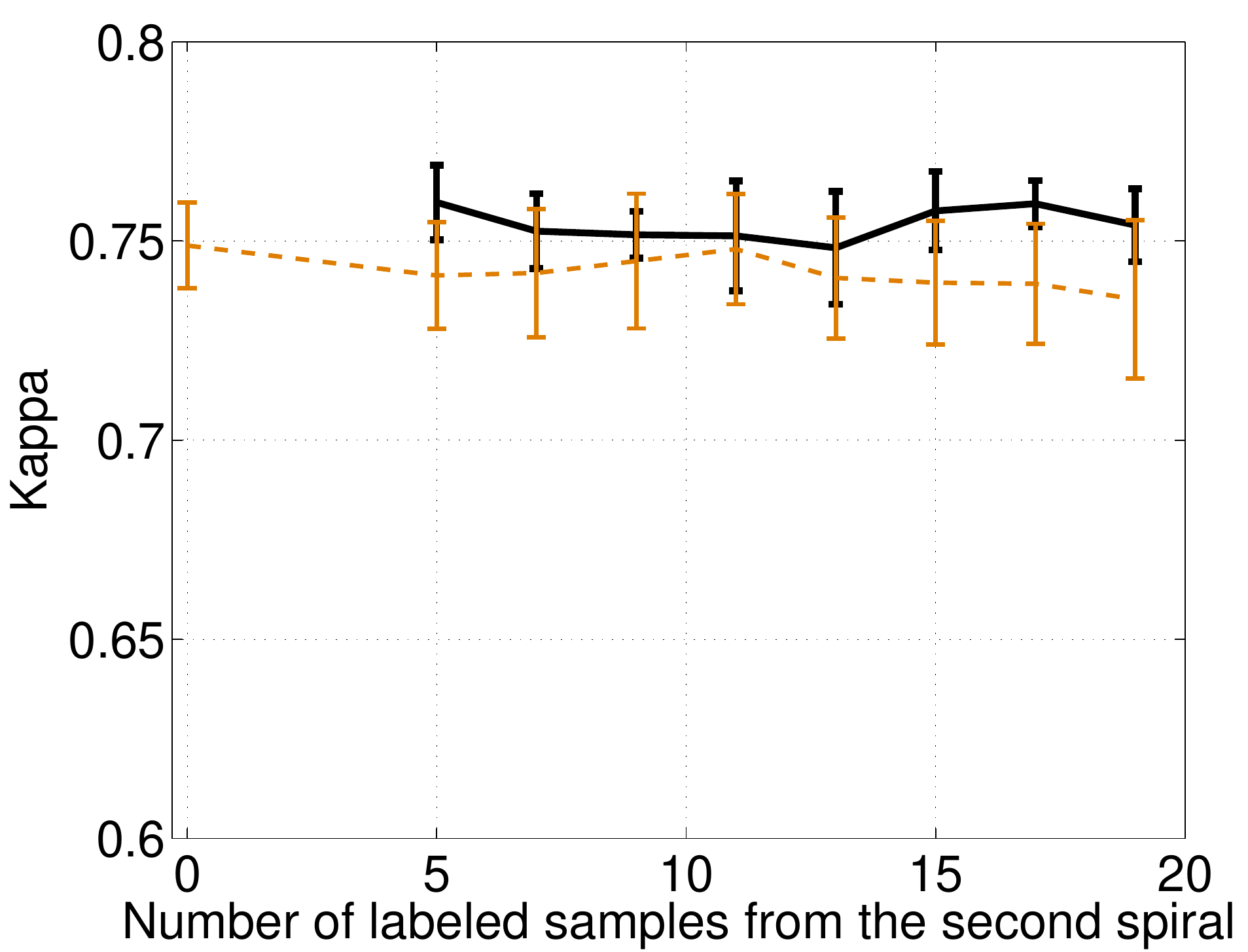}&
\includegraphics[width=3.5cm]{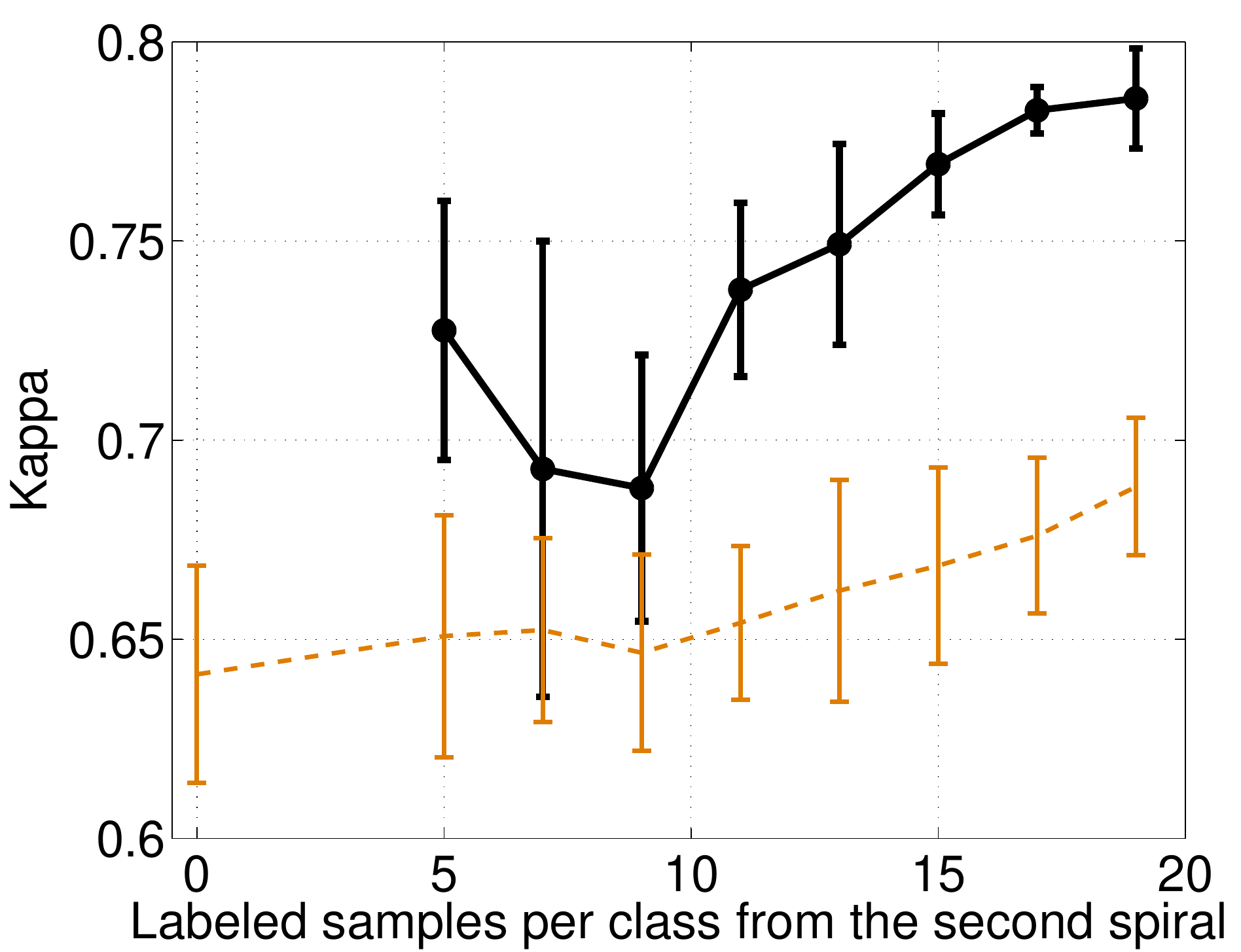}\\\hline

\rotatebox{90}{\hspace{0.5cm} S + R}&
\includegraphics[width=2.3cm]{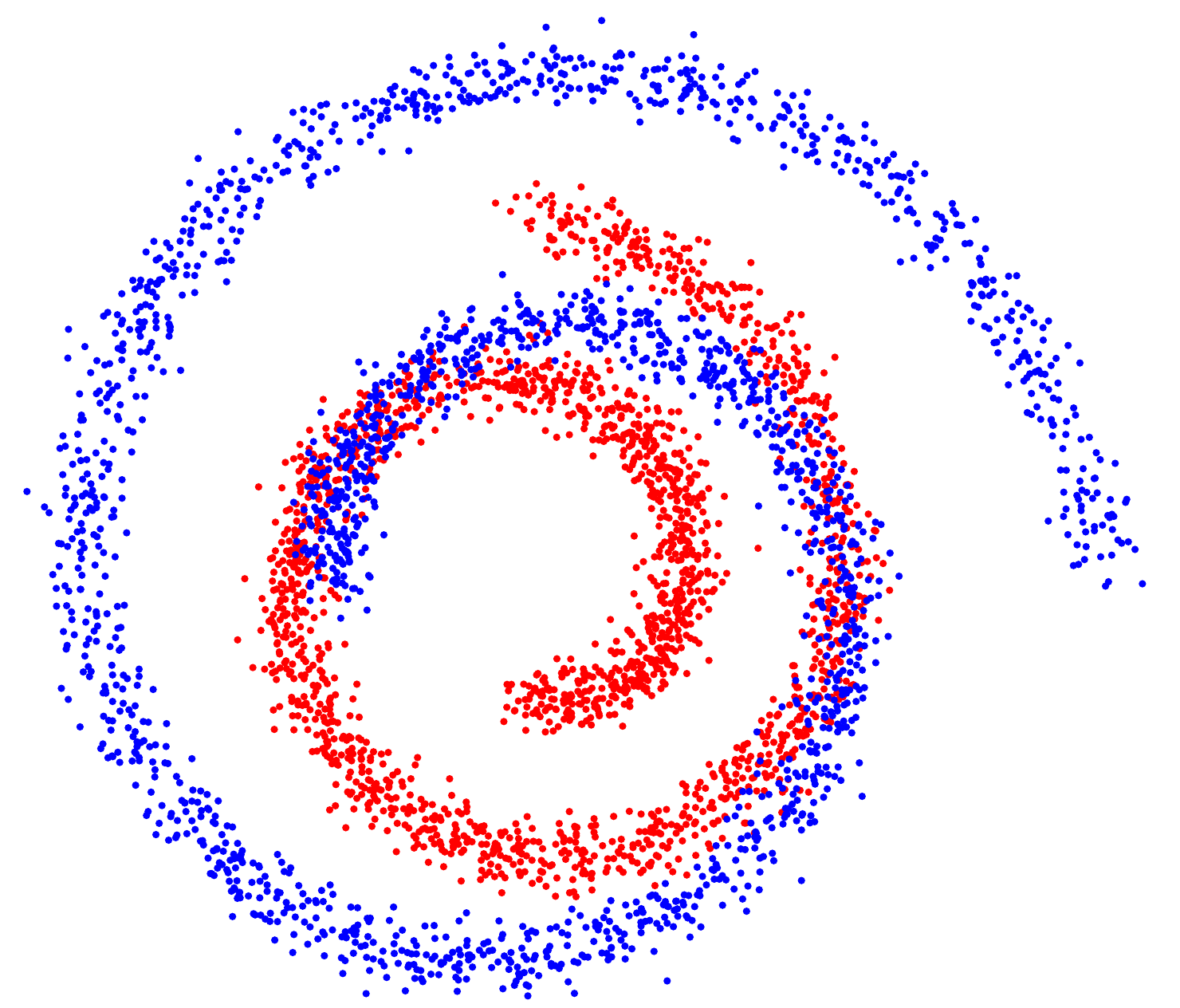}&
\includegraphics[width=2.3cm]{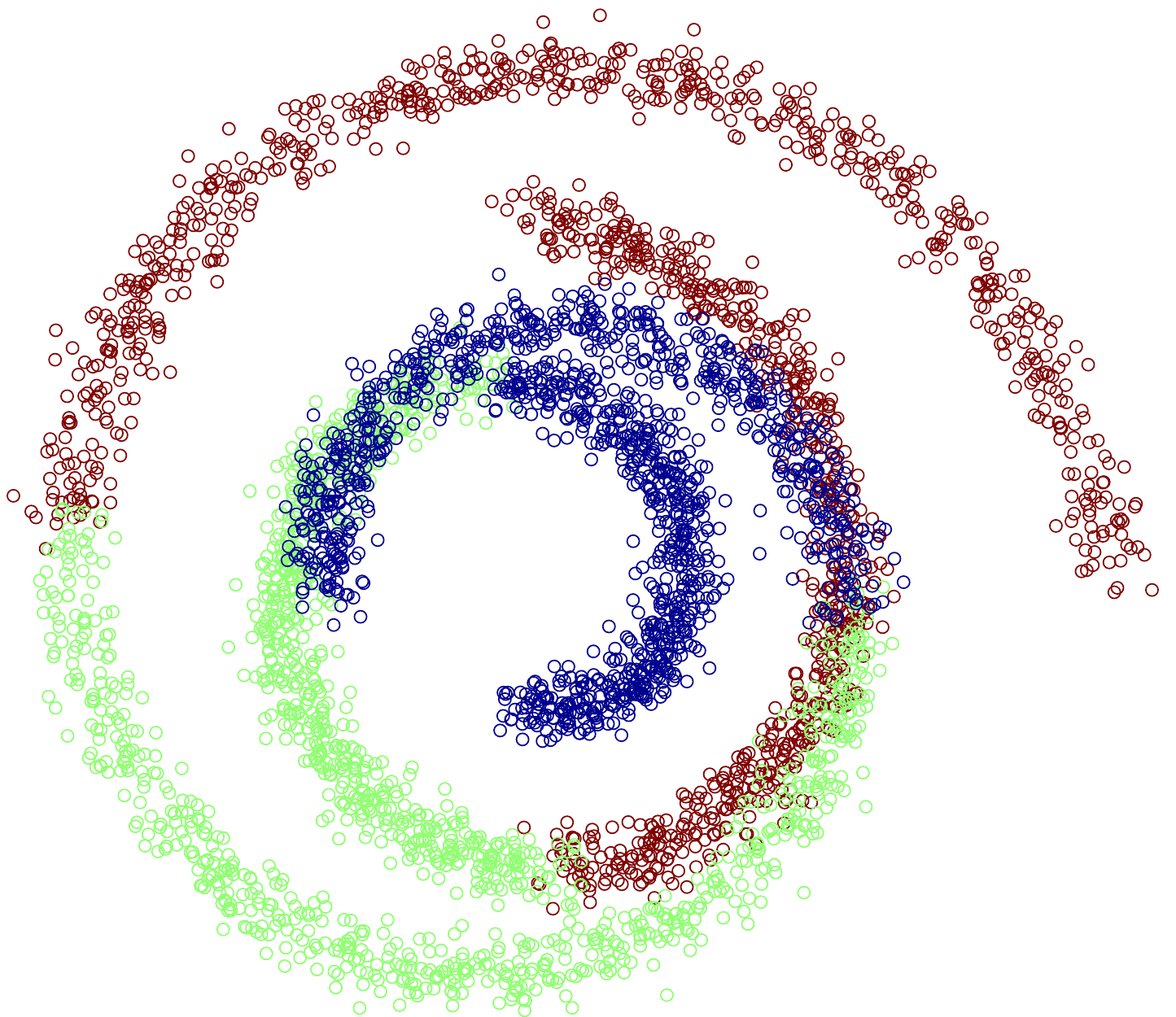}&
\includegraphics[width=2.3cm]{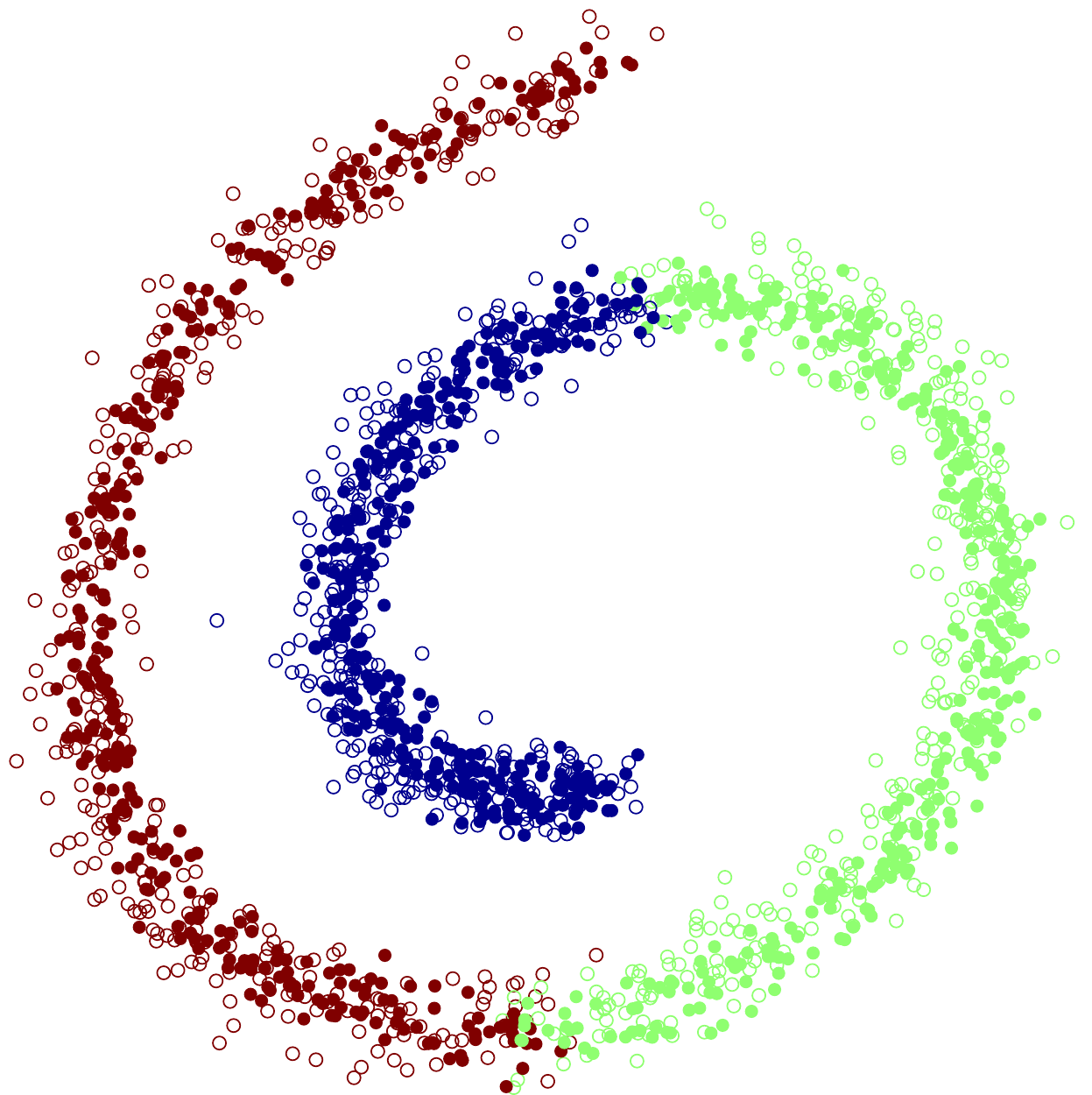}&
\includegraphics[width=3.5cm]{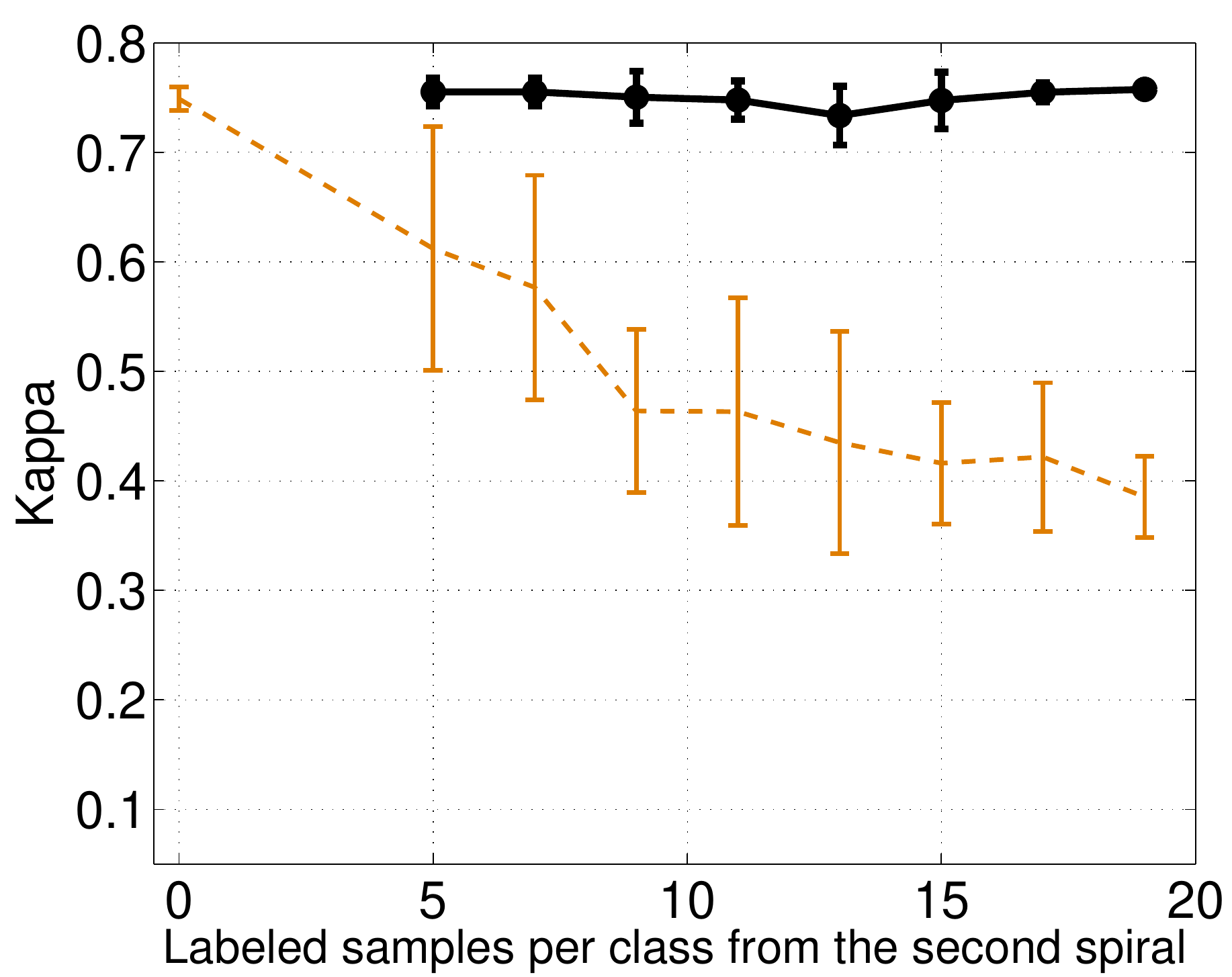}&
\includegraphics[width=3.5cm]{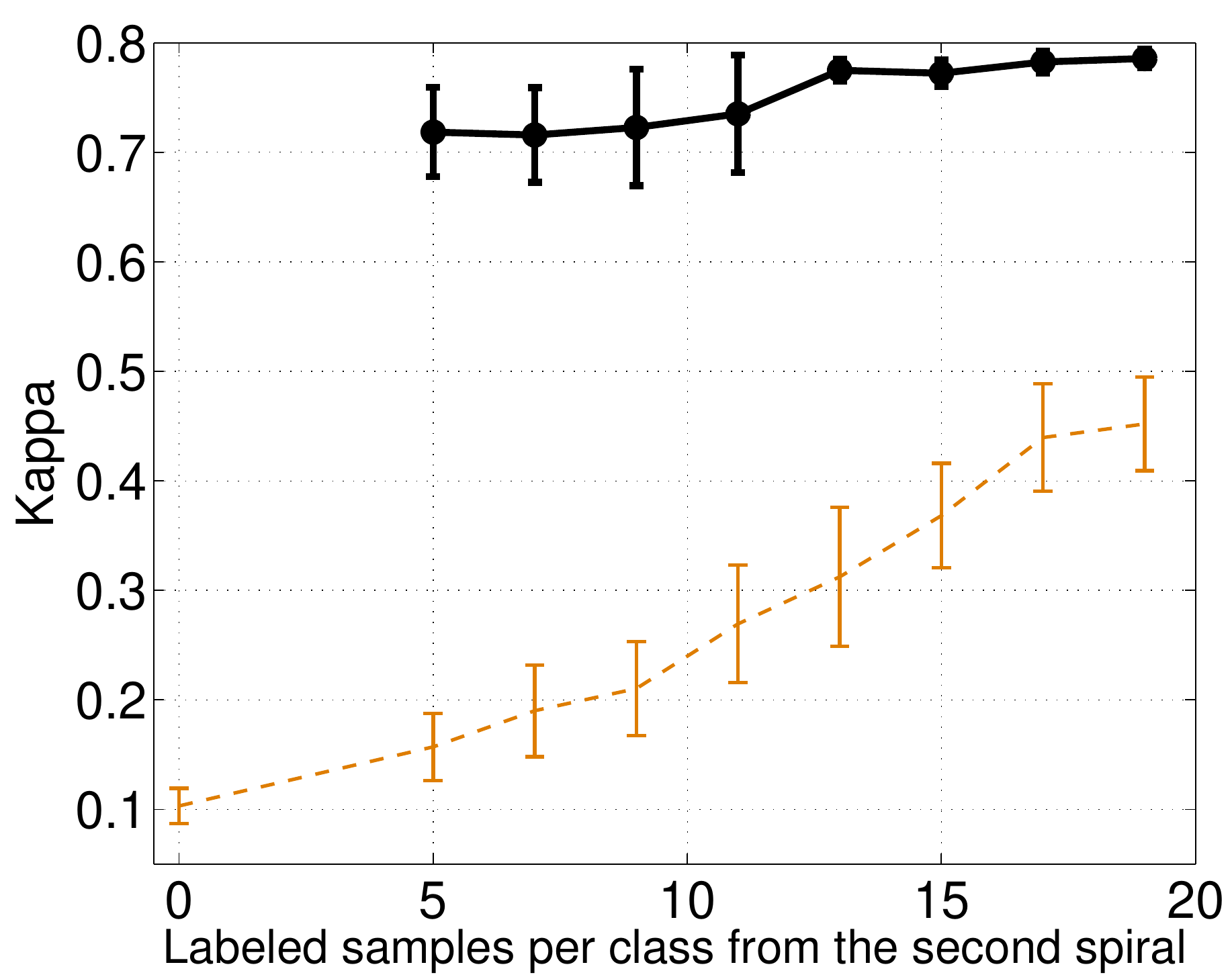}\\\hline

\rotatebox{90}{S + R + T}&
\includegraphics[width=2cm]{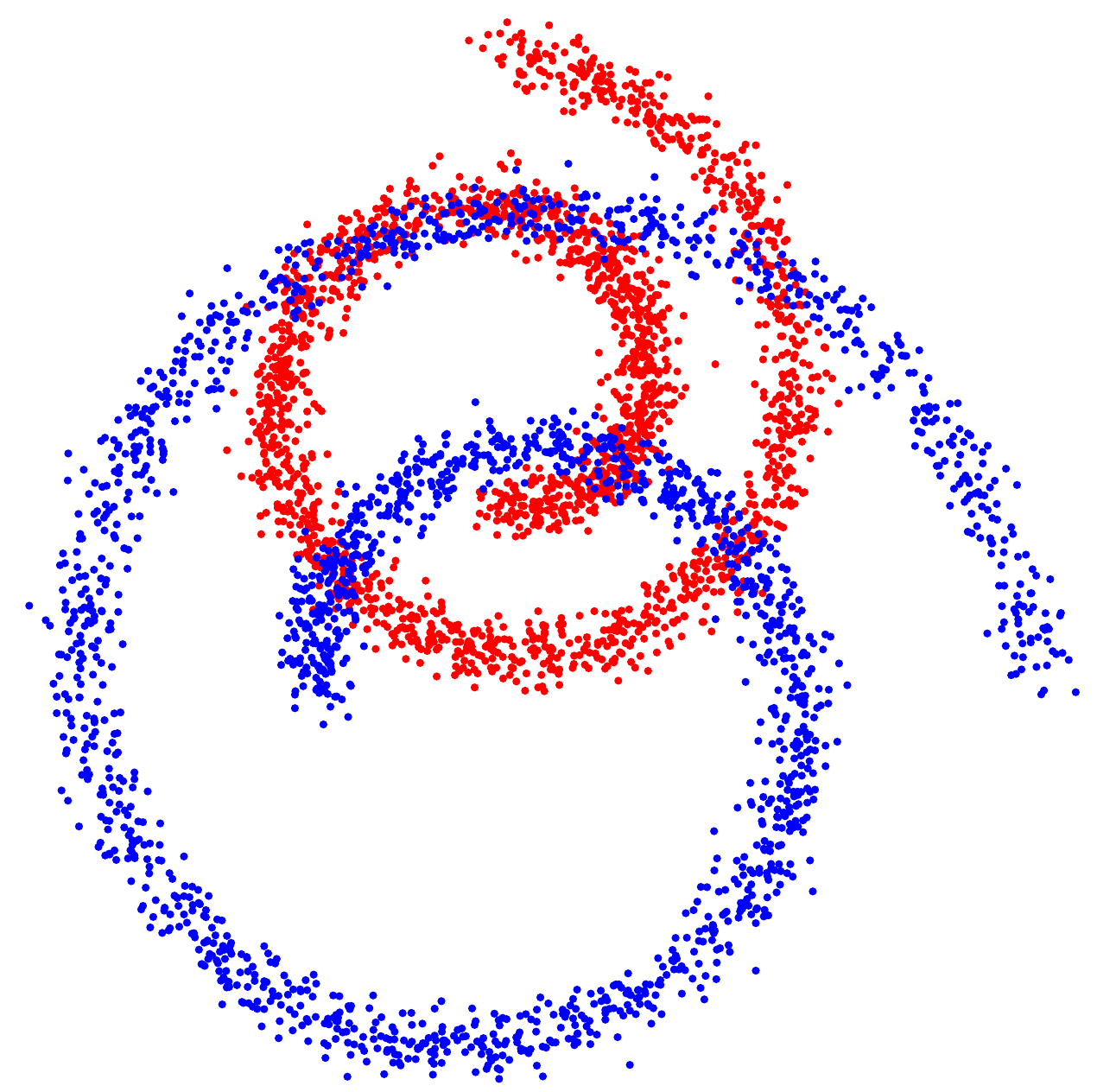}&
\includegraphics[width=2cm]{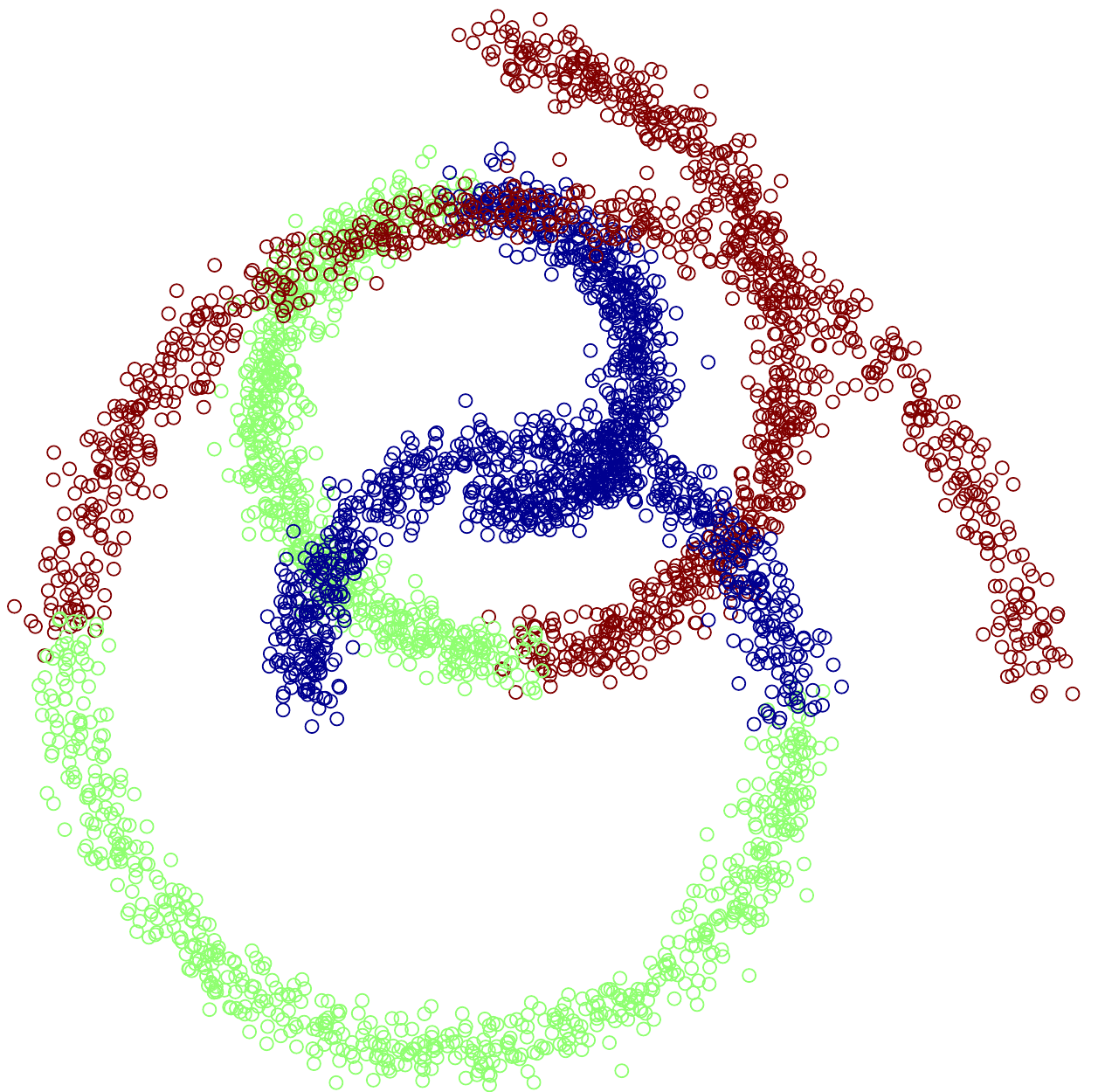}&
\includegraphics[width=2cm]{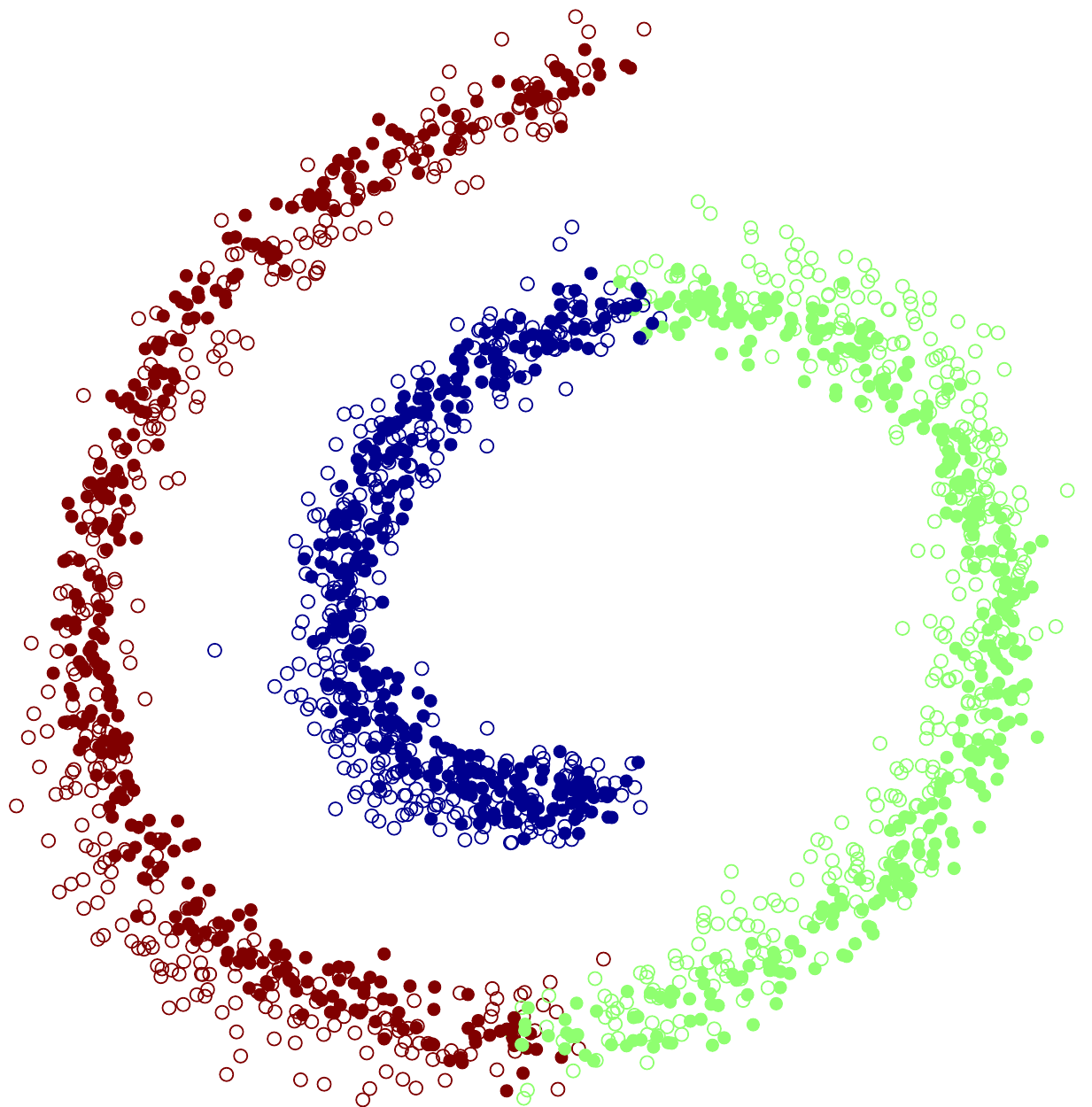}&
\includegraphics[width=3.5cm]{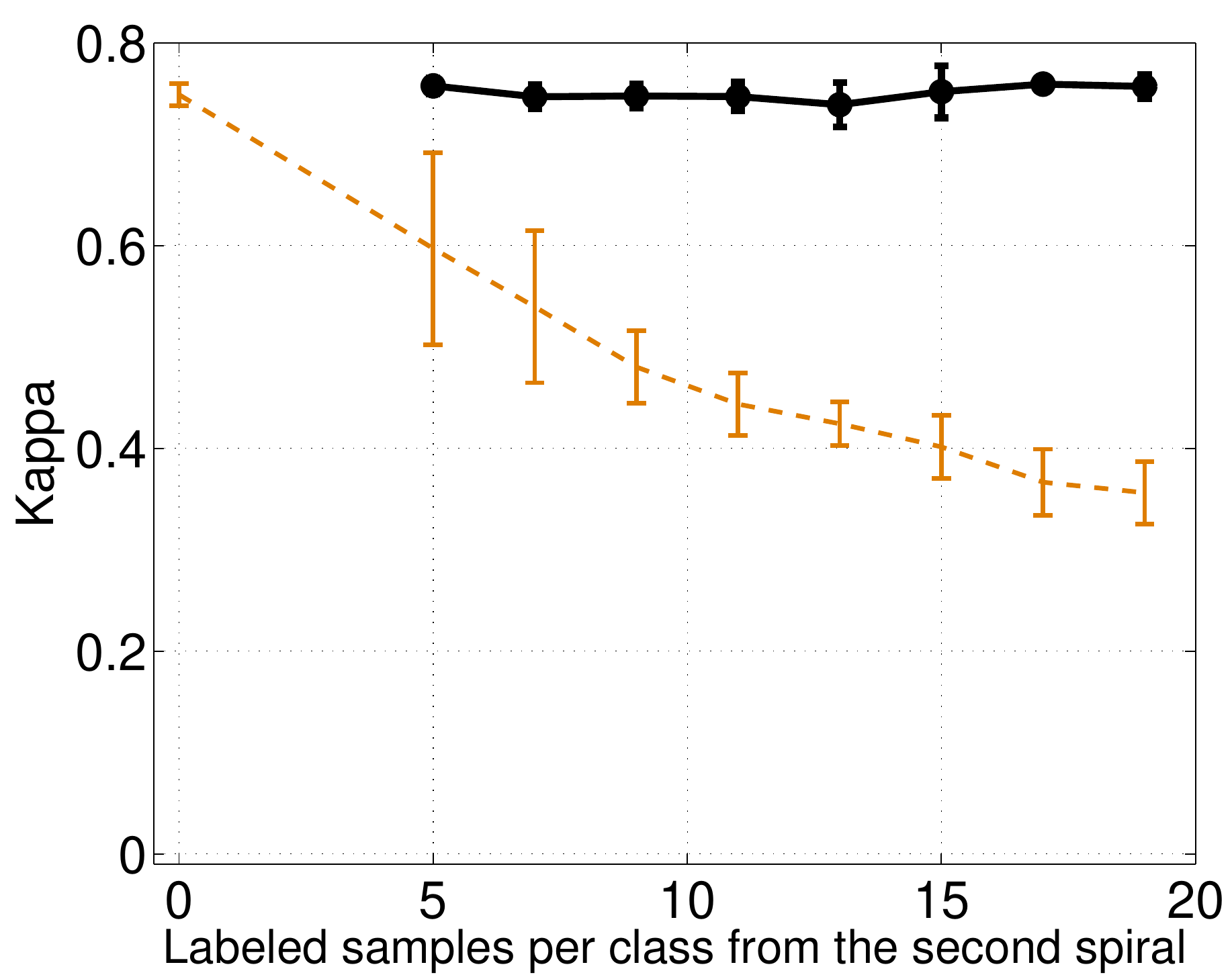}&
\includegraphics[width=3.5cm]{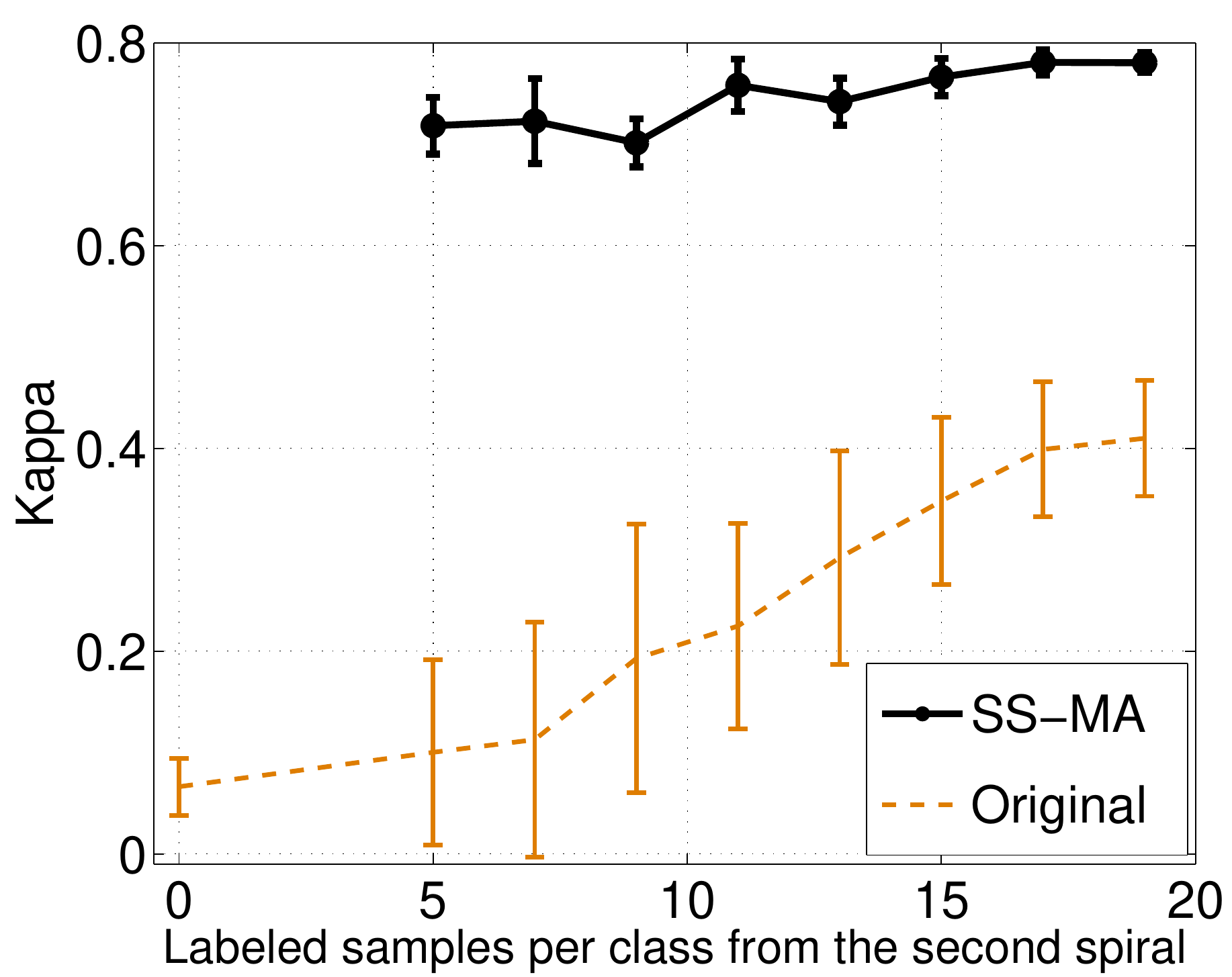}\\
\end{tabular}

\caption{\blue{Toy examples considered to test the invariance to scaling (S) rotation (R) and translation (T) of the SS-MA solution}. (a): the two sources ($\red{\bullet}$ and $\blue{\bullet}$) under different deformations (S = scaling; R = rotation; T = translation). (b) The distribution of the three classes. (c) Result of alignment with SS--MA. (d) Classification results with a linear SVM trained using 20 samples from the first source ($l_{\red{\bullet}}$= 20 {\em per} class) plus an increasing number of samples from the second ($l_{\blue{\bullet}}$ = [0, 5, ..., 20] {\em per} class) when predicting the test samples from the first spiral (Test = $\red{\bullet}$). (e) Classification results in the same conditions, but predicting the second spiral (Test = $\blue{\bullet}$).}
\label{fig:toyres}
\end{figure*}

The proposed SS--MA is tested in four scenarios:
\begin{itemize}
\item \emph{2D Toy dataset under different deformations}: the first example is designed to test the robustness of the proposed SS-MA transformation to different deformations of different levels of complexity. To do so, we \blue{generated toy datasets} composed by two spirals embedded in a 2D space (red and blue spirals on the first column of Fig.~\ref{fig:toyres}). The red spiral remains unchanged, while the blue one is either 1) scaled \blue{(S)}, 2) scaled and rotated \blue{(S + R)}, and 3) scaled, rotated  and translated \blue{(S + R + T)}. The two \blue{domains} are composed of 2000 labeled examples each, whose classes are illustrated on different colors on the second column of the figure.

\item \emph{Multiangular adaptation}: the first real experiment considers portability of a model along a series of five angular views of the city center of Rio de Janeiro, Brazil, acquired by WorldView-2~\cite{Pac11}. The interest of this dataset is that the images are acquired at a very short time interval and capture the same area, so that the spectral deformations should be only due to angular effects, since the atmosphere is unchanged along the acquisitions. Figure~\ref{fig:ang} shows an RGB composite of the five images. The number of labeled pixels available for the twelve classes available in each domain is detailed in Tab.~\ref{tab:ang}.

\item \emph{Multitemporal adaptation for the same sensor}: this is the classical scenario for domain adaptation in which, given one image with sufficient ground truth, we want to apply the best possible classifier for that image to other images taken at different times and places and with minimal effort in photointerpretation (i.e. the least number of new labeled pixels). The images in this experiment are three WorldView-2 scenes extracted from two images of the city of Lausanne, Switzerland: the Montelly and Malley images are subsets of images acquired the 29$^{th}$ of September 2010, while the Prilly subset is part of a scene acquired the 2$^{nd}$ of August 2011. All scenes have been pansharpened using the Gram-Schmidt transform. Figure~\ref{fig:im} illustrates the RGB composite and the exhaustive ground truth of these datasets. The number of labeled pixels available is detailed in Tab.~\ref{tab:pix}.

\item \emph{Multisource adaptation}: in the last experiment, we test the capability of the proposed SS-MA to align manifolds of different input dimensionality. To do so, we consider the 8-bands available in the Montelly and Prilly datasets, but use as a third image a 4-bands QuickBird image of Zurich (Switzerland) acquired the 6$^{th}$ of October 2006. The image and the corresponding ground truth are illustrated in the last row of Fig.~\ref{fig:im}. The number of labeled pixels available is detailed in Tab.~\ref{tab:pix}. All the images show periurban areas and involve a common set of classes.

\subsection{Experimental setup}

\end{itemize}

\definecolor{resi}{RGB}{255,102,51}
\definecolor{mea}{RGB}{0,235,50}
\definecolor{tr}{RGB}{51,155,0}
\definecolor{ro}{RGB}{120,120,120}
\definecolor{sh}{RGB}{51,255,255}
\definecolor{comm}{RGB}{255,0,0}
\definecolor{rail}{RGB}{204,204,204}
\definecolor{bare}{RGB}{153,102,51}
\definecolor{high}{RGB}{204,204,0}

\begin{table}[!t]
\caption{\blueR{Number of labeled pixels available for each dataset in the multiangular experiments ($\theta$ = off-nadir angle).}}
\label{tab:ang}
\begin{tabular}{c|c|c|c|c|c}
\hline
\backslashbox{Class}{$\theta$} & $-38.79^\circ$ & $ -29.16^\circ$ & $ 6.09^\circ$ & $ 26.76^\circ$ & $ 39.5^\circ$ \\\hline\hline
Water & 83260 &79937& 66084& 63492 & 54769 \\
Grass & 8127 &8127& 8127& 8127 & 8127 \\
Pools & 244&244& 223 & 195 & 195 \\
Trees & 4231 &4074& 3066& 3046 & 3046 \\
Concrete & 707 &719&719 & 719 & 696 \\
Bare soil & 790 &790& 790& 790 & 811 \\
Asphalt & 2949 &2949& 2949& 2827 & 2827 \\
Grey buildings & 6291 &6061& 5936& 4375 & 4527\\
Red buildings & 1147&1080& 1070& 1046 & 1042 \\
White buildings & 1683 &1683& 1571& 1571 & 1571 \\
Shadows & 1829 &1056& 705& 512 & 525 \\
Tarmac & 5179 &5179& 5179& 2166 & 2758 \\
\hline
\end{tabular}
\end{table}

\begin{table}[!t]
\caption{Number of labeled pixels available for each dataset in the multitemporal and multisource experiments}
\label{tab:pix}
\vspace{-0.25cm}
\begin{tabular}{c|c|cccc}
\hline
\multirow{2}{*}{Class} & Color in & \multicolumn{4}{c}{Dataset}\\
&Fig.~\ref{fig:im}& Prilly & Montelly & Malley & Zurich \\\hline\hline
Residential & \colorbox{resi}{\textcolor{resi}{m}} & 151271 & 179695 & 24898 &  78018  \\
Meadows & \colorbox{mea}{\textcolor{mea}{m}} & 148604 & 47865 &  143674 & 12347  \\
Trees & \colorbox{tr}{\textcolor{tr}{m}} & 116343 & 177203 & 63956 & 52812 \\
Roads & \colorbox{ro}{\textcolor{ro}{m}} & 141353 & 104582 & 294687 & 43005 \\
Shadows & \colorbox{sh}{\textcolor{sh}{m}} & 39404 & 218189 & 194321 & 14071 \\
Commercial & \colorbox{comm}{\textcolor{comm}{m}} & 13692 & 22506 & 437633 & 25389 \\
Rail & \colorbox{rail}{\textcolor{rail}{m}} & -- & 8570 & 76802 & --\\
Bare & \colorbox{bare}{\textcolor{bare}{m}} & -- & 2812 & 33305 & -- \\
Highway & \colorbox{high}{\textcolor{high}{m}} & -- & -- & -- & 28827 \\\hline
\end{tabular}
\end{table}

As proposed in~\cite{Wan11b}, all the similarity matrices $\W$ are scaled to have the same Frobenius norm in order to make them comparable in terms of energy. Consequently they all have the same contribution if $\mu = 1$. 
From all the available labeled pixels \blue{in each image}, 50\% are kept apart as the testing set. The remaining 50\% samples are used to extract the labeled and unlabeled pixels used in the feature extractor.

The projection matrices $\F$ are learned in a traditional domain adaptation setting, where one domain provides many labeled examples $l_1$, and the other(s) have limited examples $l_{\blue{M} \setminus 1}$. \blueRR{}{Since the projectors and models are trained with labeled samples from each domain, all the classes of all the images are represented simultaneously. This way, problems related to the appearance of new classes are obviated.} All the domains provide the same number of unlabeled samples $u_m$. The dataset sizes reported in Table~\ref{tab:n} have been considered.

\begin{table}[!t]
\caption{Number of labeled and unlabeled pixels used in the experiments.}
\label{tab:n}
\vspace{-0.25cm}
\begin{tabular}{c|c|c|c}
\hline
Experiment & $l_1$ & $l_{\blue{M} \setminus 1}$ & $u_m$ \\\hline\hline
Toy 2D & 20 per class & [5, ..., 20] per class & 300 per source\\
Multiangular & 100 per class & [10, 50] per class & 500 per source\\
Multitemporal & 100 per class & [10, ..., 90] per class & 500 per source\\
Multisource & 100 per class & [10, ..., 90] per class & 500 per source\\\hline
\end{tabular}
\end{table}

In the multiangular experiment, only the near-nadir image ($6.07^\circ$) is considered as the heavily labeled one ($l_1$). In the multitemporal and multisource experiments, all the domains are considered in turn as the heavily labeled. 

The $u_m$ unlabeled examples are selected using an iterative clustering algorithm, the bisecting $k$-means~\cite{Kas09}. As in~\cite{Mun11}, we run binary partitions of the dataset until retrieving $u_m$ clusters. The centroid of the cluster is then used as unlabeled example, thus ensuring an uniform sampling. The number of unlabeled samples used is reported for each experiment in Table~\ref{tab:n}. To build the geometrical Laplacian $\L\blueRR{}{_g}$ in Eq.~\eqref{g}, we used a series of $M$ graphs built using $k$-NN with $k=9$ and all the $n_m$ labeled and unlabeled samples. \blueRR{}{The parameter $k$ defines the locality of the graph and, in our experimental tests, seemed not to influence the final results in a significant way, as long as it is set in a reasonable range, i.e. a range where the manifold structure is not lost in an over-connected graph or oversimplified by an under-connected graph.}

After extraction of the projection matrix $\F$, the best number of dimensions for classification is cross-validated exploiting the same labeled pixels already used to minimize Eq.~\eqref{costABC}. To assess the performance of the alignment, a linear SVM has been used to classify the original and projected data. Even though one could use more sophisticated and eventually nonlinear classifiers, our interest here is to assess the quality and discriminative power of the encountered projection features. The algorithm is actually nonlinear through the definition of the graph Laplacian so in principle there is no need to perform nonlinear classification. The libSVM library~\cite{CC01a} has been used for such a classification and the regularization $C$ parameter has been cross-validated using the entire labeled set in the range $[100, \ldots, 1000]$. In the multi-angular experiment, additional results obtained with other projectors (PCA, KPCA and graph matching~\cite{Tui12a}) and classifiers (LDA) are \blue{also} reported\blue{. Finally, an in-depth analysis of the best dimensionality of the latent space is also conducted for LDA}. 

A final remark concerns the multisource experiment. Since the domains have different dimensionality, it is not possible to learn a classifier using simultaneously pixels from WorldView-2 and QuickBird. To do so, we downgraded the WorldView-2 images to the four QuickBird bands and provide the results obtained using only the four bands. 

In all cases, the  results reported are averages \blueR{of the estimated Cohen's Kappa agreement score ($\kappa$ hereafter~\cite{Foo04c})}  over five realizations of the sequence {\em projection plus classification}  in all domains with the projected data.  This means that the projections are derived once, a single classifier is trained on the projected training samples, and then applied to each of the \blue{projected} images, considered in turn as the test image. \blue{We set the random generator in a way that the samples of each run are automatically included in the corresponding run using more labeled pixels, i.e. the pixels included in the first run using 10 pixels \emph{per class} are also used in the first run using $20, 30, ...$ pixels per class.} Note that no specific training process for each test image is performed, as the aim is to classify in the latent space, regardless of the image which will be projected into it.

\vspace{1cm}
\section{Results and discussion}\label{sec:res}

This section presents and discusses the results obtained in the four case studies described in Section~\ref{sec:data}. 

\subsection{2D toy datasets}

The two first columns of Fig.~\ref{fig:toyres} represent the three deformations considered: 1) scaling \blue{(S)}, 2) scaling plus rotation \blue{(S + R)} and 3) scaling, rotation and translation \blue{(S + R + T)}. The third column illustrates the result of SS--MA in one of the five experiments performed in each case. In all cases, the proposed method compensates for the present deformations and aligns the two spirals in the correct way. Subsequent classifications by a linear SVM (reported in the fourth and fifth columns of Fig.~\ref{fig:toyres} show two trends:
\begin{itemize}
\item[-] When predicting data from the first spiral ($\red{\bullet}$, 20 labeled pixels {\em per} class), adding samples from the other one does not influence the classification results. On the contrary, when using unprojected data, a strong decrease in performance is observed in the three cases. \blue{SS--MA shows a stable behavior, since the distributions have been aligned correctly.}
\item[-] When predicting data from the second spiral ($\blue{\bullet}$, [5, ..., 20] labeled pixels {\em per} class), SS--MA can exploit the common information of the two datasets efficiently and it always outperforms the unprojected data.
\end{itemize}

\subsection{Multiangular adaptation}

\blue{The first real} experiment considers the multiangular sequence of Rio de Janeiro. \blue{As this dataset comprises a set of images from the same sensor and acquired at a very short time interval, the shifts observed are only related to angular effects of increasing strength. Besides analyzing the classification accuracy with respect to a set of competing methods, we also use this dataset to study the evolution of the performance with respect to the dimensionality of the latent space}. 

\begin{figure*}[!t]
\centerline{\begin{tabular}{cccc}
\rotatebox{90}{ Domains}&
\includegraphics[width=0.28\textwidth]{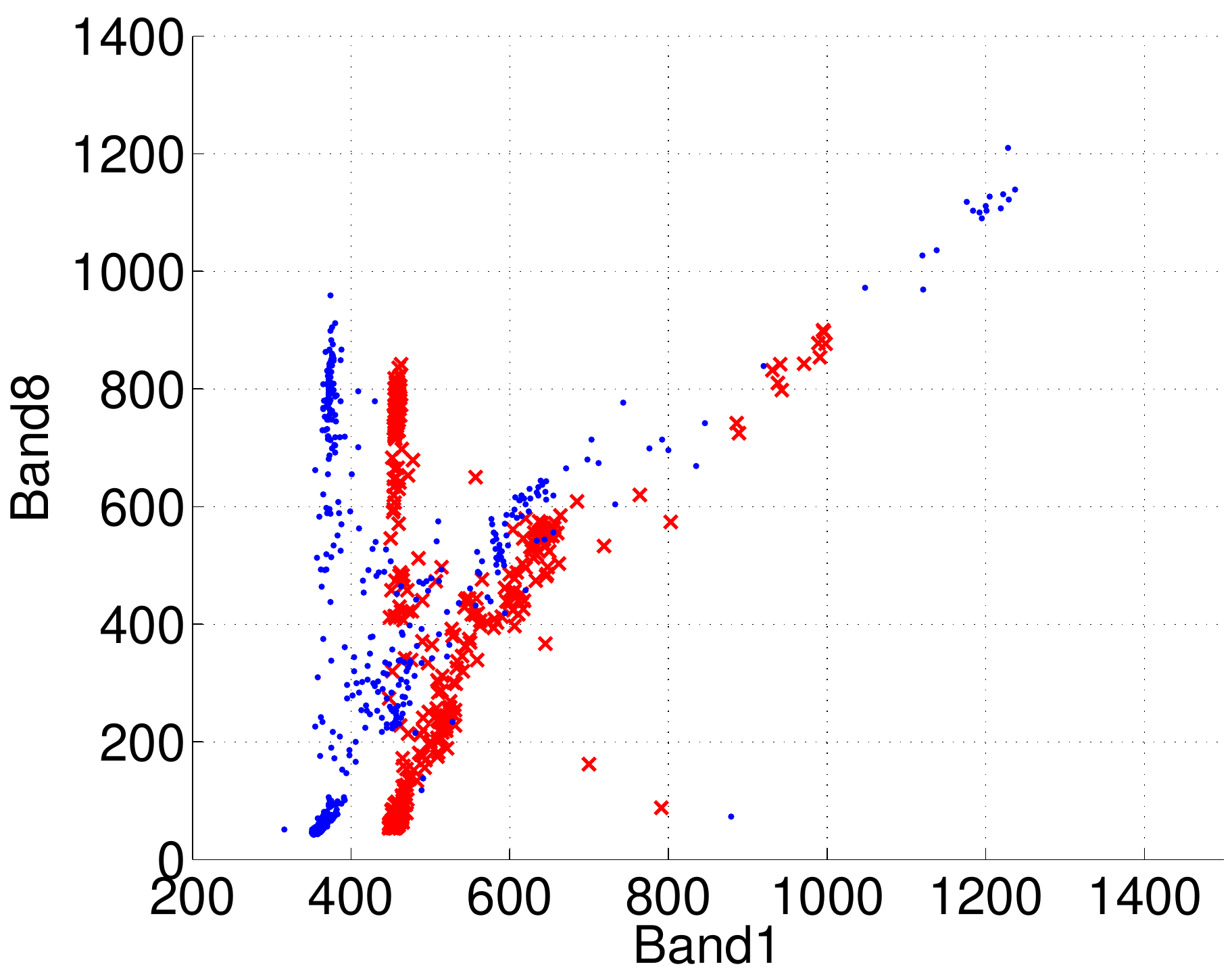}&
\includegraphics[width=0.28\textwidth]{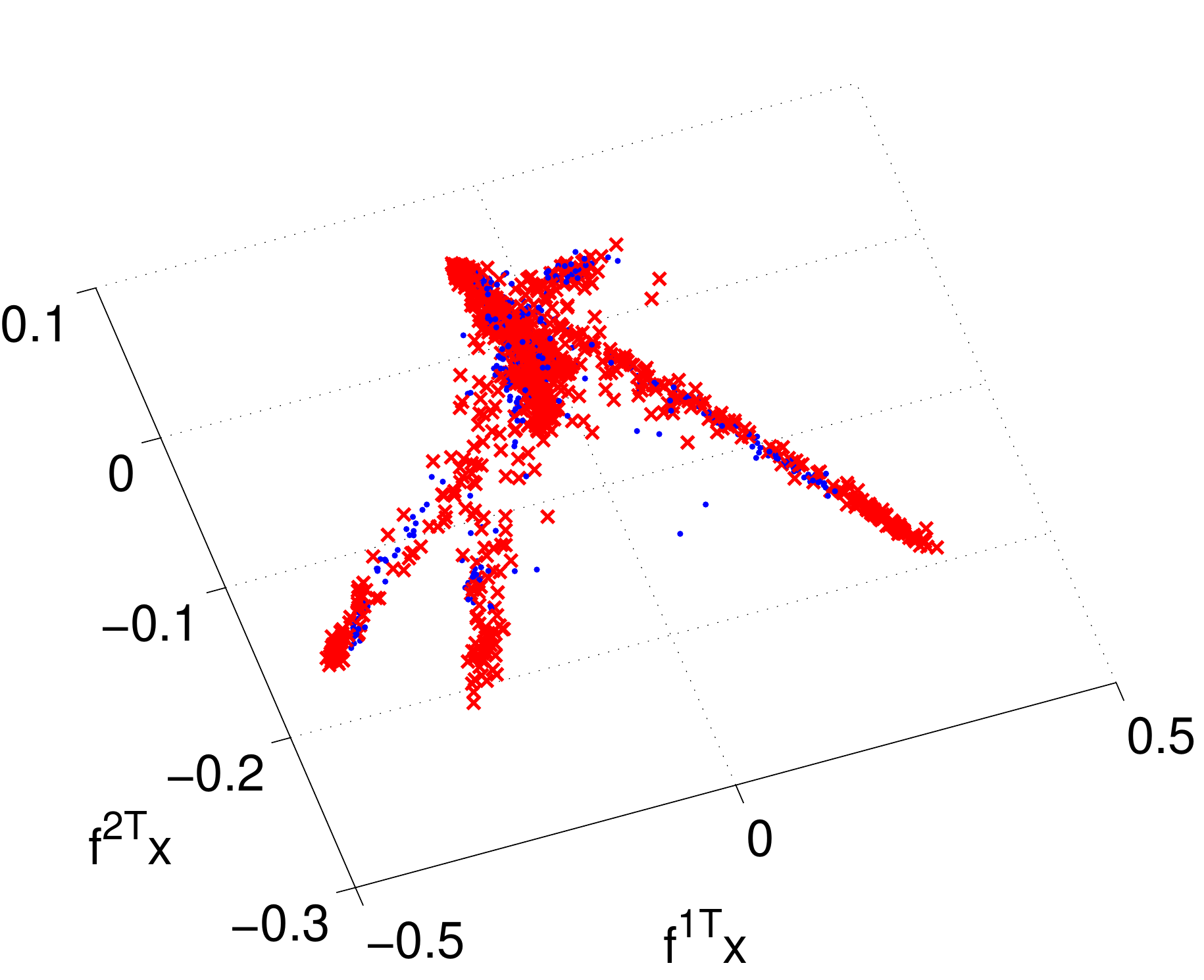}&
\includegraphics[width=0.28\textwidth]{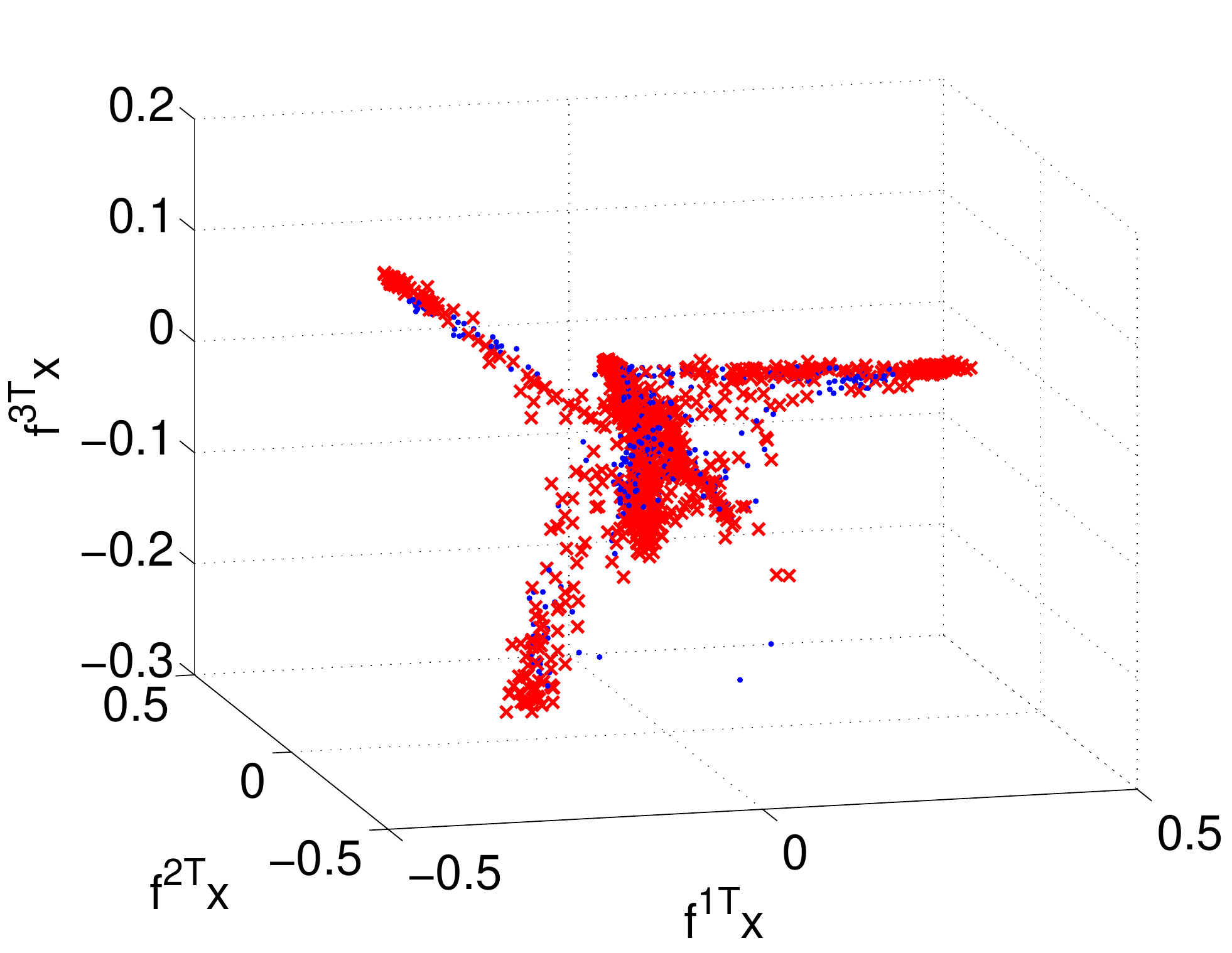}\\
\rotatebox{90}{ Classes}&
\includegraphics[width=0.28\textwidth]{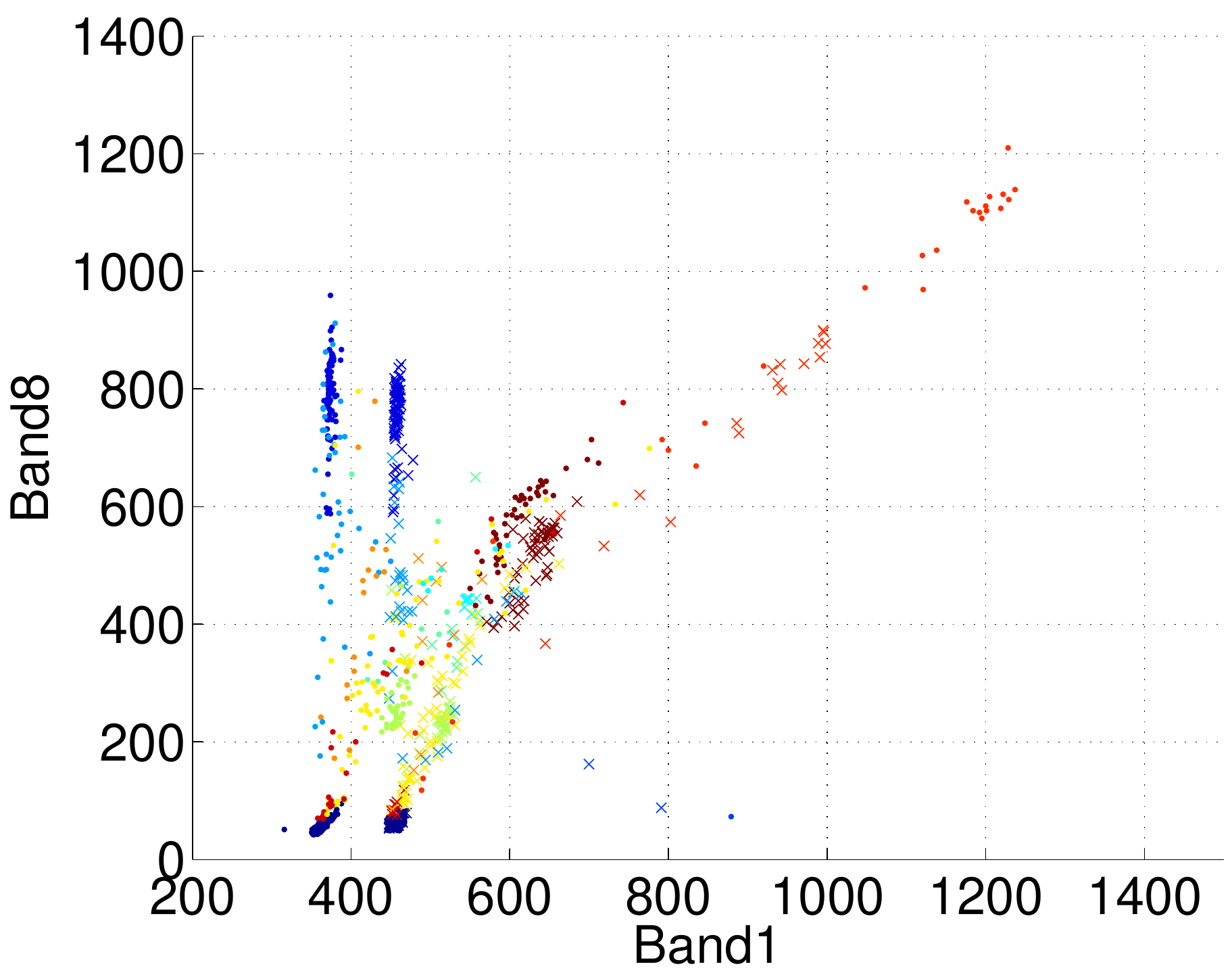}&
\includegraphics[width=0.28\textwidth]{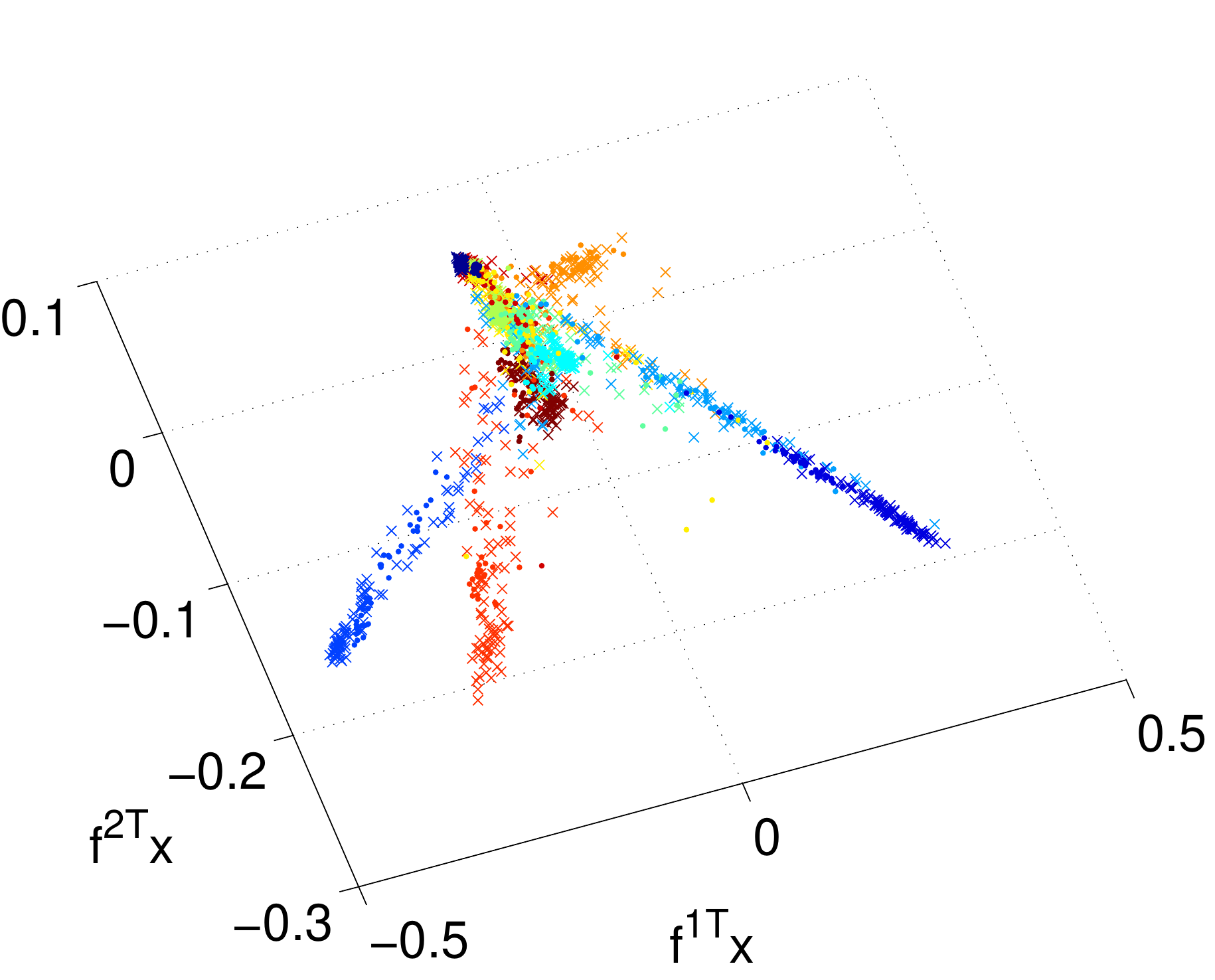}&
\includegraphics[width=0.28\textwidth]{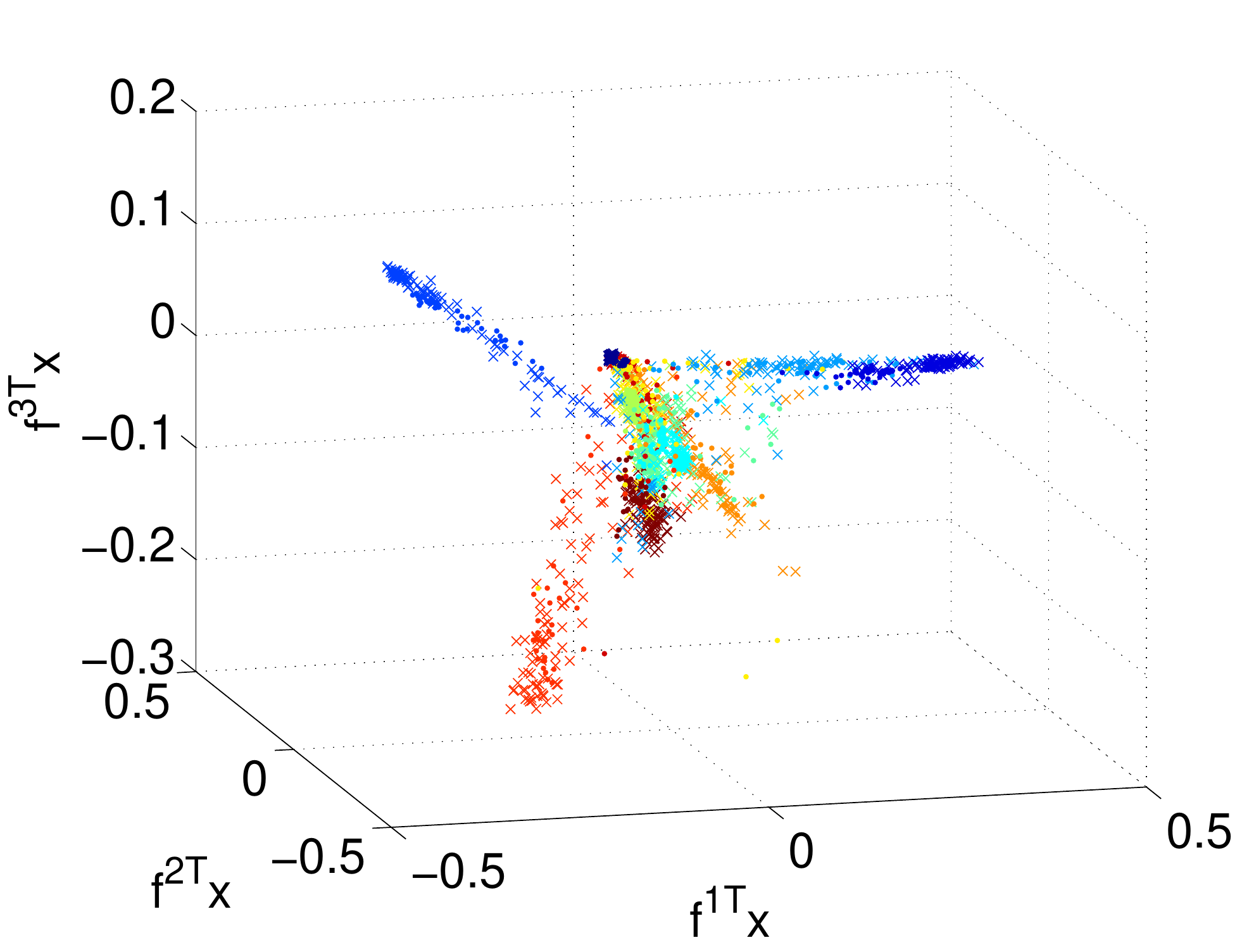}\\
&Original bands & Projected (2D) & Projected (3D)\\
\end{tabular}}
\caption{\blue{SS--MA projections. Left: original acquisitions at $6.09^\circ$ and $-38.79^\circ$. Center: two first dimensions of the latent space. Right: three first dimensions of the latent space. In the top row, each color corresponds to a domain (an image), in the bottom row, each color corresponds to a class.}}
\label{fig:proj}
\end{figure*}

\blue{First, we visualize the latent space, to assess its discriminative power. Figure~\ref{fig:proj} illustrates the projections of the images at $6.09^\circ$ and $-38.79^\circ$ in the 2-dimensional (2 first eigenvectors) and 3-dimensional (3 first eigenvectors) latent space. From the plots, we can observe that the model defined a discriminative space, where joint linear classification is facilitated.}

\begin{figure}
\centering
\begin{tabular}{ccc}
\rotatebox{90}{\hspace{1.5cm} LDA}&
\includegraphics[width=3.7cm]{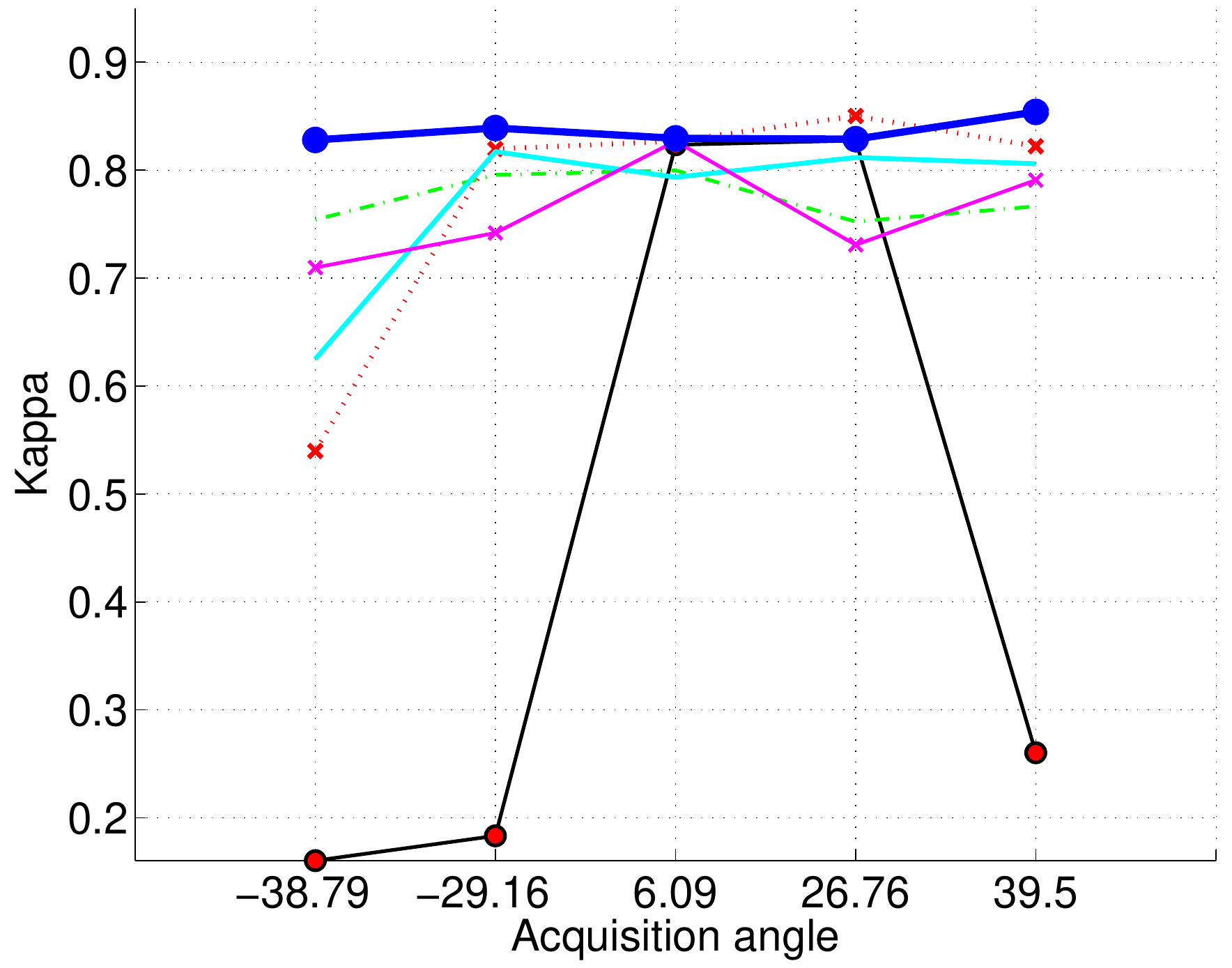}&
\includegraphics[width=3.7cm]{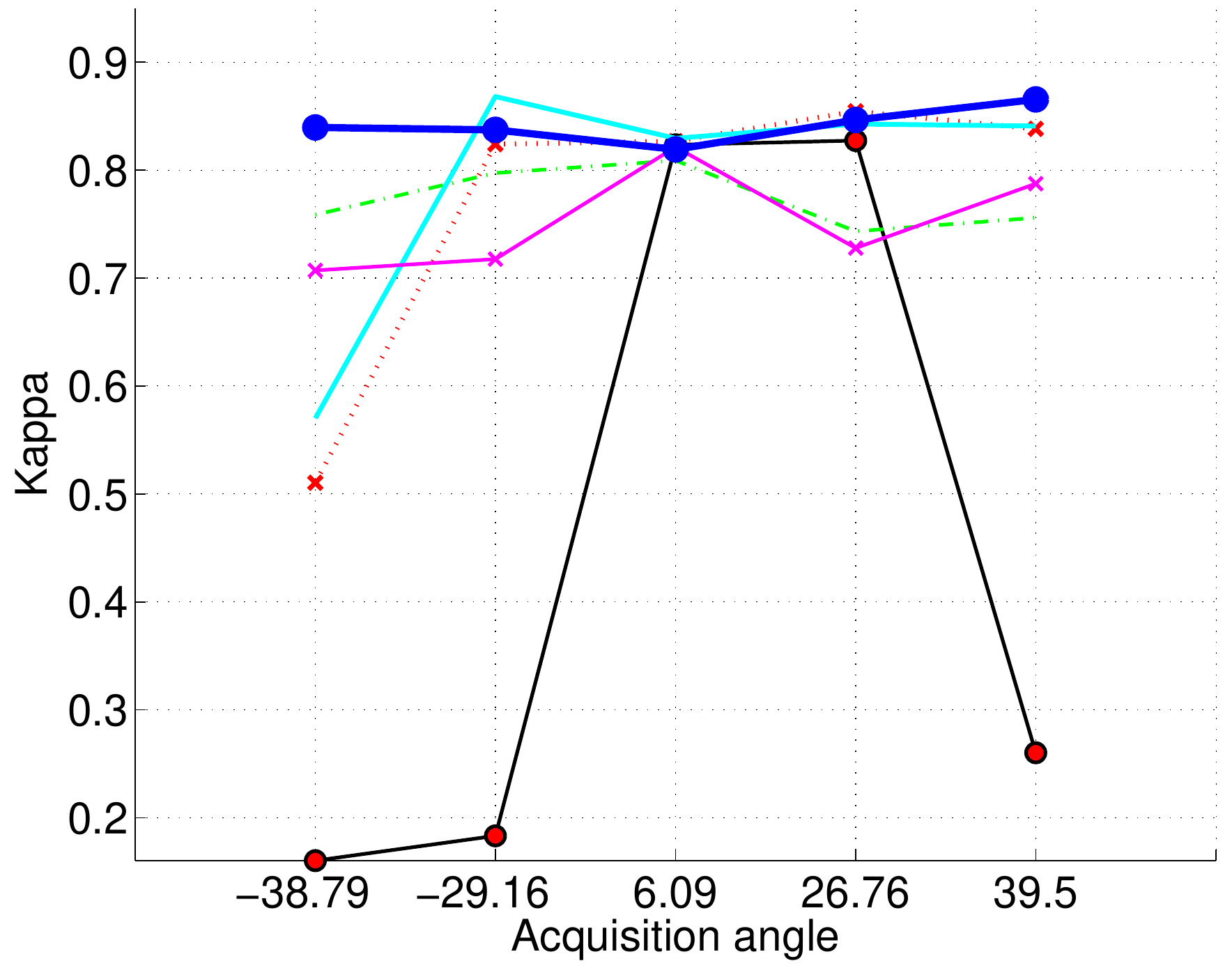}\\
\rotatebox{90}{\hspace{1.5cm} SVM}&
\includegraphics[width=3.7cm]{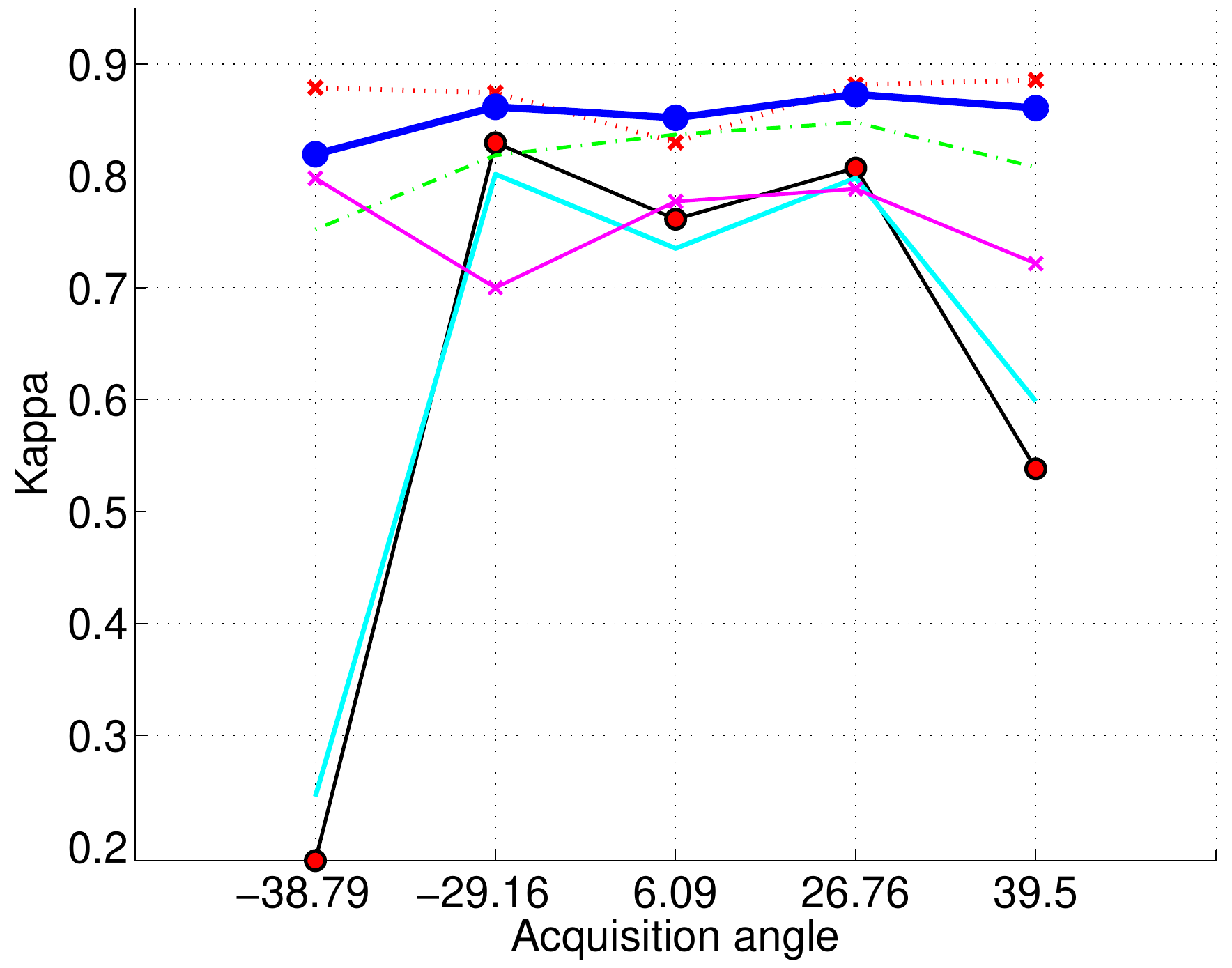}&
\includegraphics[width=3.7cm]{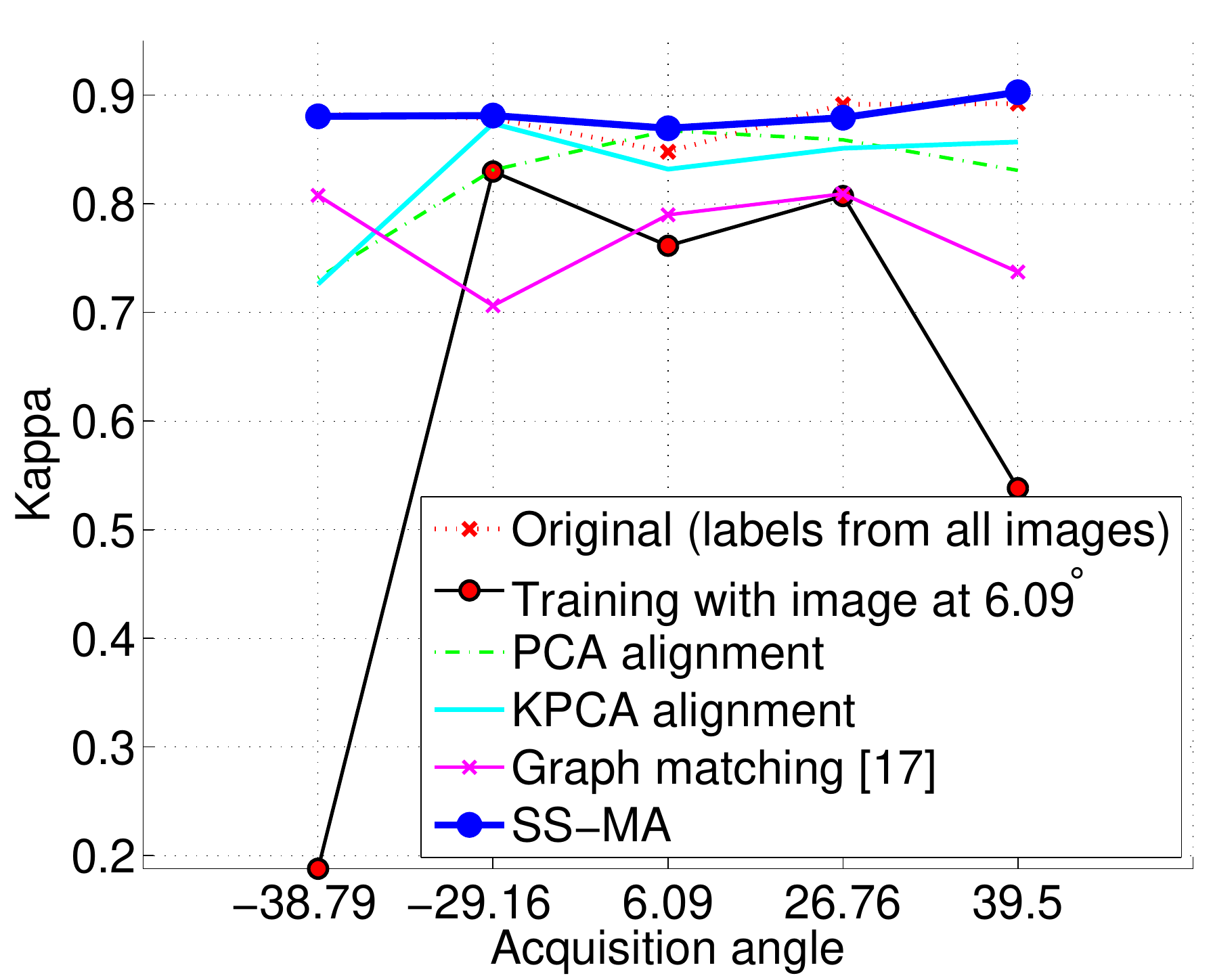}\\
& (a) & (b)  \\
& 10 pixels {\em per} class    & 50 pixels {\em per} class\\

\end{tabular}

\caption{\blue{Numerical performances of the multiangular experiments. The nadir image provides 100 labeled pixels {\em per} class, while the other are aligned with an increasing number of labeled pixels.}}
\label{fig:angres}
\end{figure}

Figure~\ref{fig:angres} illustrates the results obtained by 1) the baseline (using only the labeled pixels from the $6.09^\circ$ image as training set), 2) the joint use of all the acquisitions, but without projection, 3) the joint use of all the acquisitions after projection with the proposed SS--MA \blue{and 4) three aligning methods such as PCA, KPCA and graph matching~\cite{Tui12a}. To be fair in the evaluation, the projections for these methods are obtained in an unsupervised way, but then the classifier is trained using the original training points from the nadir acquisition, stacked to the transformed labeled pixels of the domain to be tested}.

Using only the nadir acquisition leads to poor results for increasing \blueRR{}{off-nadir} angles, at the point that the $\kappa$ statistic observed are close to 0.2 (\blue{black solid line with red markers}). Adding labeled information from the specific viewing angle helps and improves notably the results (\blue{red dashed line}). However, and especially when using a simple parametric model as LDA, the decrease in performance is still observed, especially for strong off-nadir acquisition. The proposed SS--MA (\blue{blue solid line}) results in an almost flat $\kappa$ surface along the acquisitions, thus showing that projection in the latent space permits 1) to align efficiently the spectra of the acquisitions, and 2) to reduce all the image-specific problems to the same (latent) classification problem. \blue{The results observed for SS--MA are always accurate and stable along the acquisitions: they join the stability of the unsupervised method designed specifically for domain adaptation (graph matching\blue{, pink line}) and the higher numerical results of a generic supervised one (the one depicted by the red curve, where samples from all acquisitions are used to train the model). SS--MA always outperforms PCA alignment (\blue{green line}) and is always at least as accurate as KPCA (which is also nonlinear -- \blue{cyan line}). Compared to KPCA, SS--MA shows more  stable results along the acquisitions, regardless of the classifier and of the number of labeled pixels used.}

\begin{figure}
\begin{tabular}{cc}
\includegraphics[width=4cm]{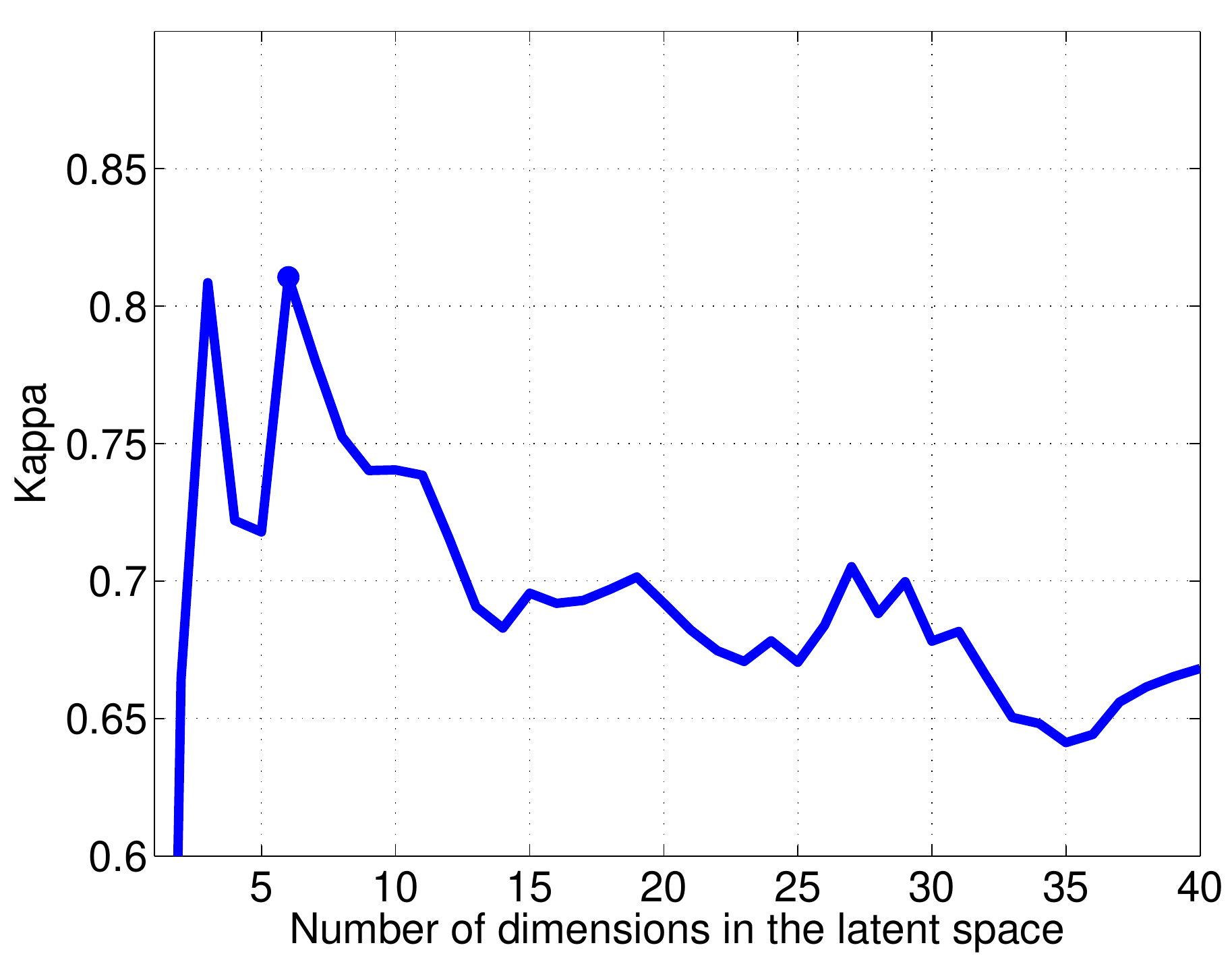}&
\includegraphics[width=4cm]{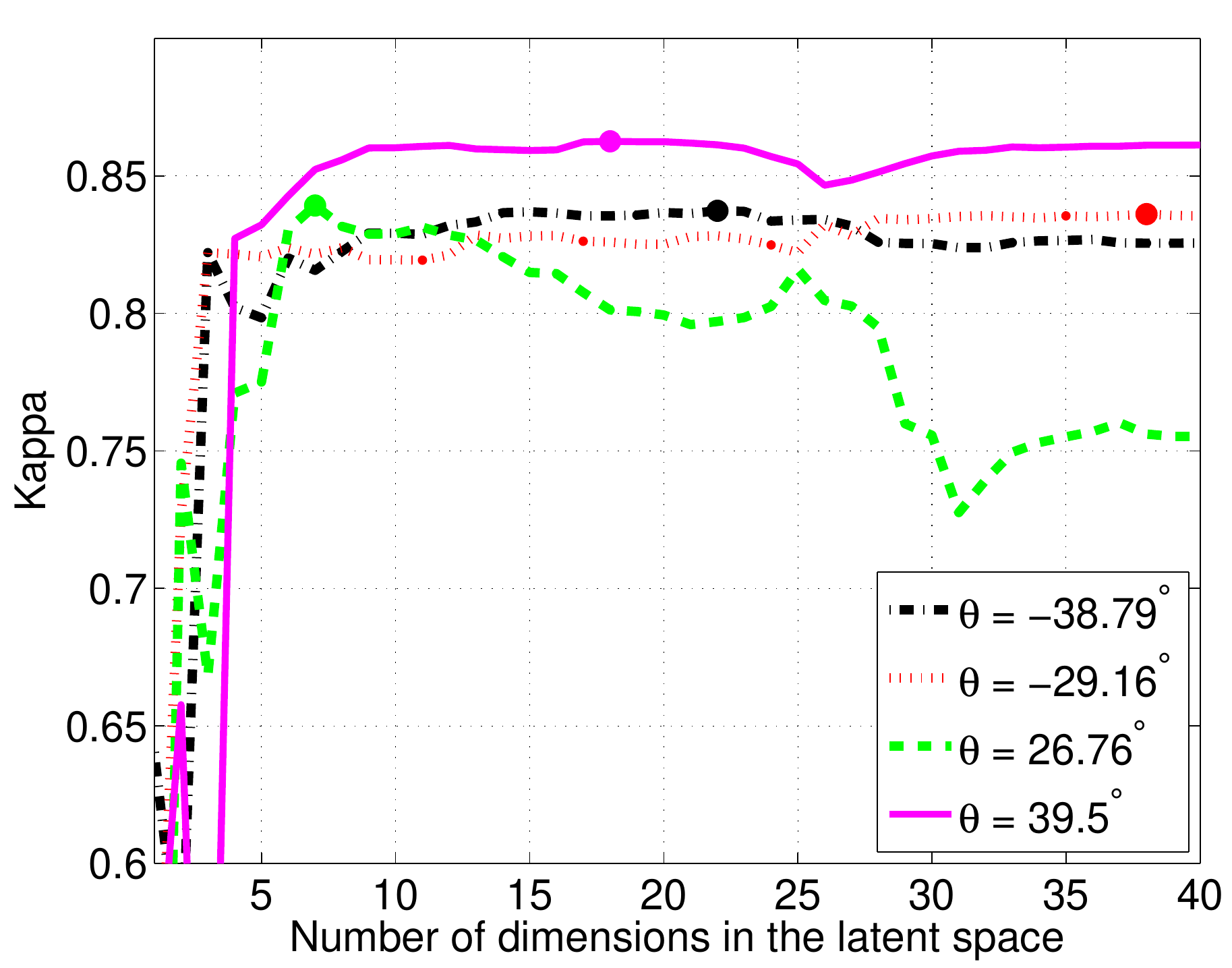}\\
(a)  & (b)  \\
\end{tabular}
\caption{\blue{Evolution of the average $\kappa$ along the dimensions of the latent space defined by the five domains ($d = 5 \times 8 = 40$) for the experiment using $50$ pixels \emph{per} class. (a) Predicting the most nadiral domain ($\theta = 6.09^\circ$); (b) predicting the four other domains. The filled dot corresponds to the dimension returning the higher average $\kappa$}}
\label{fig:dimLDA}
\end{figure}

\blue{The use of the LDA classifier also allows to assess the discriminative power of the latent space across its dimensions. Unlike SVM, which is robust to high dimensionality with noisy dimensions, LDA is strongly affected by non-discriminative dimensions and by non-Gaussian class-conditional distributions. Therefore, we studied the evolution of accuracy, when increasing the dimension of the latent space in which the classifier is trained. Fig.~\ref{fig:dimLDA}(a) shows the evolution of the $\kappa$ statistic for one run of the algorithm in the nadir case ($\theta = 6.09^\circ$). Fig.~\ref{fig:dimLDA}(b) shows the same evolution for the off-nadir angles. From these plots, we can conclude that when the shift between the domain providing most of the labeled samples and the destination domain is low (as for the $6.09^\circ$ and the $26.76^\circ$ images), a lower number of dimensions is sufficient to achieve the best performances. On the contrary, when the shift is larger, the latent space provides informative dimensions up to the last components. Another measure of the complexity of the alignment problem can be provided by studying the dimension of the latent space corresponding to the highest classification performance in each experiment (and not on a global average, as the dimension represented by the filled dots in Fig.~\ref{fig:dimLDA}): Table~\ref{tab:dim} illustrates the average dimension for the five angles and varying number of training examples. Two tendencies are observed: the most nadir acquisition ($\theta = 6.09^\circ$) requires the smallest number of dimensions of the latent space to achieve accurate classification. This is not surprising, as this domain is the one providing most of the labeled examples. For the other domains, a larger number of dimensions is necessary, ranging from 12 to 18 for the acquisition at $26.75^\circ$ and higher dimensionality for the other acquisitions, related to larger shift (cf. the lower performance of the LDA baseline in Fig. ~\ref{fig:angres}). The only exception observed is the result for 10 pixels \emph{per} class at $-39.79^\circ$, which shows an unusually small average best dimensionality. This is due to the instability of the best dimensionality along the experiments, most likely due to the small number of pixels per class used to align the $-39.79^\circ$ domain: for two experiments it required three dimensions and for the rest about 30. However, note that this does not impact the classification performance reported in Fig.~\ref{fig:angres}, which is stable around 0.82 in $\kappa$ statistic.}

\begin{table}[!t]
\caption{\label{tab:dim} \blue{Average number of dimensions required by SS--MA to obtain the best classification performance (blue curve in Fig.~\ref{fig:angres})}}
\begin{tabular}{c|c|c|c|c|c}
\hline
& \multicolumn{5}{c}{Optimal dimensionality for $\theta$}\\
& $-38.79^\circ$ & $ -29.16^\circ$ & $ 6.09^\circ$ & $ 26.76^\circ$ & $ 39.5^\circ$ \\\hline
10 \emph{per} class & 10.6 & 34.4 & 3.0 & 18.0 & 32.2 \\
50 \emph{per} class & 25.8 & 24.2 & 6.2 & 11.6 & 23.8 \\
90 \emph{per} class & 25.0 & 31.6 & 3.0 & 11.2 & 21.4 \\
\hline
\end{tabular}
\end{table}

\subsection{Multitemporal adaptation for the same sensor}

\begin{figure*}
\begin{tabular}{ll|c|c|c}
\setlength{\tabcolsep}{1pt}

&& \multicolumn{3}{c}{Image predicted} \\
&& Prilly & Montelly & Malley \\\hline

\multirow{3}{*}{\rotatebox{90}{\hspace{-0.5cm} Leading training image}}&

\rotatebox{90}{Prilly} & 
\includegraphics[width=5.2cm]{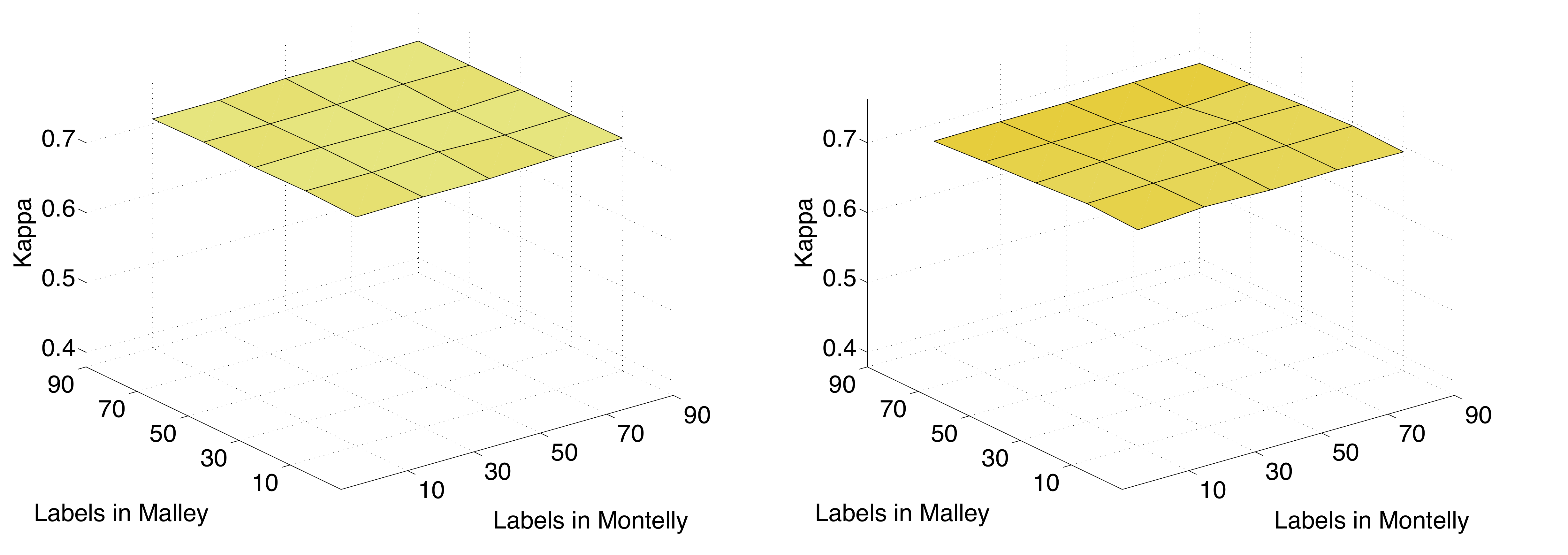}&
\includegraphics[width=5.2cm]{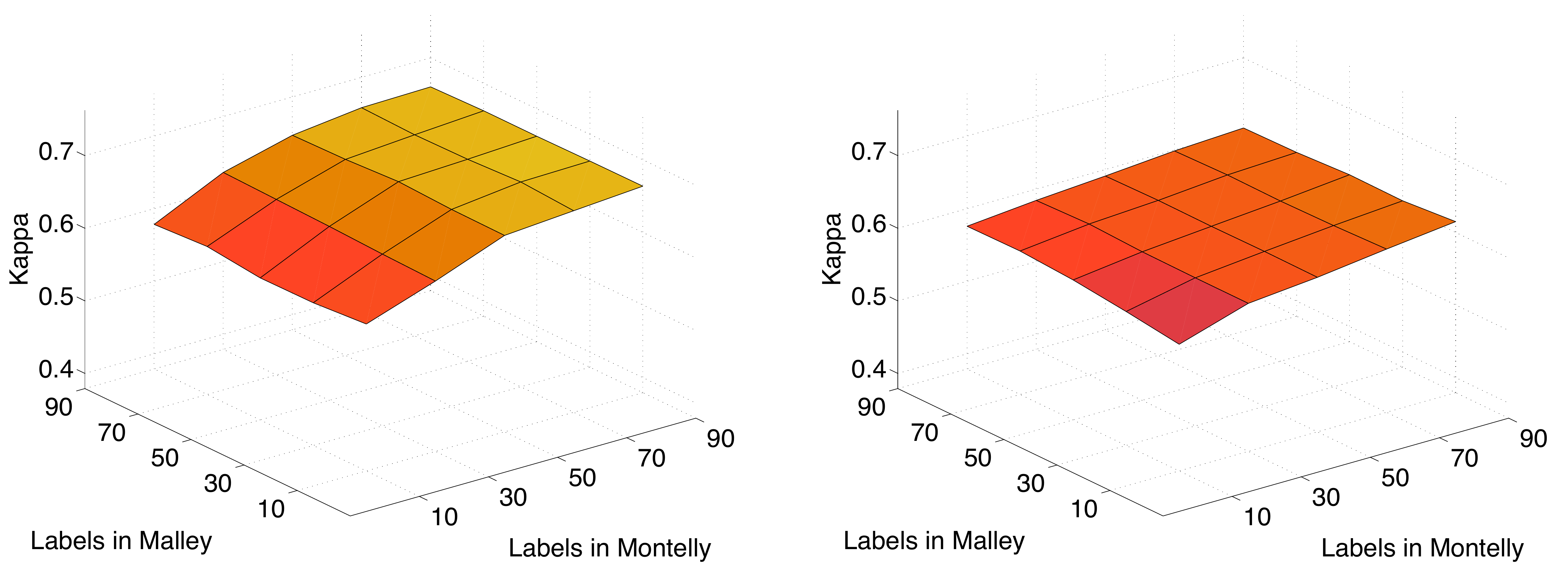} &
\includegraphics[width=5.2cm]{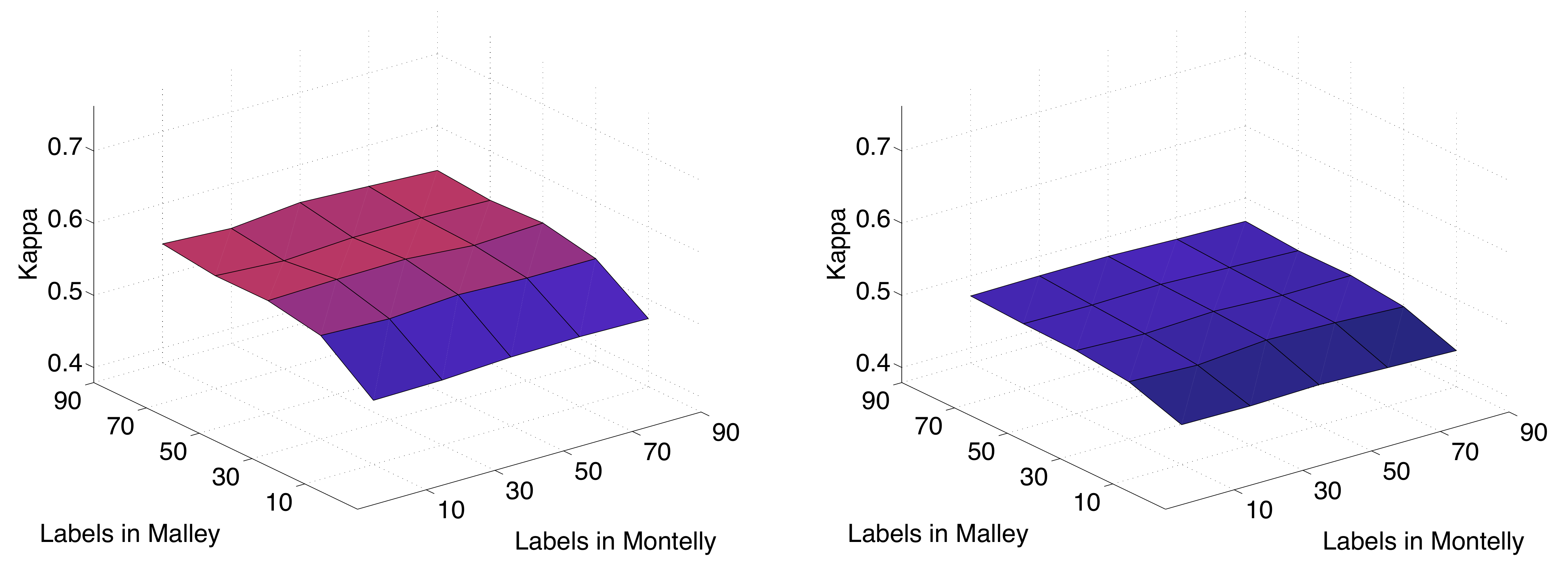}  \\\cline{2-5}

& \rotatebox{90}{Montelly} & 
\includegraphics[width=5.2cm]{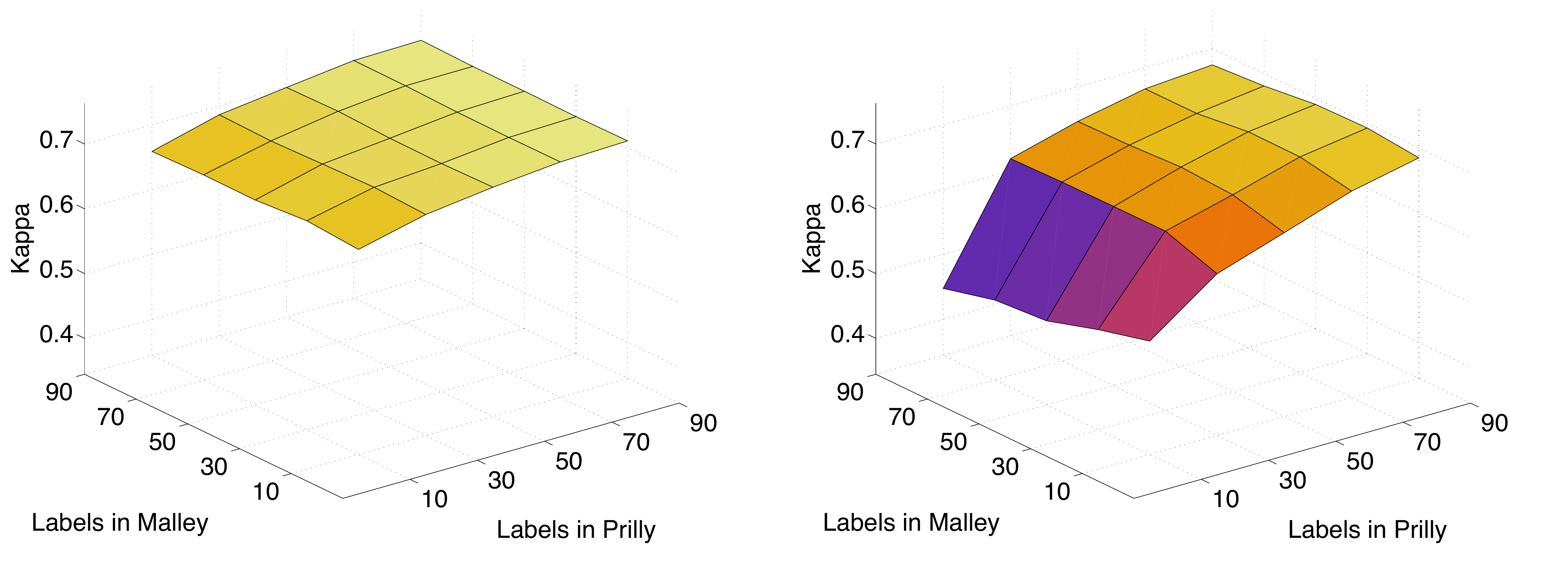}&
\includegraphics[width=5.2cm]{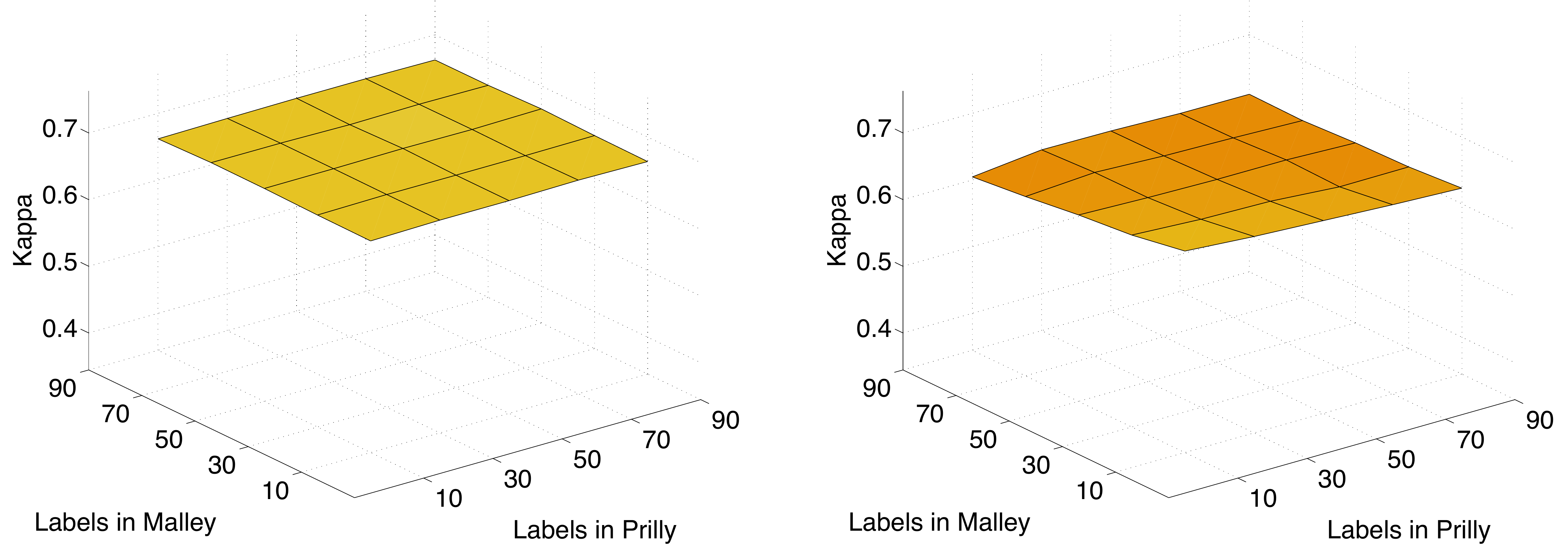}&
\includegraphics[width=5.2cm]{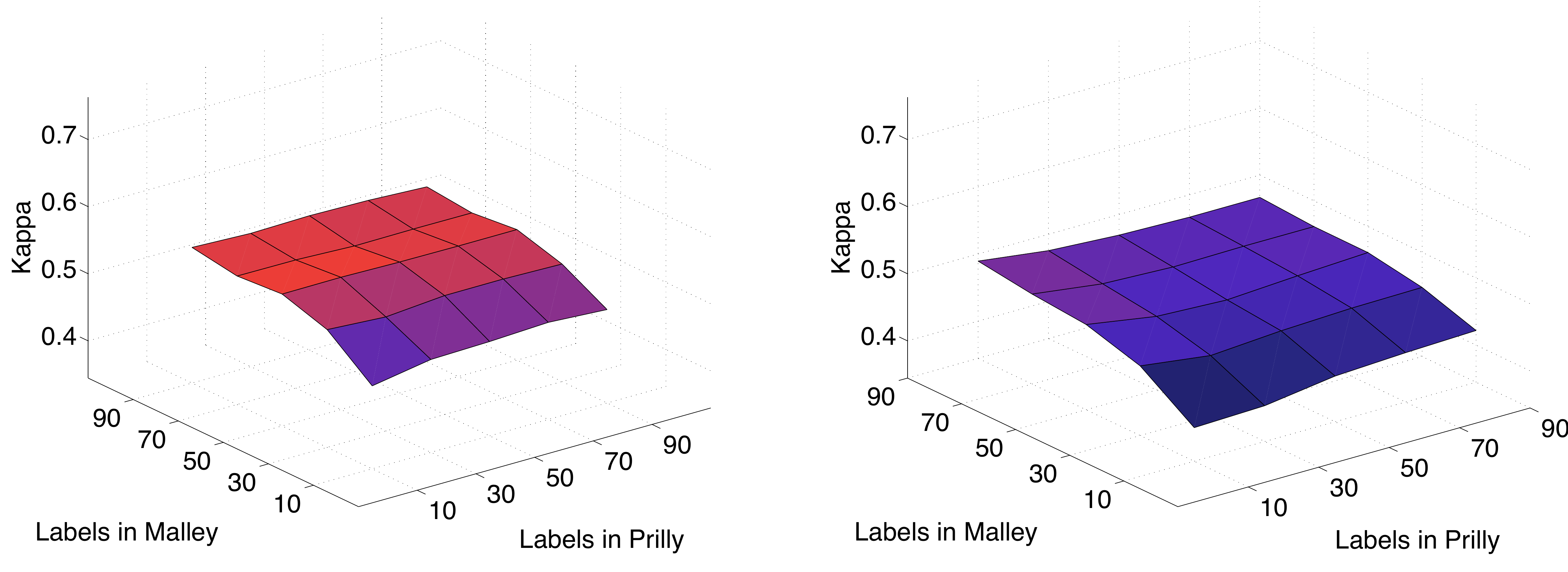}\\\cline{2-5}

& \rotatebox{90}{Malley} &
\includegraphics[width=5.2cm]{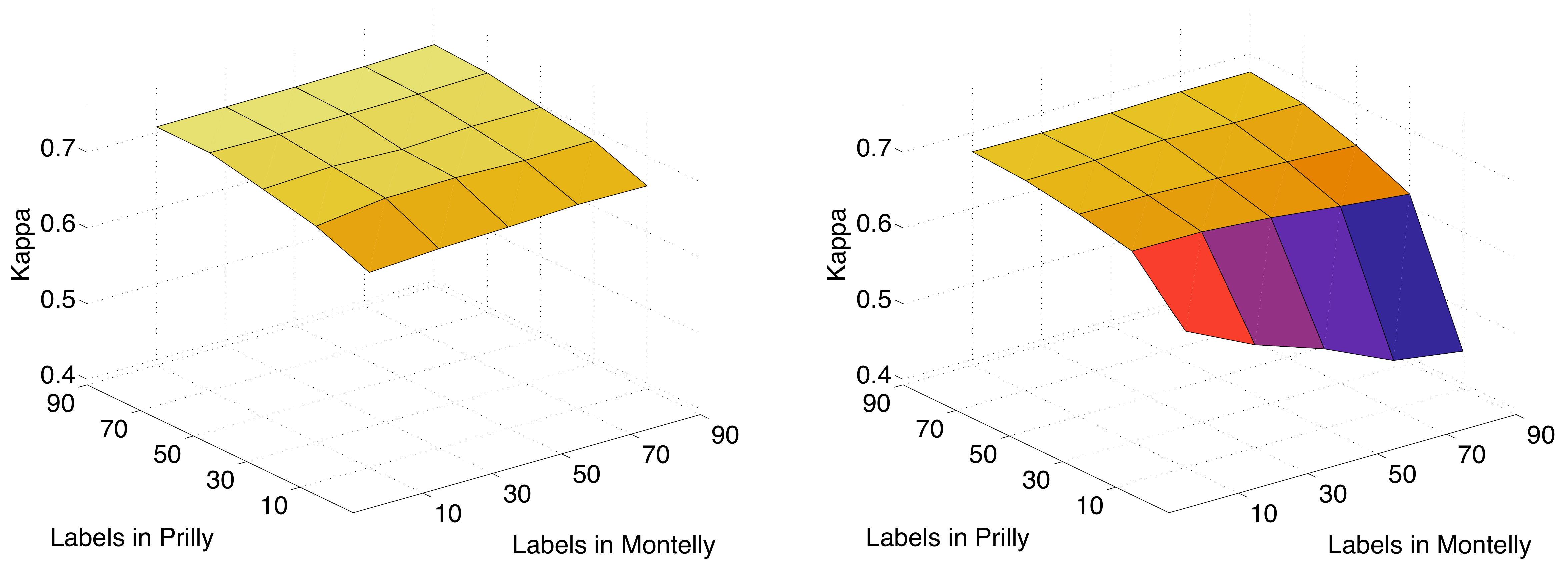}&
\includegraphics[width=5.2cm]{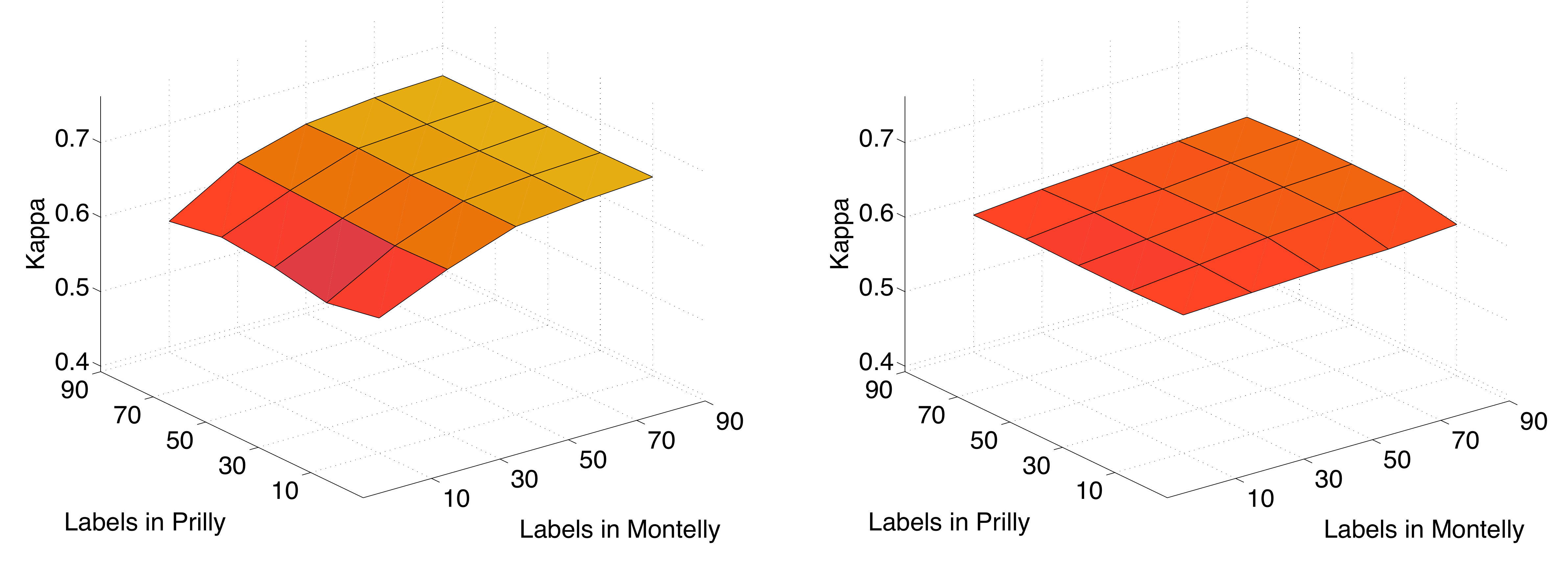}&
\includegraphics[width=5.2cm]{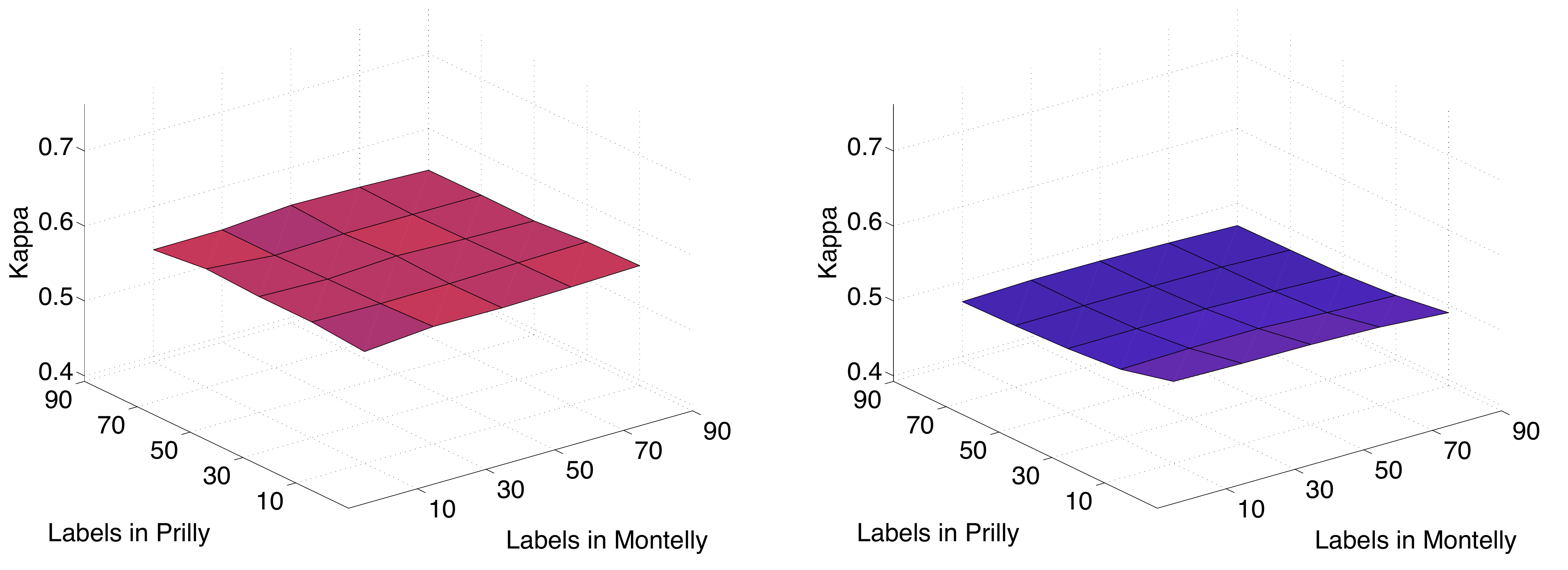}\\\hline

& &SS--MA \white{-------------} Original & SS--MA \white{-------------} Original &SS--MA \white{-------------} Original  \\
                              
\end{tabular}

\caption{Numerical results for the multitemporal experiment. Rows indicate the image from which 100 labeled pixels {\em per} class are used ($l_1 = 100$ {\em per} class). $\kappa$ performance for increasing number of labeled pixels in the two other images ($l_{2,3} = [10, ..., 90]$ {\em per} class) are reported. \blueR{Brighter tones correspond to higher $\kappa$.} Columns correspond to the image that has been used for testing. }
\label{fig:3domainsWV2}

\end{figure*}

\begin{figure*}[!t]
\begin{tabular}{llccc}
\setlength{\tabcolsep}{1pt}

&& \multicolumn{3}{c}{Image predicted} \\
&& Prilly & Montelly & Malley \\
\multirow{3}{*}{\rotatebox{90}{\hspace{-0.5cm} Leading training image}}&

\rotatebox{90}{Prilly} & 
\includegraphics[width=5.2cm]{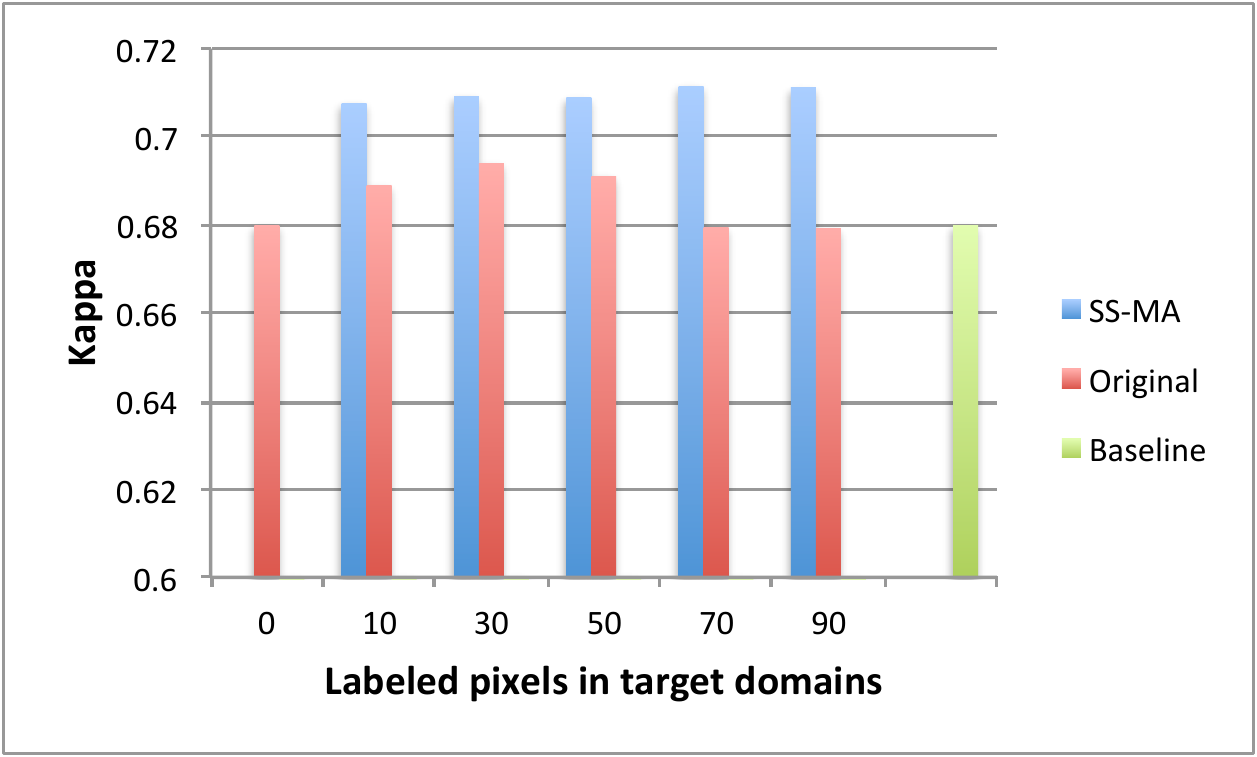}&
\includegraphics[width=5.2cm]{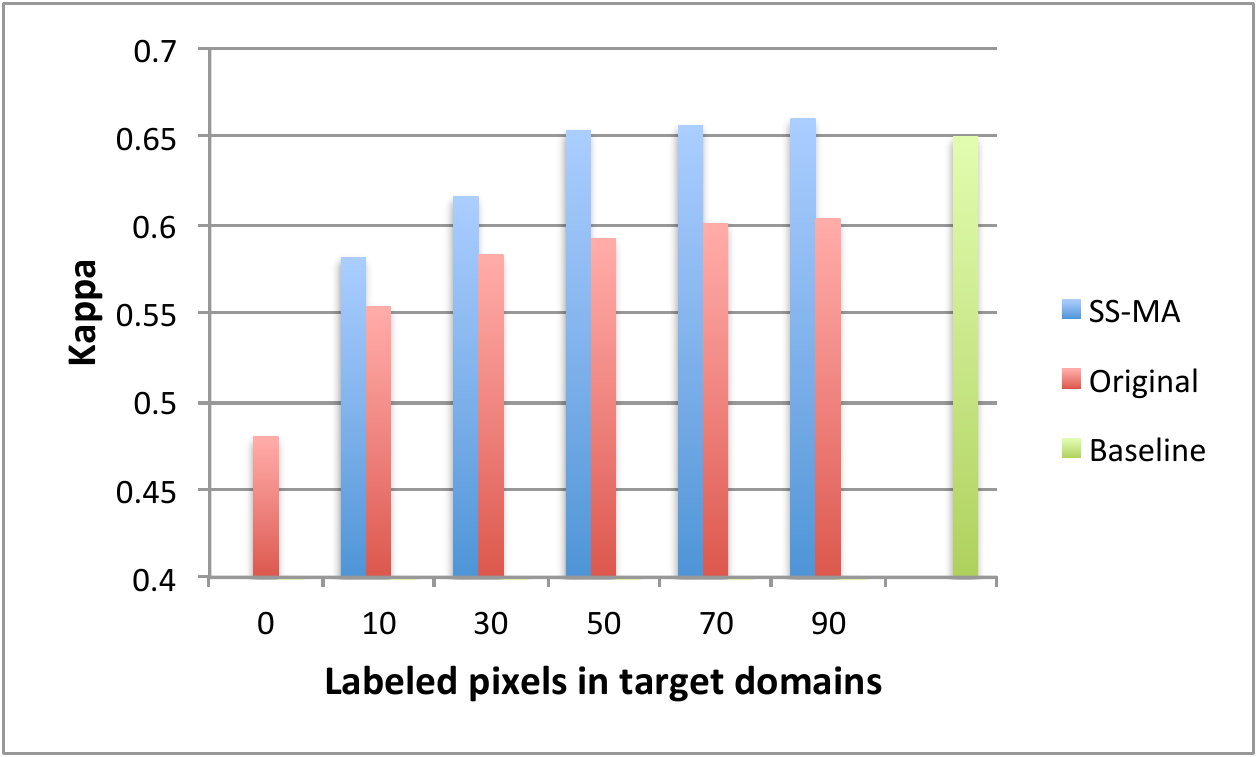} &
\includegraphics[width=5.2cm]{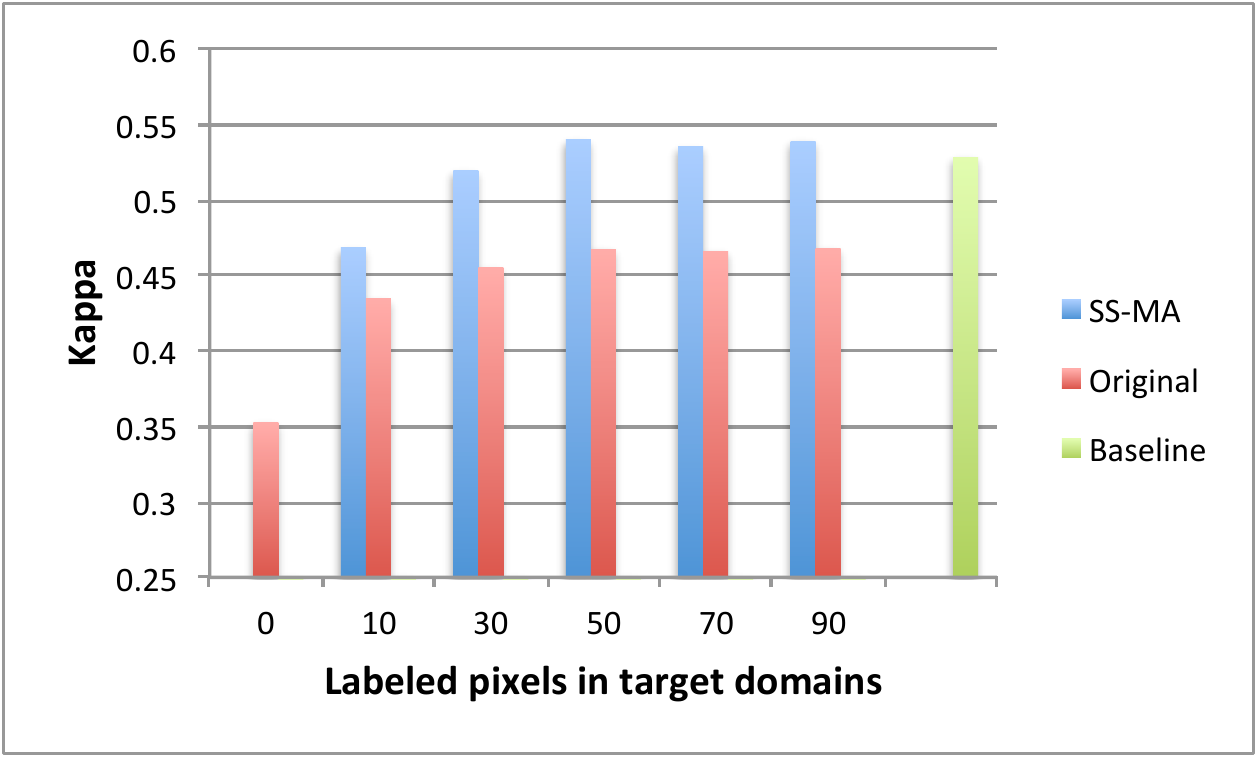}  \\
& \rotatebox{90}{Montelly} & 
\includegraphics[width=5.2cm]{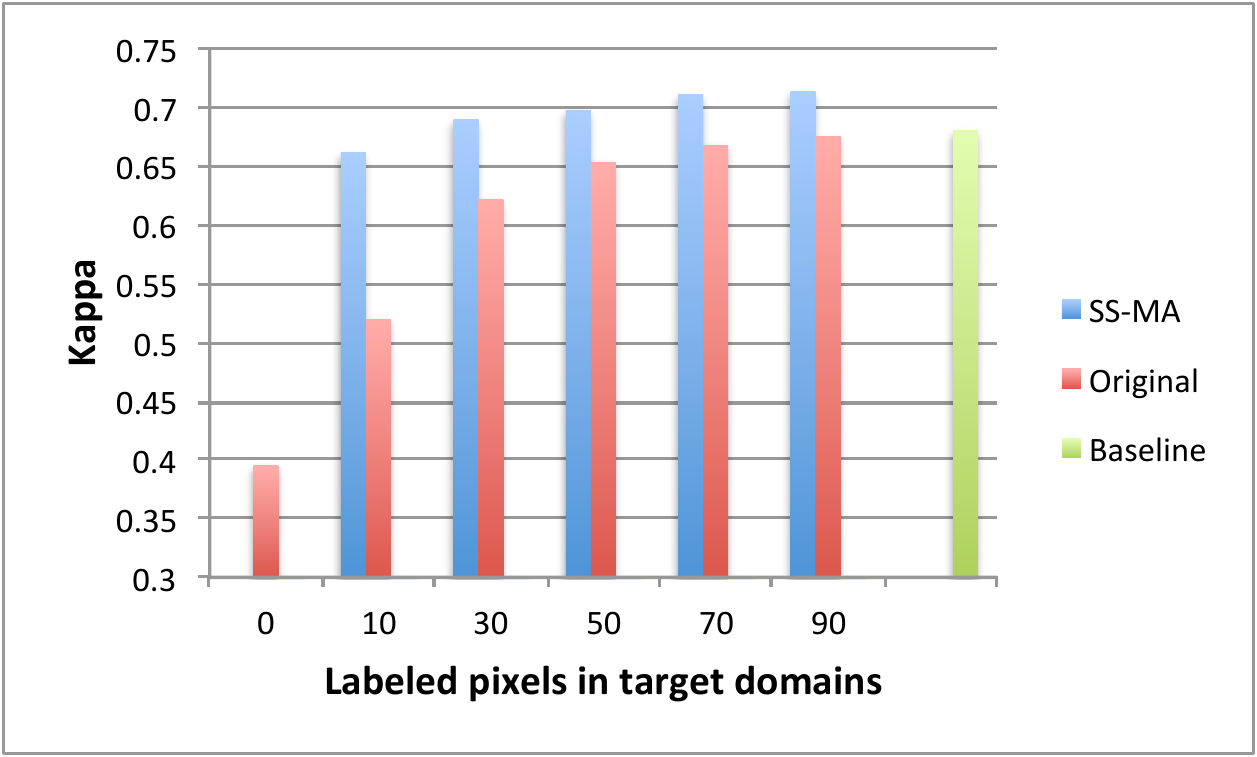}&
\includegraphics[width=5.2cm]{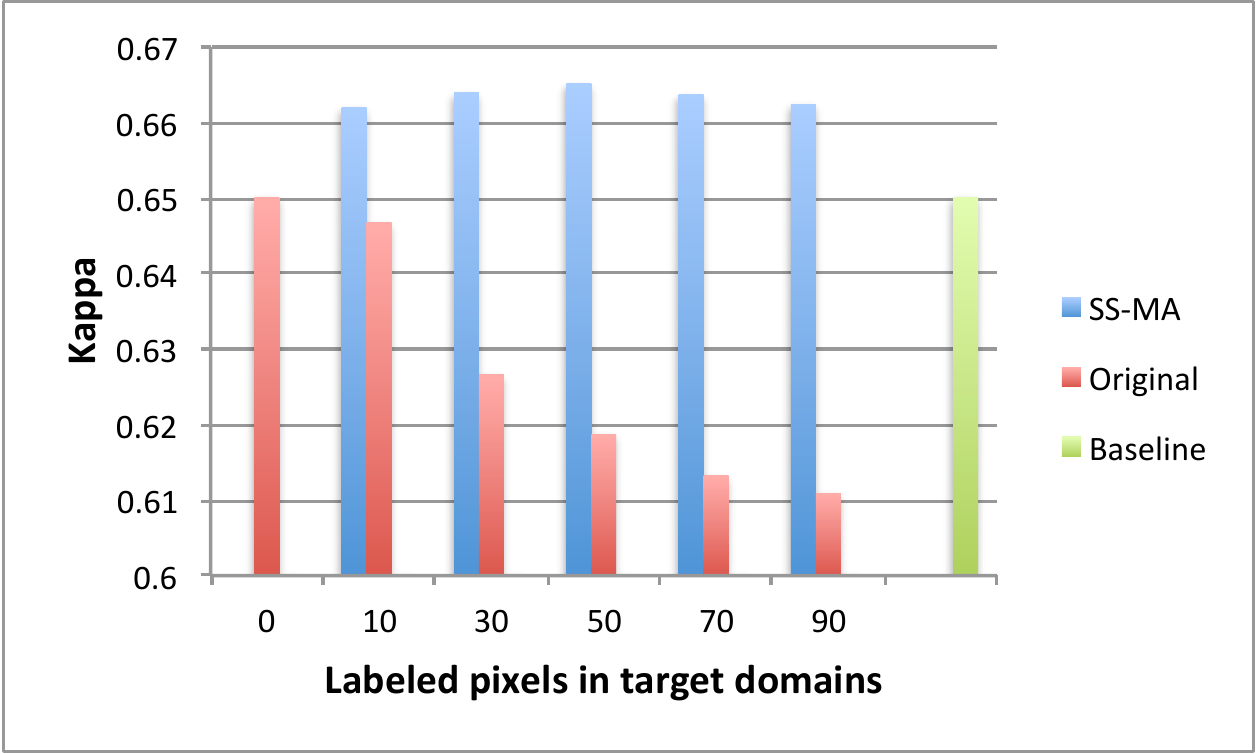}&
\includegraphics[width=5.2cm]{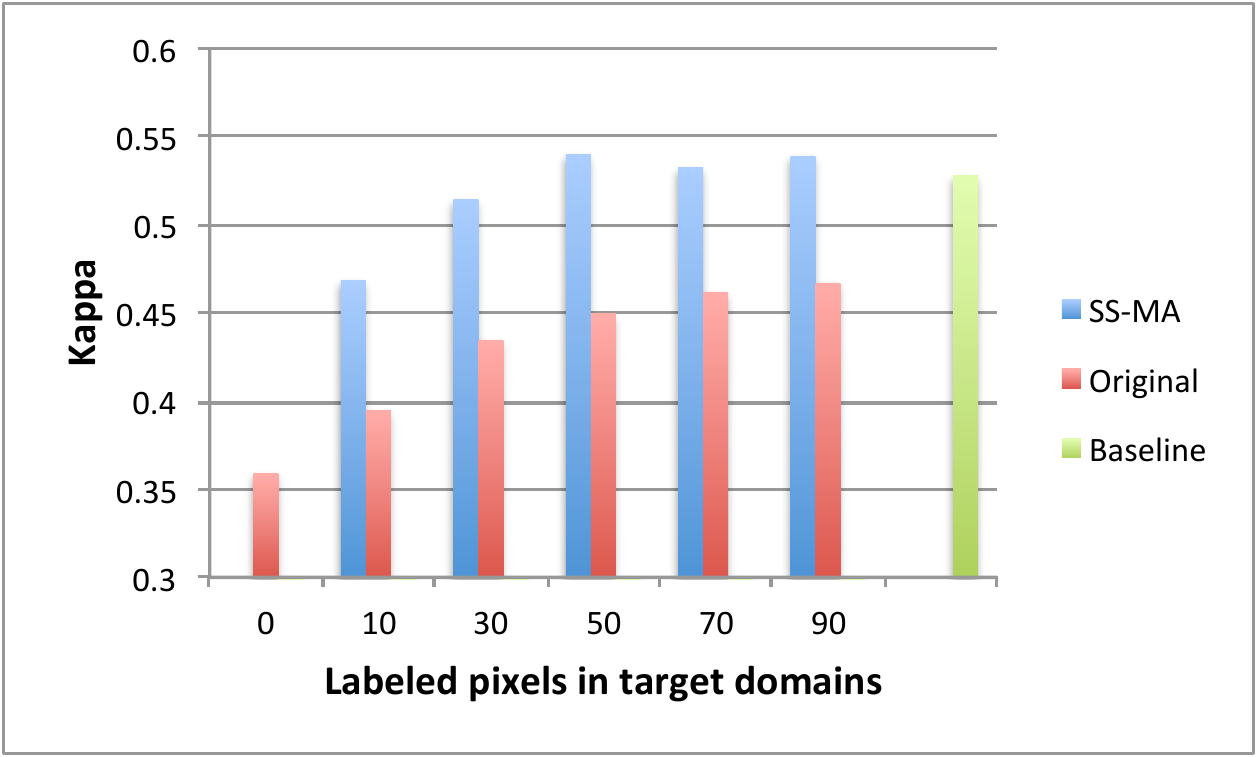}\\

& \rotatebox{90}{Malley} &
\includegraphics[width=5.2cm]{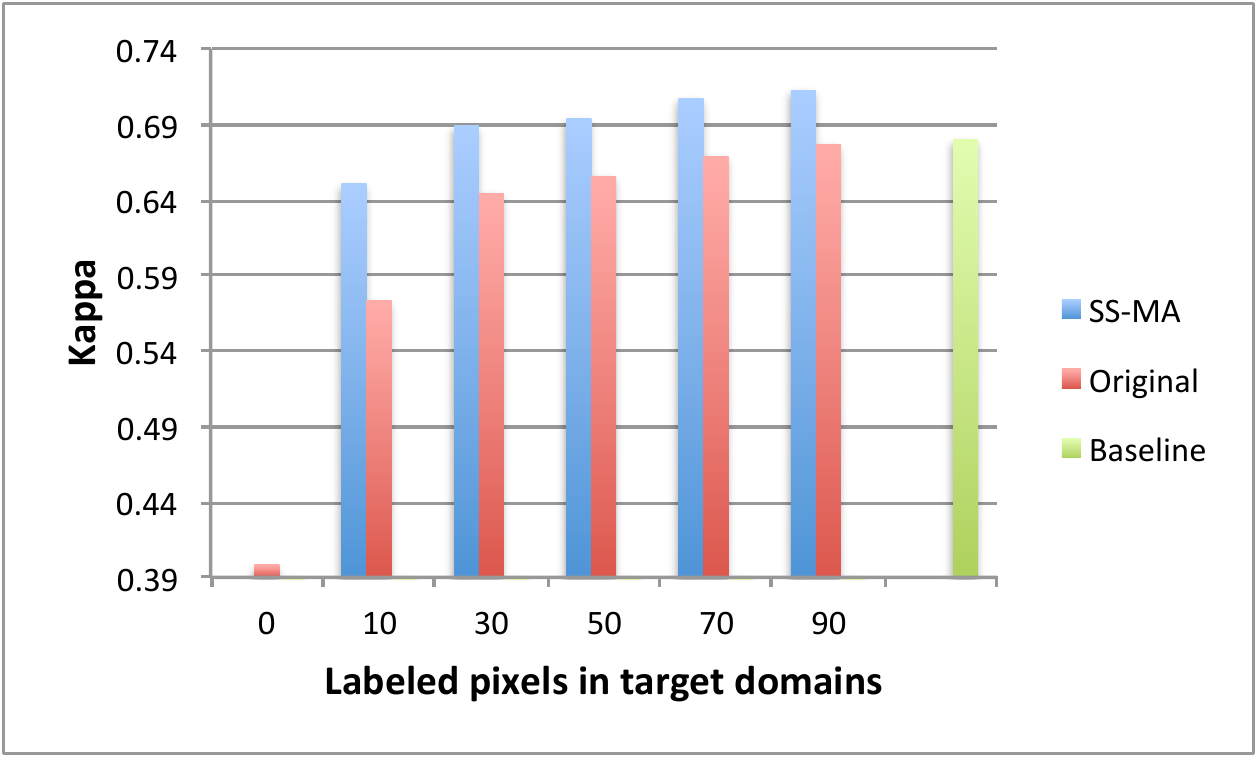}&
\includegraphics[width=5.2cm]{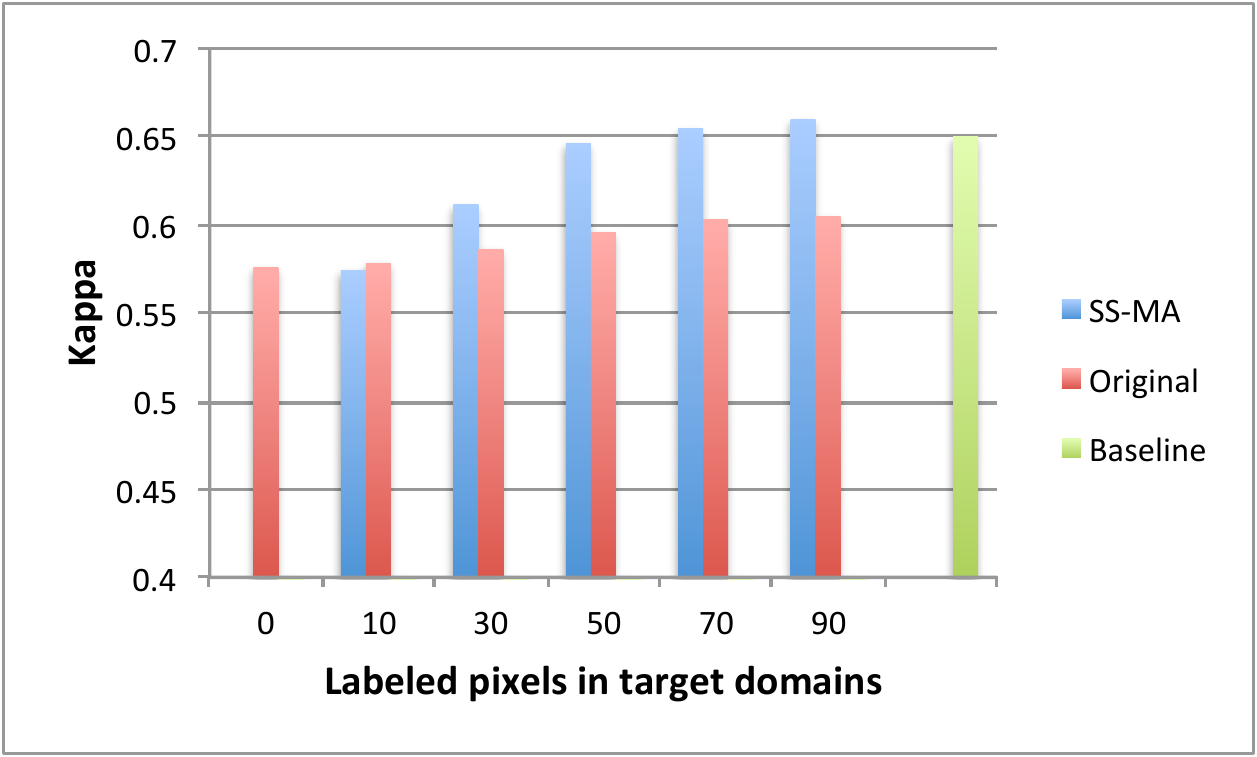}&
\includegraphics[width=5.2cm]{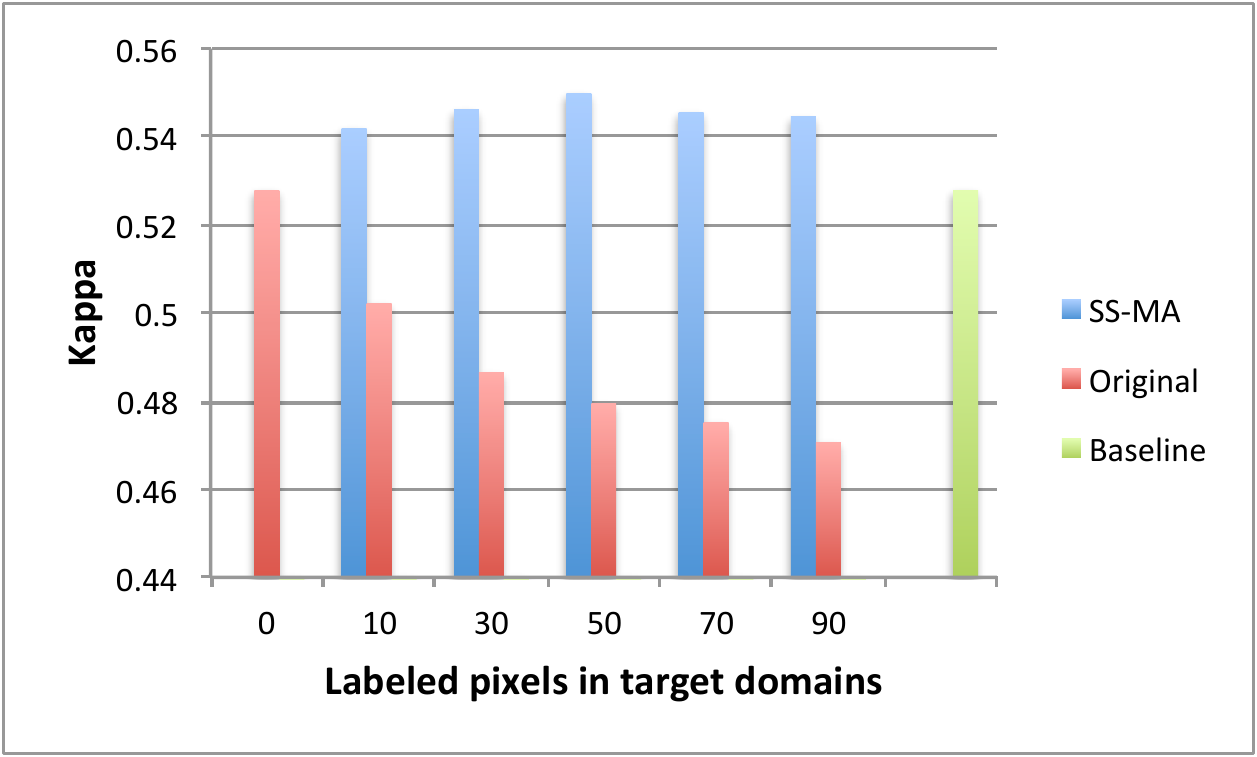}\\
\end{tabular}

\caption{Numerical results for the multitemporal experiment. Rows indicate the image from which 100 labeled pixels {\em per} class are used ($l_1 = 100$ {\em per} class). $\kappa$ performances for increasing number of labeled pixels in the testing images ($l_{2} = l_3 = [10, ..., 90]$ {\em per} class) are reported. The results correspond to the diagonal of the \blue{surfaces} in Fig.~\ref{fig:3domainsWV2}. Columns correspond to the image that has been used for testing. The baseline is the model obtained using 100 pixels {\em per} class from that image only.}
\label{fig:3domainsWV2diag}
\end{figure*}

Figures~\ref{fig:3domainsWV2} and~\ref{fig:3domainsWV2diag} illustrate the numerical results for the multitemporal experiments involving three World-View 2 scenes. The first figure shows the $\kappa$ surfaces for increasing number of labeled samples in the auxiliary domains $\X^2$ and $\X^3$ for both the original data and the data aligned with the SS--MA method. The leading domain $\X^1$ is always sampled with $l_1 = 100$ labeled samples {\em per} class. Figure~\ref{fig:3domainsWV2diag} summarizes the former by illustrating the diagonal of the surfaces, and compares it to the baseline case (in green), where only the tested domain is used for training the model. This baseline is what would happen in a traditional supervised classification study.

In both cases, each row corresponds to the leading training domain $\X^1$ and each column correspond to a different testing image (the baseline is thus the same column-wise). We remind that in  the `SS--MA' case, one set of projections {\em per} combination of labeled samples in the three domains is extracted and all the images are then classified with the same SVM model, trained on the labeled pixels of the three domains. The three blocks of \blue{each row in the figures} (one {\em per} testing image) are issued from the same model. \blueRR{}{The same holds for the `original' experiments: a common model is trained on the labeled pixels in the original space and then each image is predicted separately.}

For all tested scenarios, the proposed SS--MA leads to higher $\kappa$ scores than the method without alignment. In general, the inclusion of labeled pixels from the testing domain (e.g., from the Prilly scene, when testing on the Prilly scene) is beneficial, so that the ideal case is  met when having labeled samples from all domains. Moreover, some interesting observations can be done:
\begin{itemize}
\item[-] When predicting the same image as the leading training one (diagonal blocks of Figs.~\ref{fig:3domainsWV2} and~\ref{fig:3domainsWV2diag}), adding pixels from the other images results in a decrease of the performance \blue{if no alignment is performed. This is}  particularly visible in the Montelly and Malley cases, \blueRR{}{where the performance of the `original' model (the red bars in the diagonal panels of Fig.~\ref{fig:3domainsWV2diag}) decrease with the increase of labeled samples from the other domains}. On the contrary, when aligning the datasets with the proposed SS--MA, the $\kappa$ performance remains stable at a level generally higher than the baseline \blueRR{}{(blue bars in the same panels of Fig.~\ref{fig:3domainsWV2diag})}. Since it aligns domains in a discriminative sense,  SS--MA  allows to use many domains without affecting the quality of the classifier in the original domain. 
\item[-] When predicting other images than the one with a large amount of labeled samples (the off-diagonal blocks of Figs.~\ref{fig:3domainsWV2} and~\ref{fig:3domainsWV2diag}) both the original and SS--MA classification lead to an improvement of the results of the case where only pixels from the leading domain are used (visible as the first bar of the histograms in Fig.~\ref{fig:3domainsWV2diag}. The alignment procedure boosts the performance of the classifier and leads to $\kappa$ scores that are between 0.1 and 0.3 $\kappa$ points higher with respect to the case, where only pixels from the leading domain are used for training.
\item[-] In all cases, SS--MA leads to more drastic improvements than the non-projected method and can meet the baseline performance in all the cases. This shows that, when using the SS--MA projections, adding information from other domains improves the classification performances, provided the higher number of labeled samples and the discriminative nature of the projection functions. 
\end{itemize}

\begin{figure*}[!t]
\begin{tabular}{ll|c|c|c}
\setlength{\tabcolsep}{1pt}

&& \multicolumn{3}{c}{Image predicted} \\
&& Prilly & Montelly & Zurich (4 bands) \\\hline

\multirow{3}{*}{\rotatebox{90}{\hspace{-0.5cm} Leading training image}}&

\rotatebox{90}{Prilly} & 
\includegraphics[width=5.2cm]{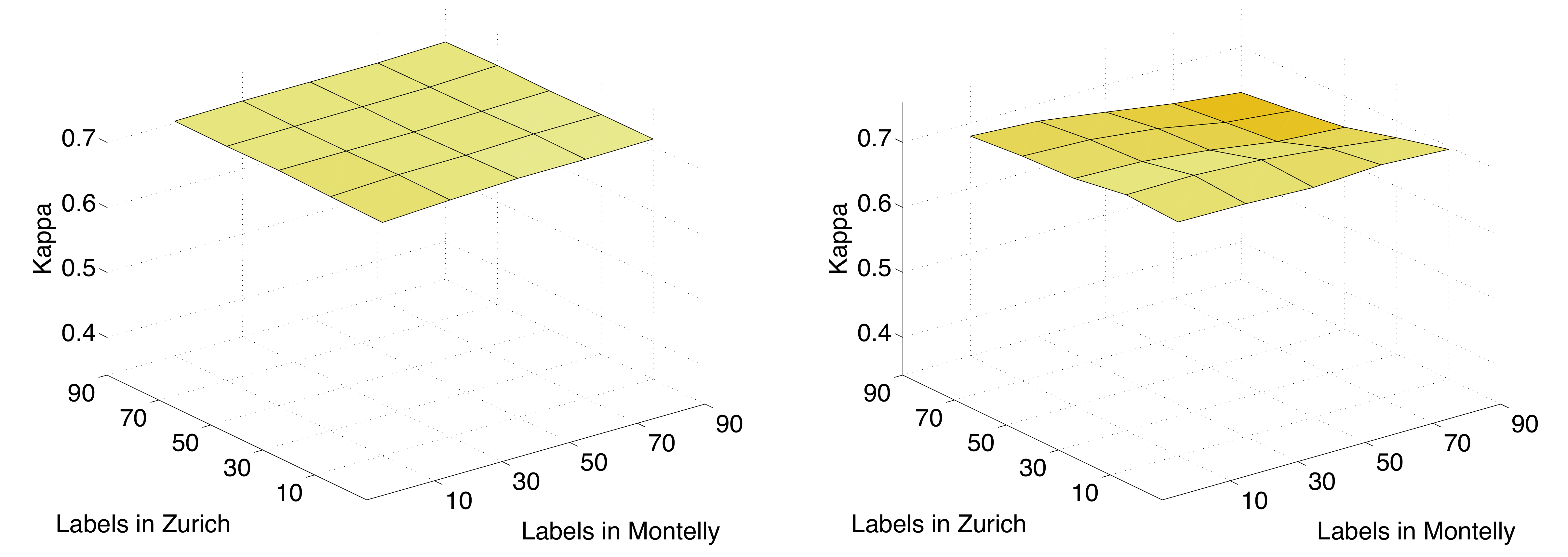}&
\includegraphics[width=5.2cm]{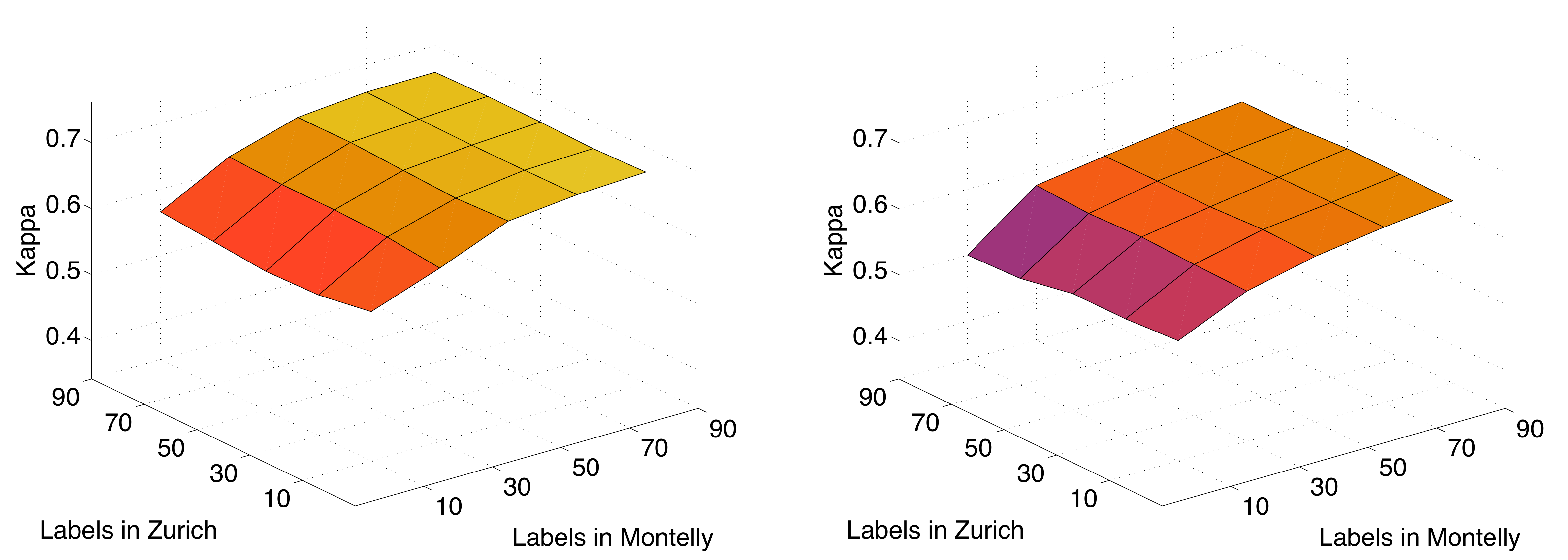} &
\includegraphics[width=5.2cm]{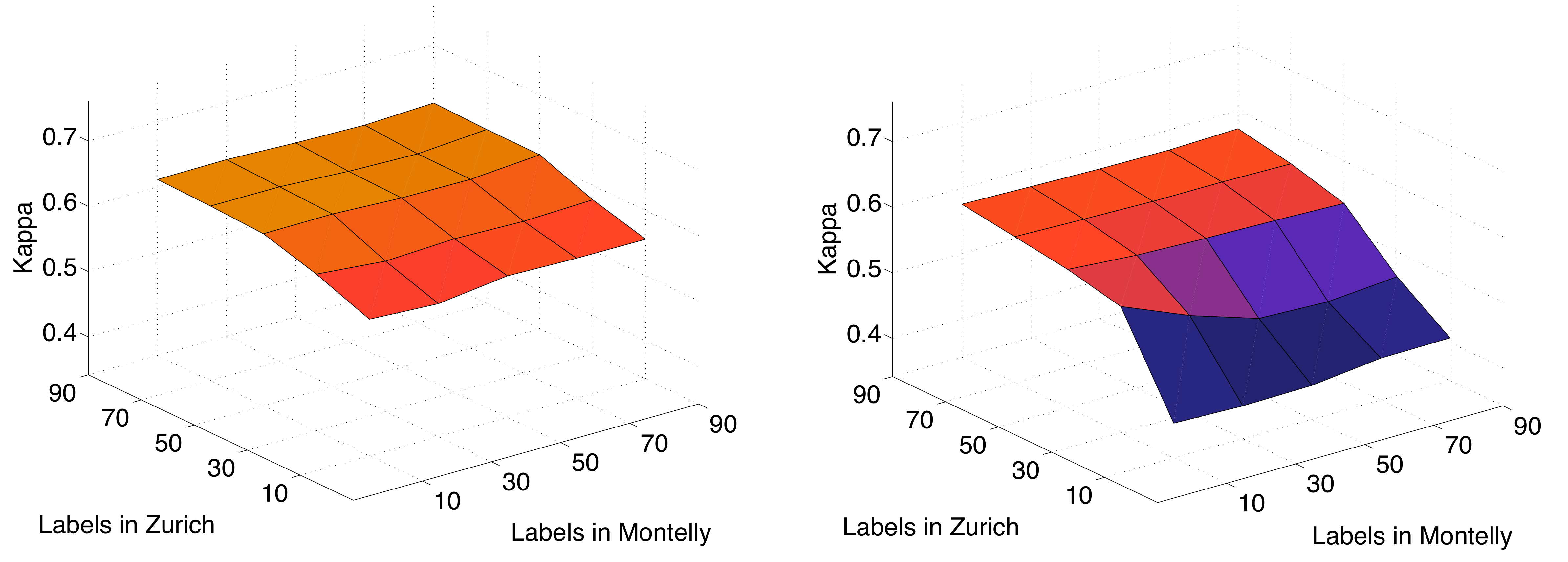}  \\\cline{2-5}

& \rotatebox{90}{Montelly} & 
\includegraphics[width=5.2cm]{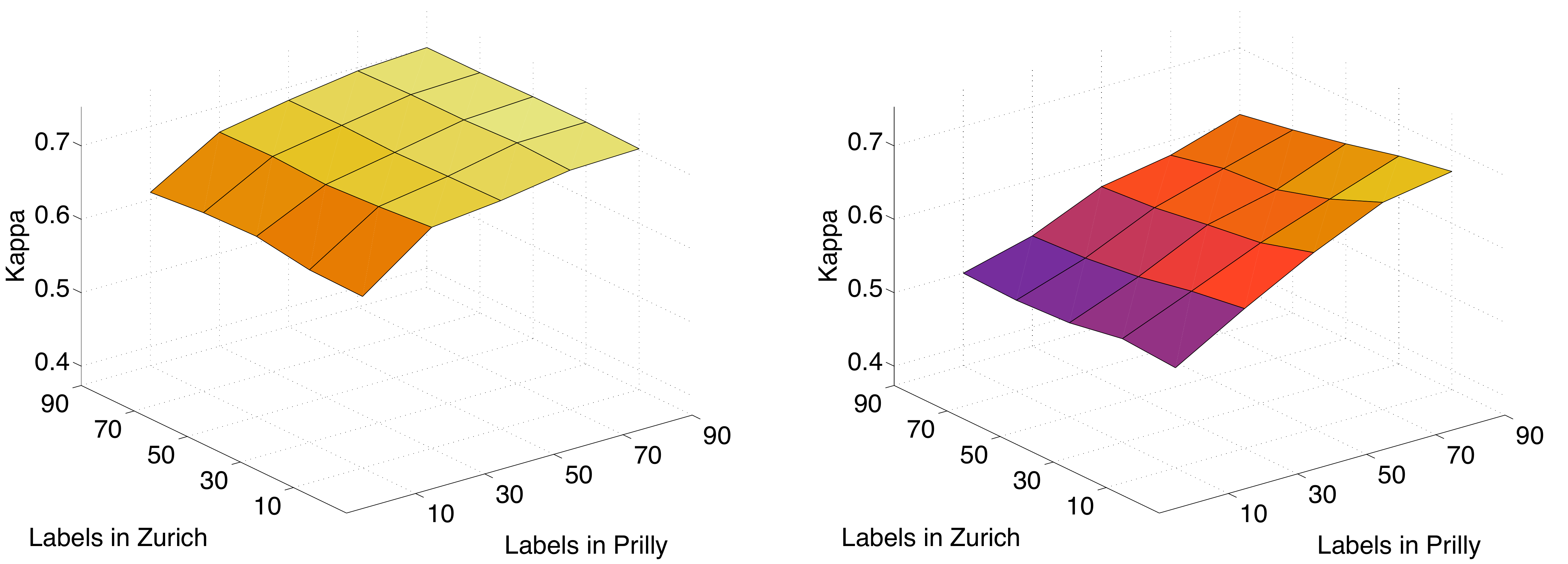}&
\includegraphics[width=5.2cm]{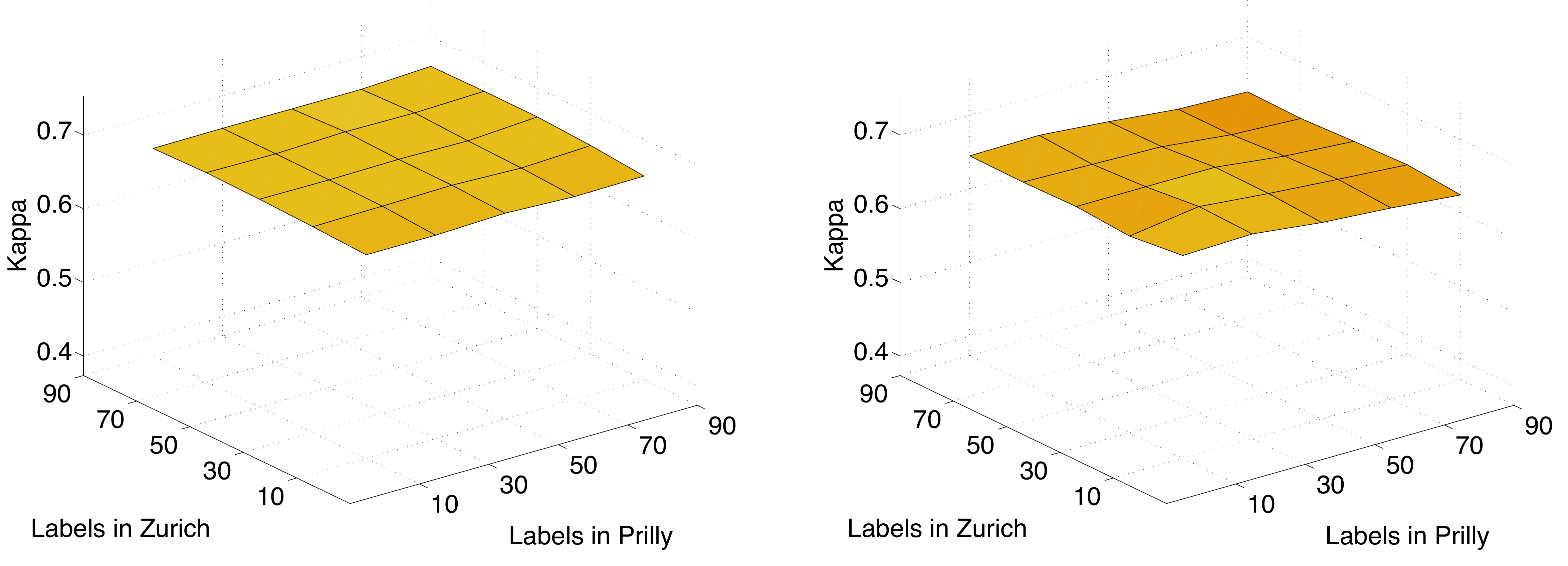}&
\includegraphics[width=5.2cm]{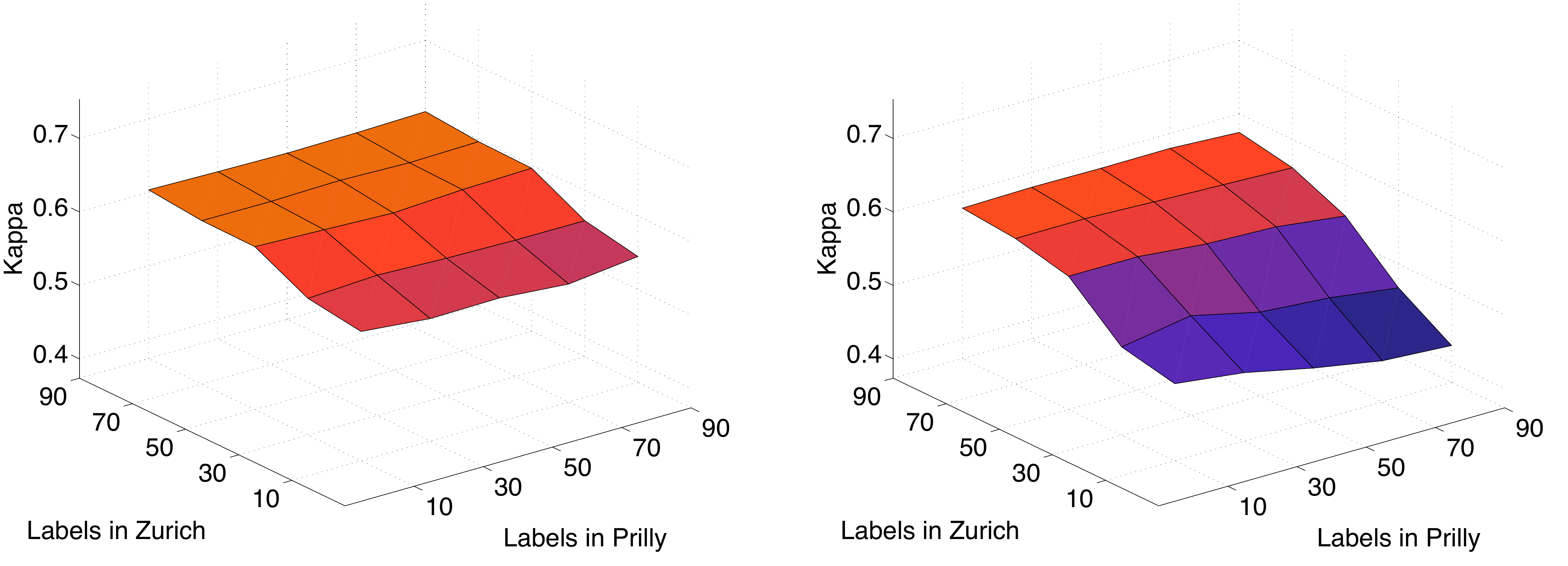}\\\cline{2-5}

& \rotatebox{90}{Zurich (4 bands)} &
\includegraphics[width=5.2cm]{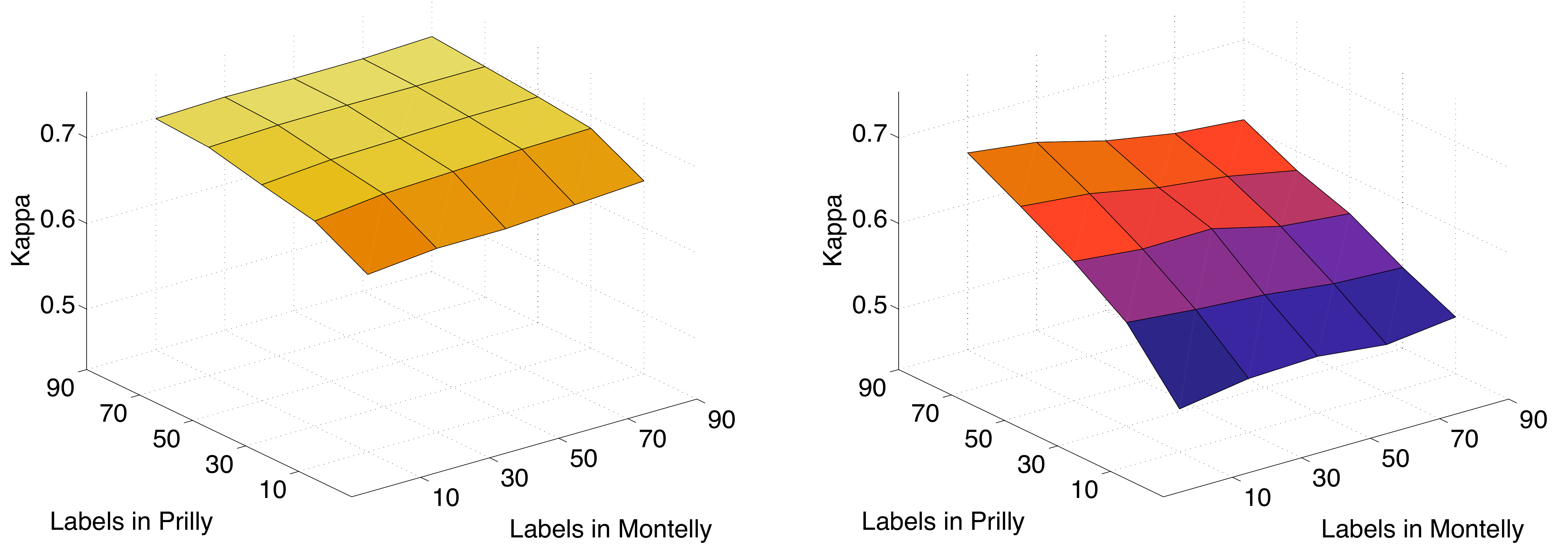}&
\includegraphics[width=5.2cm]{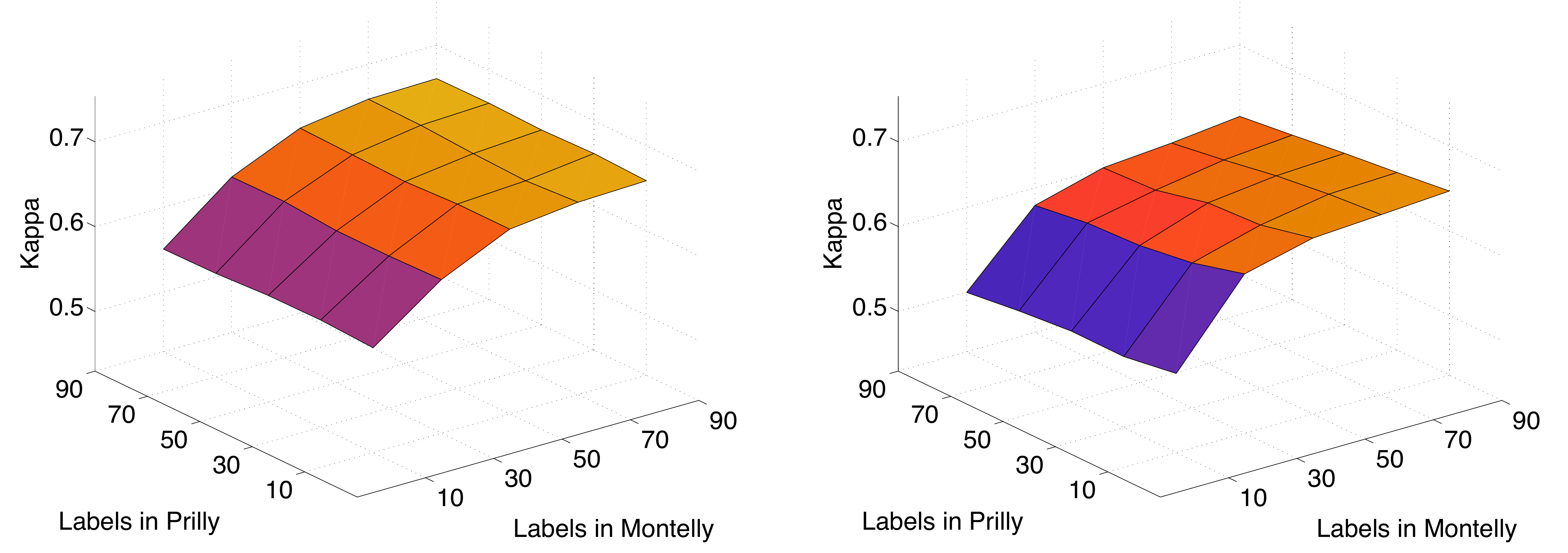}&
\includegraphics[width=5.2cm]{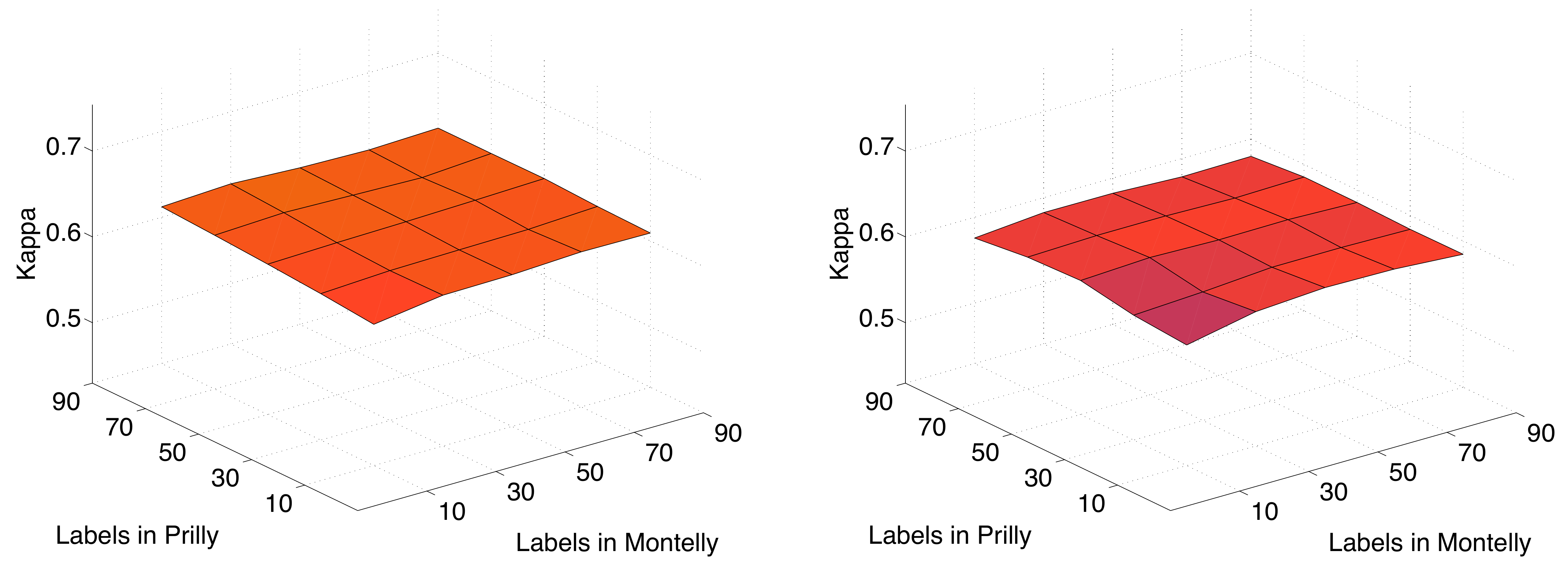}\\\hline
& &SS--MA \white{-------------} Original & SS--MA \white{-------------} Original &SS--MA \white{-------------} Original  \\
\end{tabular}

\caption{Numerical results for the multisource experiment. Rows indicate the image from which 100 labeled pixels {\em per} class are used ($l_1 = 100$ {\em per} class). $\kappa$ performances for increasing number of labeled pixels in the two other images ($l_{2,3} = [10, ..., 90]$ {\em per} class) are reported. \blueR{Brighter tones correspond to higher $\kappa$.} Columns correspond to the image that has been used for testing. }
\label{fig:3domainsWVQB}

\end{figure*}

\subsection{Multisource adaptation}

\begin{figure*}[!t]
\begin{tabular}{llccc}
\setlength{\tabcolsep}{1pt}

&& \multicolumn{3}{c}{Image predicted} \\
&& Prilly & Montelly & Zurich (4 bands) \\
\multirow{3}{*}{\rotatebox{90}{\hspace{-0.5cm} Leading training image}}&

\rotatebox{90}{Prilly} & 
\includegraphics[width=5.2cm]{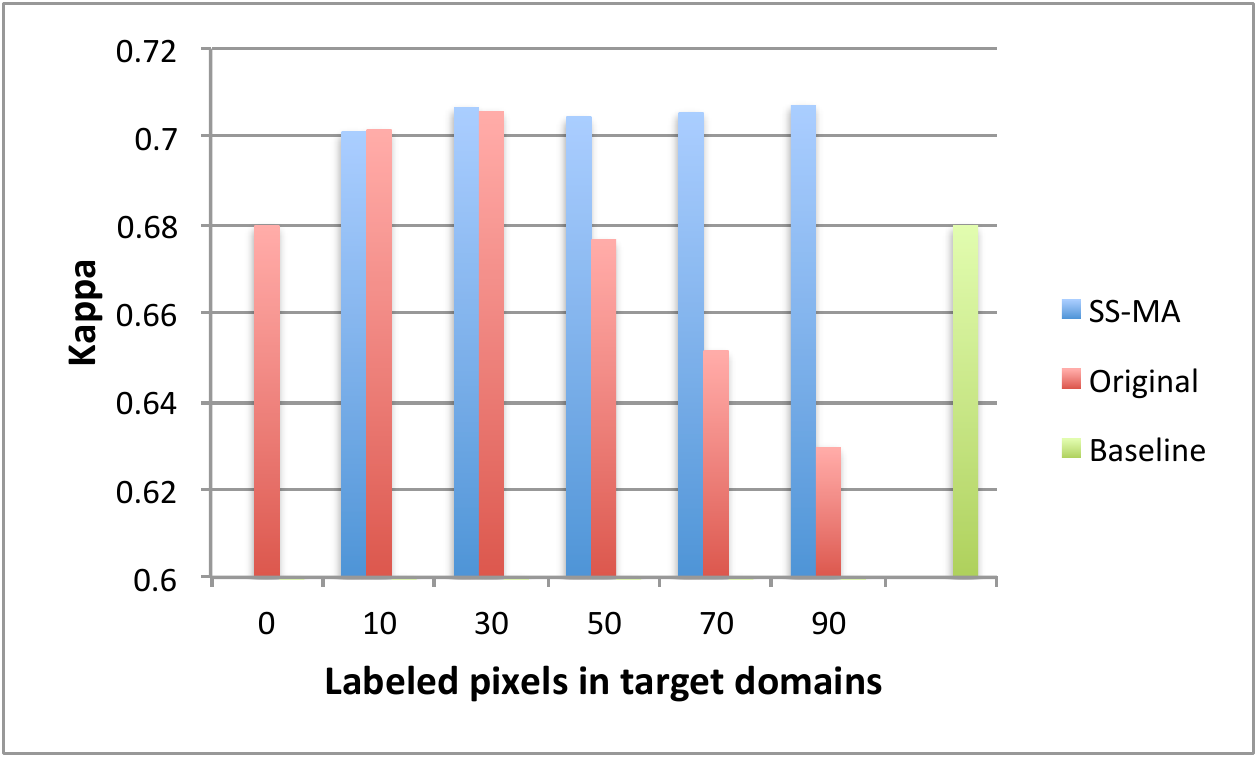}&
\includegraphics[width=5.2cm]{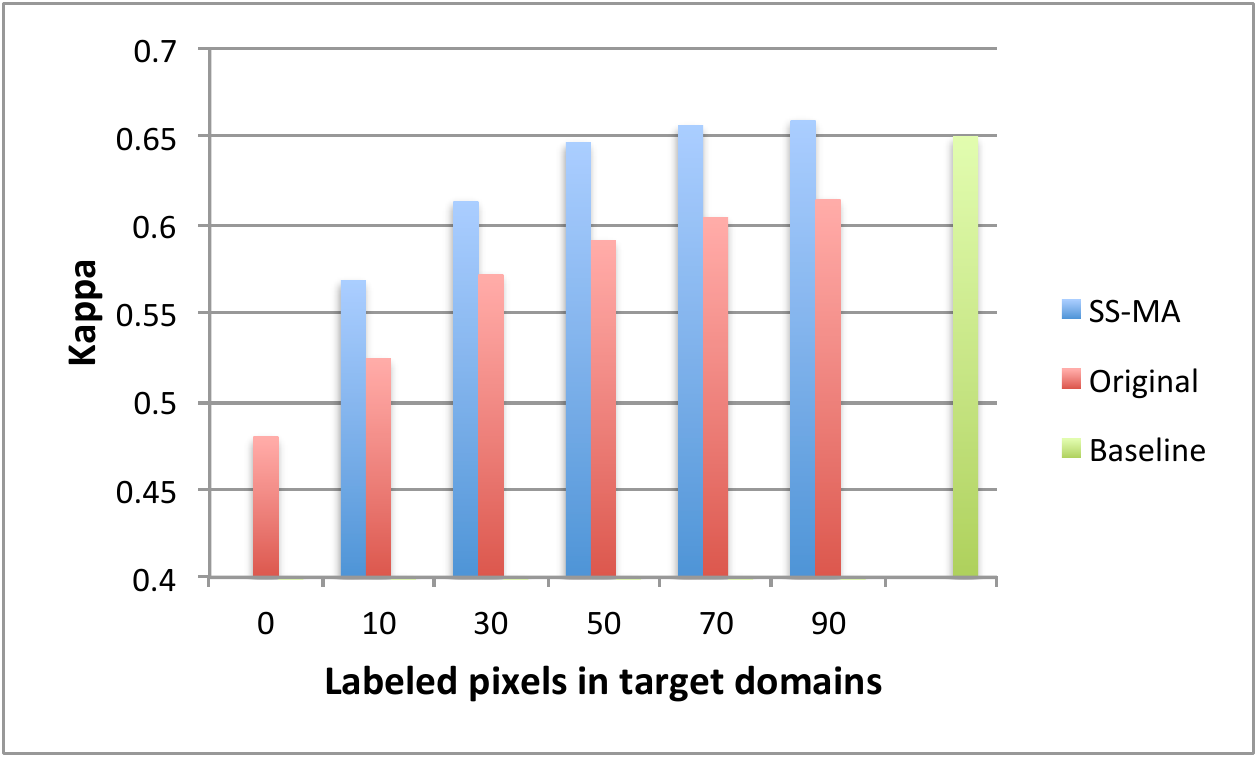} &
\includegraphics[width=5.2cm]{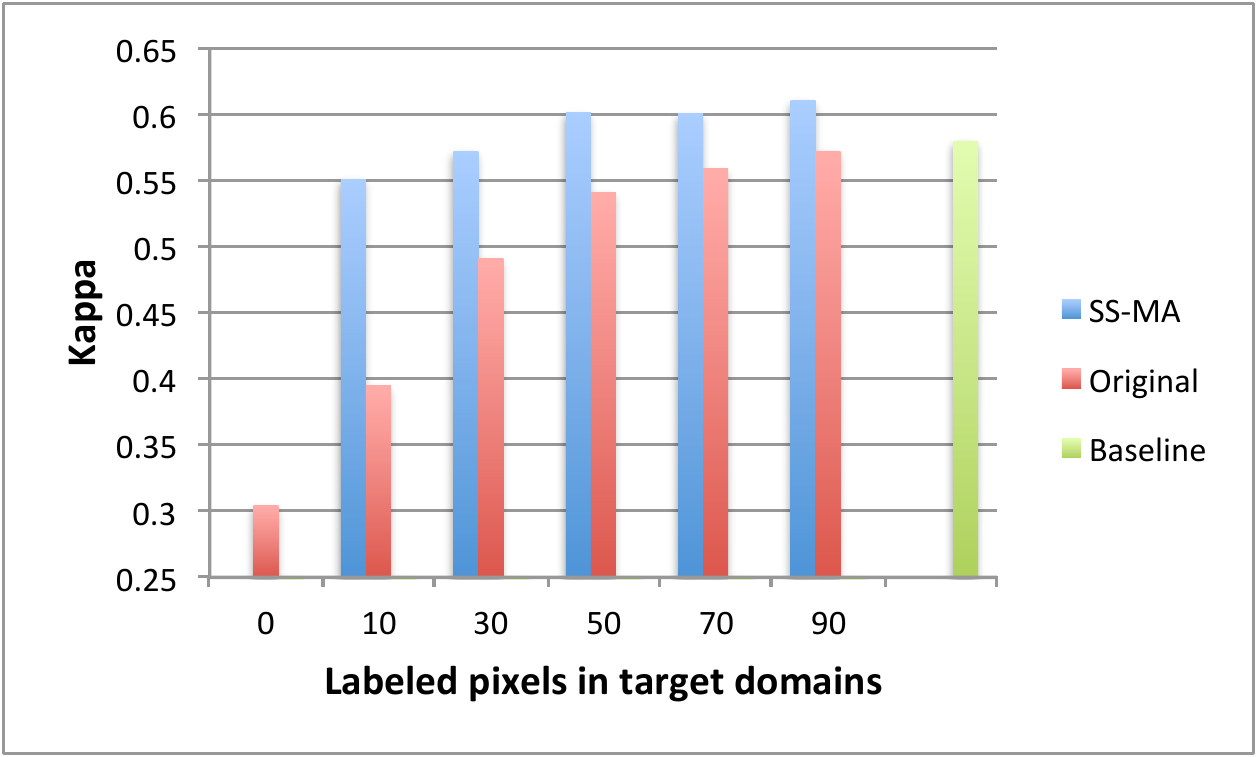}  \\
& \rotatebox{90}{Montelly} & 
\includegraphics[width=5.2cm]{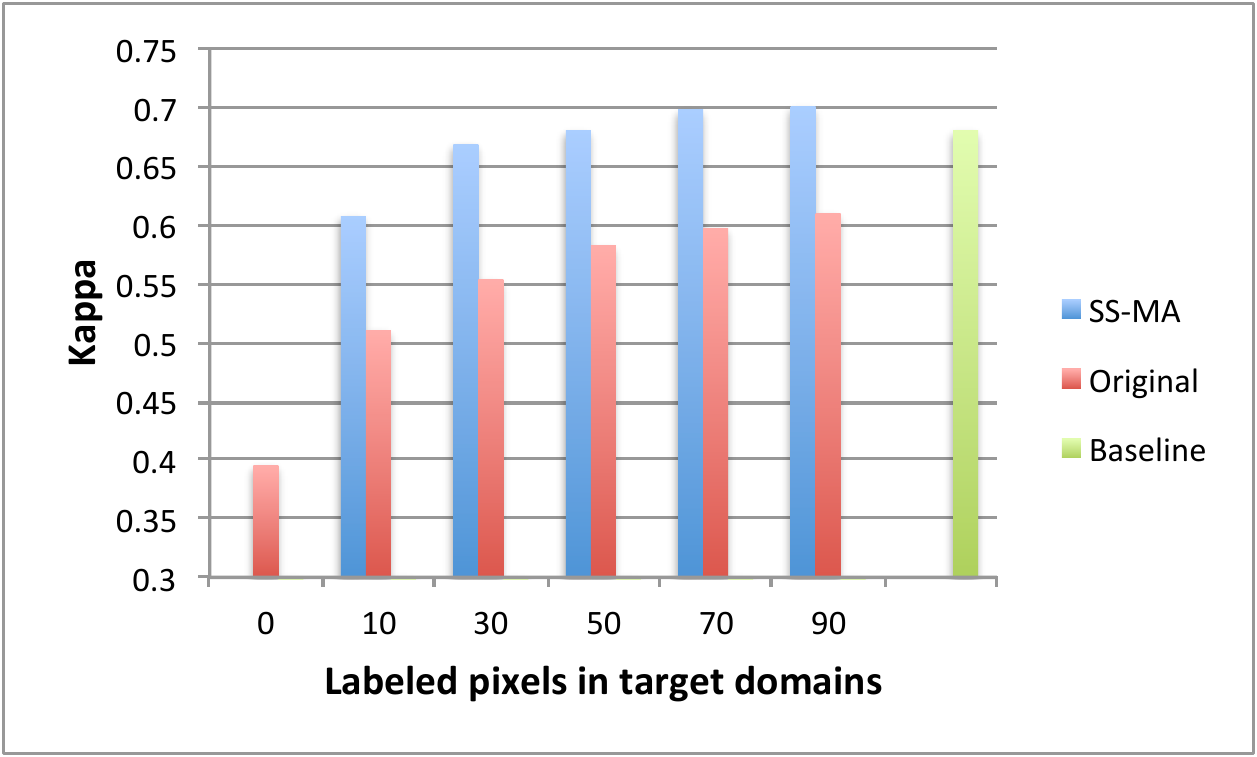}&
\includegraphics[width=5.2cm]{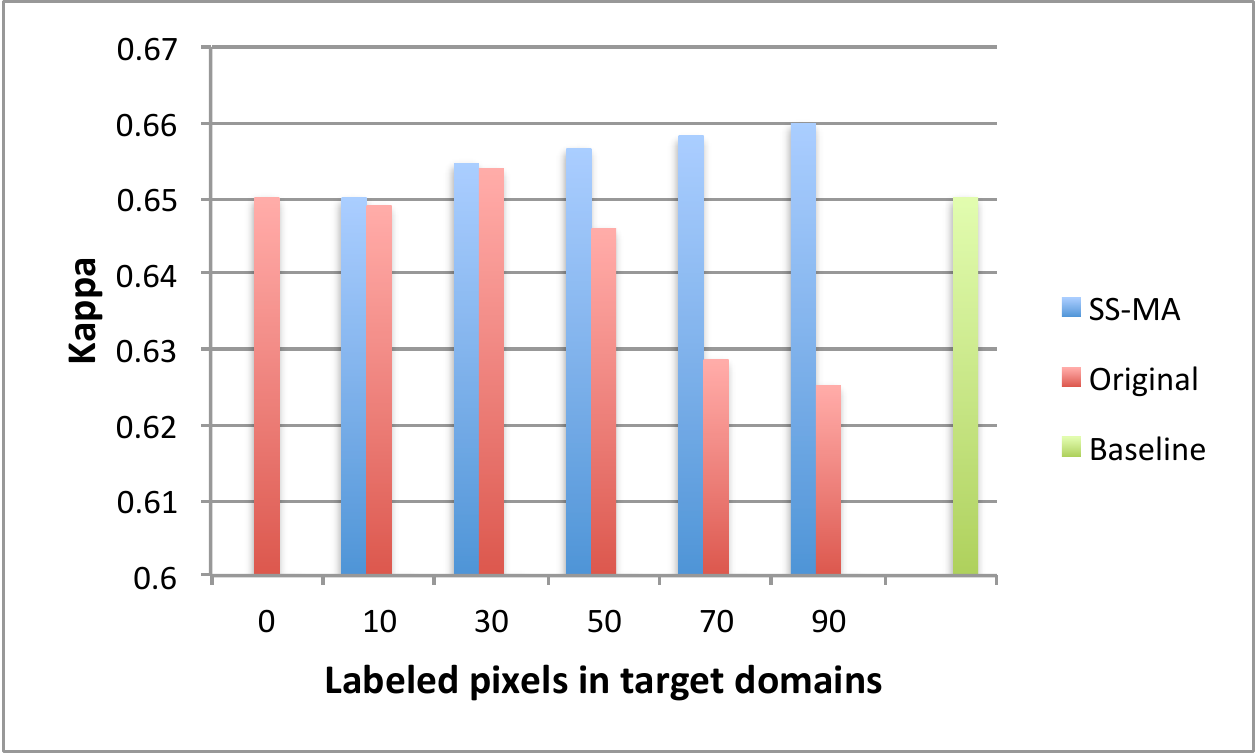}&
\includegraphics[width=5.2cm]{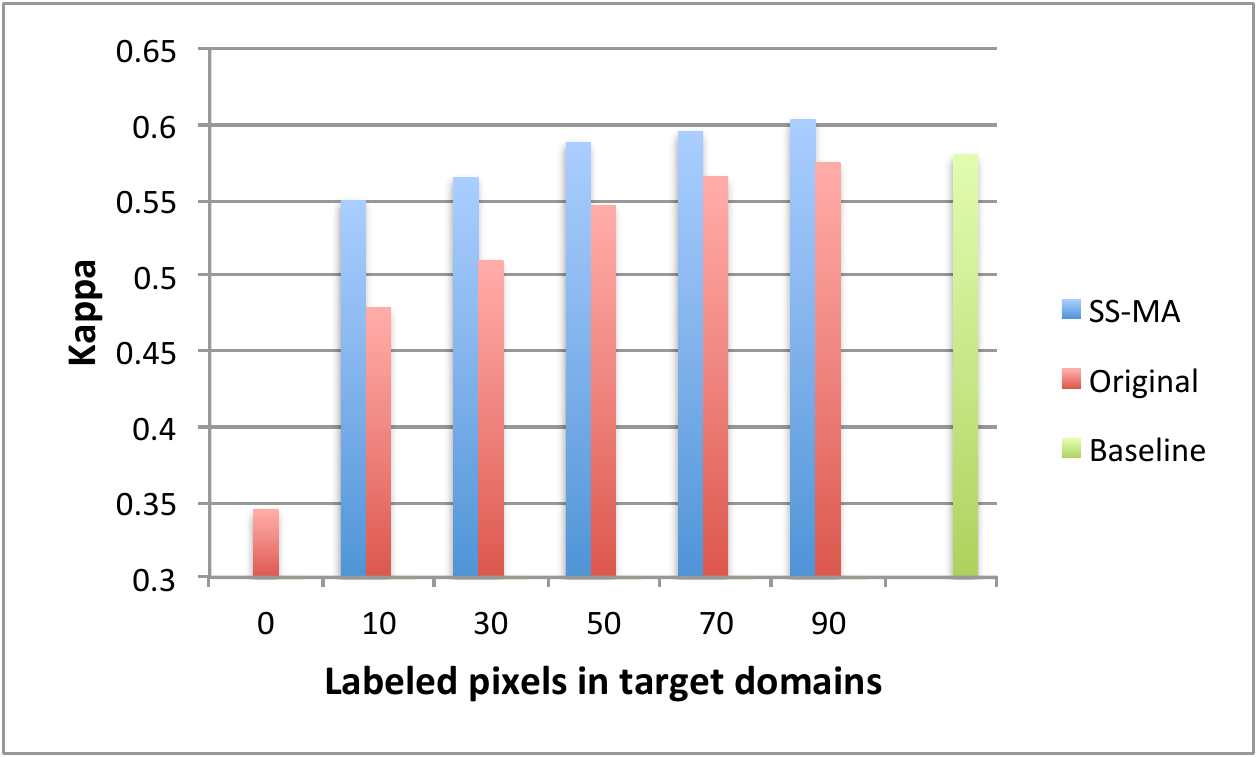}\\

& \rotatebox{90}{Zurich (4 bands)} &
\includegraphics[width=5.2cm]{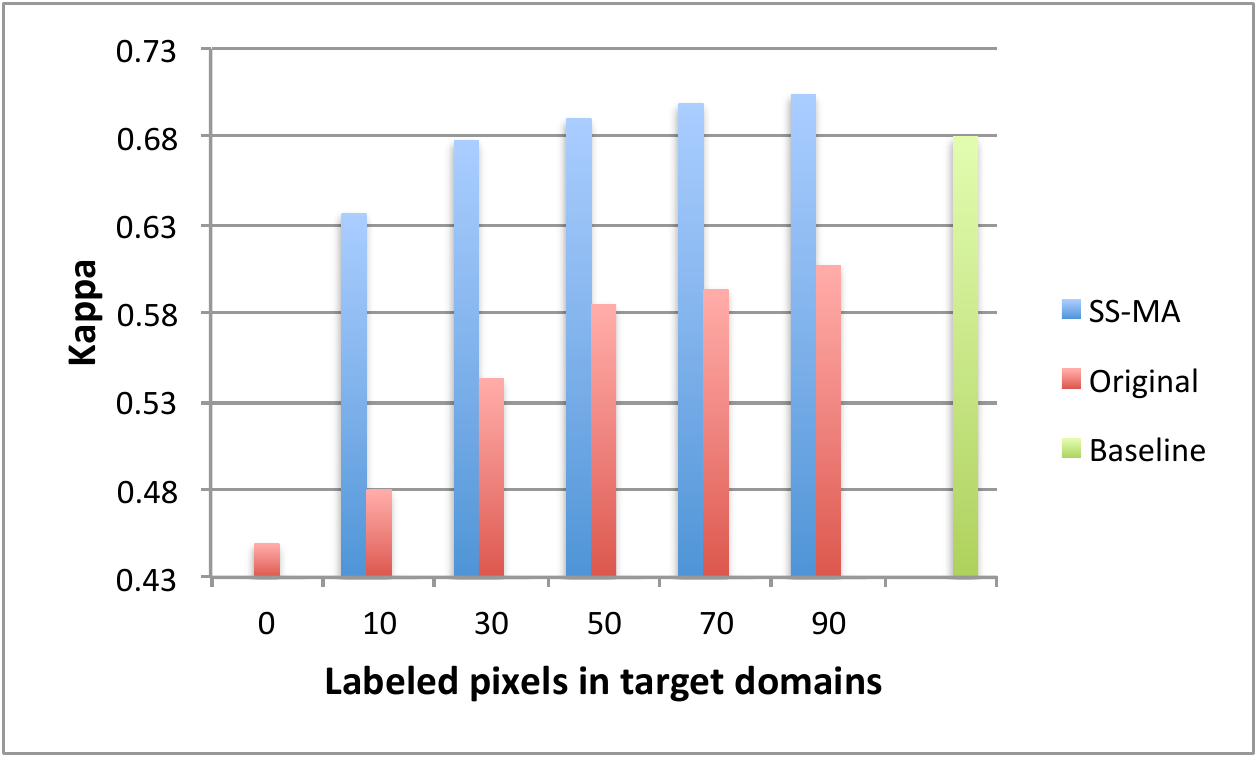}&
\includegraphics[width=5.2cm]{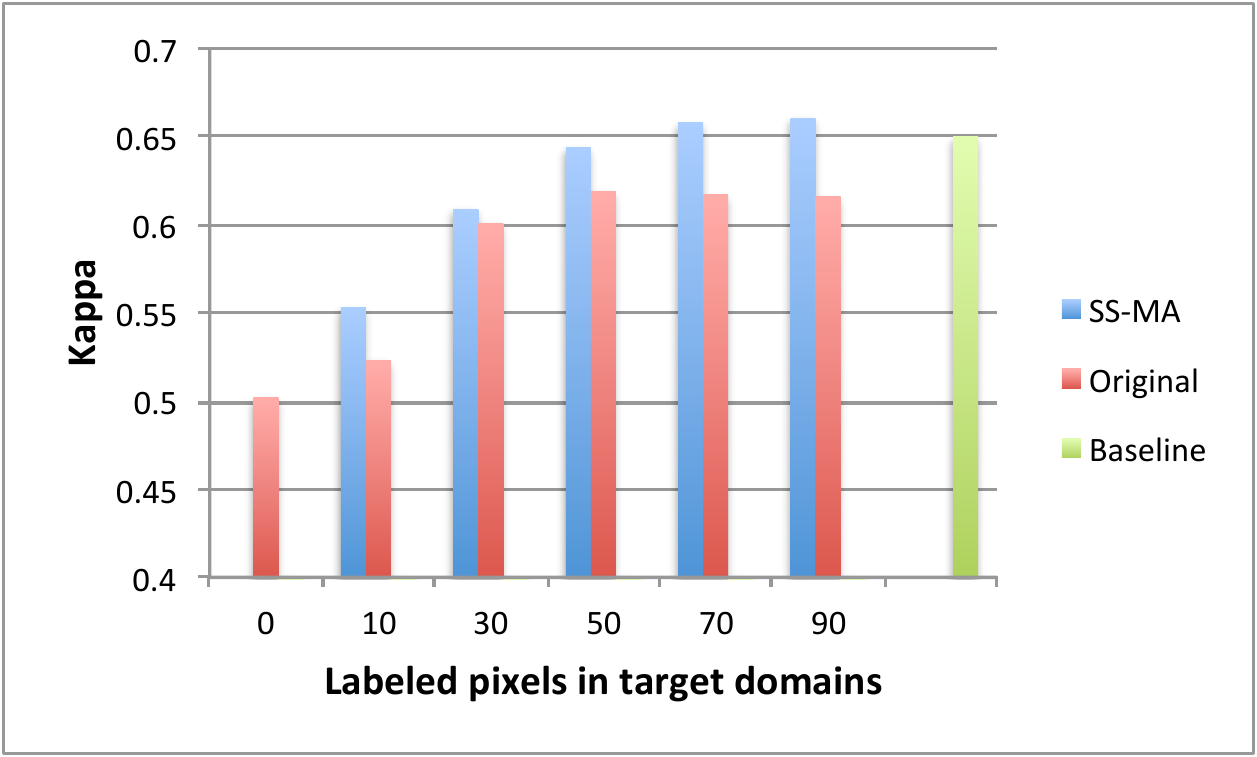}&
\includegraphics[width=5.2cm]{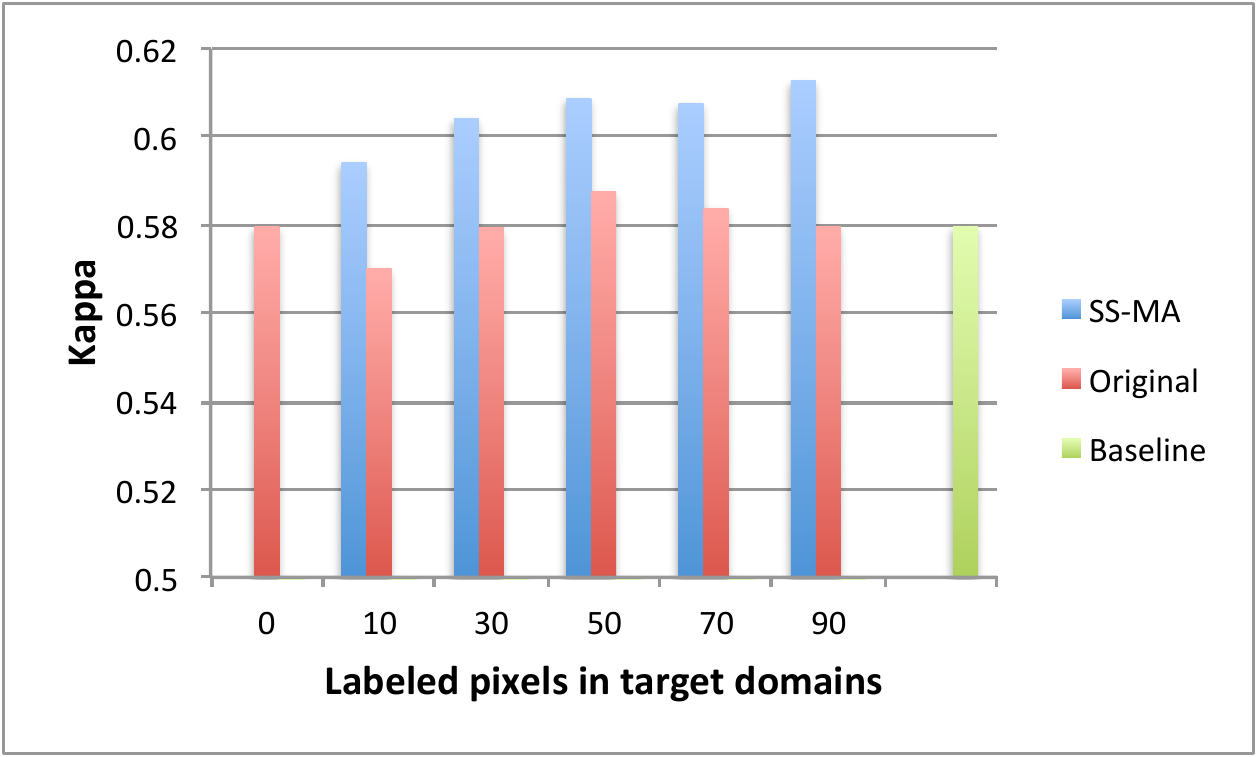}\\
\end{tabular}

\caption{Numerical results for the multisource experiment. Rows indicate the image from which 100 labeled pixels {\em per} class are used ($l_1 = 100$ {\em per} class). $\kappa$ performances for increasing number of labeled pixels in the two other images ($l_{2} = l_3 = [10, ..., 90]$ {\em per} class) are reported. The results correspond to the diagonal of the \blue{surfaces} in Fig.~\ref{fig:3domainsWVQB}.  Columns correspond to the image that has been used for testing. The baseline is the model obtained using 100 pixels {\em per} class from that image only.}
\label{fig:3domainsWVQBdiag}

\end{figure*}

The last experiment deals with multisource data. Figures~\ref{fig:3domainsWVQB} and~\ref{fig:3domainsWVQBdiag} illustrate the performance of the SS--MA projection against cases where no projection is performed and a baseline, where only pixels of the testing domain are used (green bar in Fig.~\ref{fig:3domainsWVQBdiag}). In all the results reported, the richness of the joint latent space provides better classification results, regardless of the dimensionality of the input space.

The trends observed in this experiment are similar to those observed in the multitemporal case, with the difference that the performance of the classifiers trained without adaptation is decreased: to allow multisource classification, the four bands missing in QuickBird were removed, and the performance of the SVMs exploiting a four dimensional input space are thus lower than in the previous case. On the contrary, the aligned data exploit the joint latent space of dimension $d = 8 + 8 + 4 = 20$ and return very high improvements in performances with respect to the unprojected counterpart.

\section{Conclusions}\label{sec:concl}
In this paper, we presented a semisupervised method to align \blue{manifolds of} remote sensing images\blue{. Alignment is performed} in order to enhance model portability in multitemporal, multisource and multiangular classification. The method aligns the datasets by pulling close together pixels of the same class, while pushing those of different classes apart. At the same time, it preserves the geometry of each manifold along the transformation. The method performs well even \blue{in case of} strong deformations and provides models that can classify target domains at least as accurately as if a large set of labeled pixels for that domain were available.

We reported experiments in three problems involving VHR images acquired by different sensors and with different angular properties, that cannot be tackled \blue{in a satisfactory way} with current \blue{relative normalization} strategies, such as univariate histogram matching or PCA matching. None of the image datasets considered in this work were co-registered, thus showing the main interest of the proposed SS--MA with respect to methods such as canonical correlation analysis. Of course, the presence of labeled pixels in each image is a limitation, that we will address in future research.

\section*{Acknowledgements}

The authors would like to thank the IADF Technical Committee of the IEEE and Digital Globe for the access to the multiangular data over Rio. They also would like to thank Adrian Moriette (trainee at LaSIG, EPFL) who, diligently and relentlessly, prepared the ground truths of the World-View 2 images considered in this work. 

\bibliographystyle{IEEEbib}
\bibliography{refURBAN,align}

\end{document}